%% file: acl_latex.tex
\pdfoutput=1

\documentclass[11pt]{article}

\usepackage[]{acl}

\usepackage{times}
\usepackage{latexsym}

\usepackage[T1]{fontenc}

\usepackage[utf8]{inputenc}

\usepackage{microtype}

\usepackage{inconsolata}

\input{math_commands.tex}

\usepackage{hyperref}
\usepackage{url}
\usepackage{graphicx}
\usepackage{adjustbox}
\usepackage{multirow}
\usepackage{multicol}
\usepackage{booktabs}
\usepackage{algorithm}
\usepackage{algpseudocode}
\usepackage{subcaption}
\usepackage{amsmath}
\usepackage{bbm}
\usepackage{fancyvrb} 
\usepackage{enumitem}

\usepackage{listings}

\newcommand*{\affaddr}[1]{#1} 
\newcommand*{\affmark}[1][*]{\textsuperscript{#1}}
\newcommand*{\email}[1]{\texttt{#1}}

\title{Extracting and Encoding: Leveraging Large Language Models and Medical Knowledge to Enhance Radiological Text Representation}

\author{%
Pablo Messina\affmark[1,4,5], René Vidal\affmark[2], Denis Parra\affmark[1,4,5], Álvaro Soto\affmark[1,5], Vladimir Araujo\affmark[3]\\
\affaddr{\affmark[1]Pontificia Universidad Católica de Chile},
\affaddr{\affmark[2]University of Pennsylvania},
\affaddr{\affmark[3]KU Leuven}\\
\affaddr{\affmark[4]Millennium Institute for Intelligent Healthcare Engineering (iHEALTH), Chile}\\
\affaddr{\affmark[5]National Center for Artificial Intelligence 
(CENIA), Chile}\\
\email{\{pamessina,dparras,vgaraujov\}@uc.cl},
\email{vidalr@seas.upenn.edu},\\
\email{asoto@ing.puc.cl}
}

\newcommand{\wrappablebold}[2]{\begin{minipage}[c]{#1}\centering \textbf{#2} \end{minipage}}
\newcommand{\wrappable}[2]{\begin{minipage}[c]{#1}\centering #2 \end{minipage}}

\newcommand{\cxrfactencodershortname}{CXRFE}
\newcommand{\cxrfactencodermetricname}{CXRFEScore}

\begin{document}

\maketitle

\begin{abstract}

Advancing representation learning in specialized fields like medicine remains challenging due to the scarcity of expert annotations for text and images. 
To tackle this issue, we present a novel two-stage framework designed to extract high-quality factual statements from free-text radiology reports in order to improve the representations of text encoders and, consequently, their performance on various downstream tasks.
In the first stage, we propose a \textit{Fact Extractor} that leverages large language models (LLMs) to identify factual statements from well-curated domain-specific datasets. 
In the second stage, we introduce a \textit{Fact Encoder} (CXRFE) based on a BERT model fine-tuned with objective functions designed to improve its representations using the extracted factual data.
Our framework also includes a new embedding-based metric (CXRFEScore) for evaluating chest X-ray text generation systems, leveraging both stages of our approach.
Extensive evaluations show that our fact extractor and encoder outperform current state-of-the-art methods in tasks such as sentence ranking, natural language inference, and label extraction from radiology reports. Additionally, our metric proves to be more robust and effective than existing metrics commonly used in the radiology report generation literature.
The code of this project is available at \url{https://github.com/PabloMessina/CXR-Fact-Encoder}.

\end{abstract}

\section{Introduction}
\label{sec:intro}

In the context of medical image analysis, radiology reports serve as a rich source of information. Radiologists routinely generate these free-text reports, which typically include sections such as \textit{comparison}, \textit{indication}, \textit{findings}, and \textit{impression}, as illustrated in Figure \ref{figure:iu-xray-example}.

Radiology reports have been employed for various purposes, including label extraction for structured supervision in medical image tasks \cite{irvin2019chexpert, chestimagenome, radgraph}, training data for models in report generation \cite{messina2022survey, miura-etal-2021-improving, tanida2023interactive}, summarization tasks \cite{chen-etal-2023-toward, ma2023impressiongpt}, and the development of multimodal models capable of jointly understanding medical images and text \cite{wang2022multi, boecking2022making, bannur2023learning}.

\input{tables/iu-xray-example}

A crucial aspect in addressing such tasks is the accurate comprehension of factual information within the report. Specifically, the \textit{findings} and \textit{impression} sections can be considered as repositories of factual information regarding the imaging examination. These factual statements encompass various elements, including observations (such as abnormalities, diseases, or devices), interpretations derived from observations, references to anatomical locations, discussions on severity or confidence levels, comparisons to previous studies, and more. For instance, in Figure \ref{figure:iu-xray-example}, one factual statement indicates \textcolor{orange}{no acute bone abnormality} (a normal observation), while another describes a \textcolor{purple}{stable calcified granuloma within the right upper lung} (an abnormality found in a specific anatomical site).

Despite the various aforementioned applications that recent research has explored in the use of radiology reports,
the persistent absence of effective methods for precise fact extraction and encoding from medical reports remains a critical challenge. As demonstrated in our experimental evaluation (Section \ref{sec:exp}), existing encoders and label extraction techniques frequently struggle to capture the nuanced details within free-text radiology reports, resulting in incomplete or inaccurate depictions of clinical information. This is evident in various aspects. For instance, existing text encoders developed for the medical domain may struggle to generate consistent representations of paraphrased statements (Table \ref{table:triplet-and-sent-rank-results}) or to differentiate between similar sentences conveying contradictory meanings (Table \ref{table:nli-entcont-results}), a crucial requirement to prevent encoding erroneous diagnoses. Similarly, current label extraction methods often rely on rigid labeling schemes based on manually crafted rules, leading to incomplete capture of all factual statements within a report (Table \ref{table:report-recovery-results}). Similar limitations are also observed in commonly used evaluation metrics for radiology text generation (Table \ref{table:metrics-results}).


In this work, we propose a novel approach that leverages the capabilities exhibited by Large Language Models (LLMs) such as ChatGPT, which have showcased outstanding performance in medical contexts \cite{liu-etal-2023-exploring-boundaries, katz2023gpt, liu2023a, adams2023leveraging}, to improve factual statement representation. Our methodology also takes advantage of the existing knowledge in expert-annotated datasets.
These datasets 
offer indispensable training data 
and also serve as a benchmark to enhance our model's clinical terminology and context comprehension.
Concretely, our contributions are three-fold:

\begin{itemize}
  
    \item A fact extractor: a novel and simple approach to extracting facts that leverages ChatGPT and a fine-tuned version of T5 \cite{raffel2020exploring} to capture relevant information from Chest X-ray radiology reports without requiring annotations from radiologists.

    \item A fact encoder: \textbf{C}hest \textbf{X}-\textbf{r}ay \textbf{F}act \textbf{E}ncoder (CXRFE), a CXR BERT-based model \cite{boecking2022making} fine-tuned with a multi-task approach that leverages domain expertise from radiologists as well as ChatGPT and T5 generated annotations. \cxrfactencodershortname\ exhibits significant improvement in fact comprehension, demonstrated on sentence ranking and natural language inference tasks.
   
    \item A new evaluation metric for radiology text generation: \cxrfactencodermetricname, which measures the factual accuracy of a generated text relative to a real text, by extracting and comparing the similarity of fact embeddings.
    
\end{itemize}

We release the weights of all our models, as well as the data and code necessary to replicate the results. We also release CXRFEScore as a Python library for ease of use by the research community.

\section{Related Work}
\label{sec:related-work}

In this section, we discuss prior work on BERT-based approaches to radiology text representation and label extraction from radiology reports, and leave discussion of prior work on evaluation of factual correctness in radiology text generation, applications of LLMs to medical text, and knowledge distillation from LLMs to
Appendix~\ref{sec:appendix:rel-work}.

\paragraph{BERT-based Approaches for Radiology Text Representation.} The advent of BERT \cite{devlin-etal-2019-bert} has sparked notable progress in numerous NLP domains. This has inspired researchers to customize BERT for specific applications, including the medical field. Pioneering works such as BioClinicalBERT \cite{alsentzer2019publicly}, PubMedBERT \cite{pubmedbert}, and BioLinkBERT \cite{yasunaga2022linkbert} have applied the masked language modeling (MLM) objective introduced by BERT to domain-specific corpora, such as PubMed paper abstracts and MIMIC-III \cite{johnson2016mimic}, an electronic health records dataset.

More recently, specialized variants like CXR-BERT \cite{boecking2022making} and BioViL-T \cite{bannur2023learning} have been developed, targeting the unique challenges posed by CXR reports. CXR-BERT provides both a general version, pretrained with MLM on PubMed abstracts and MIMIC-III documents, and a specialized version, fine-tuned with MLM coupled with a radiology section matching loss specifically tailored for reports from the MIMIC-CXR dataset \cite{johnson2019mimiccxrjpg}. BioViL-T adopts the same pretraining strategy as CXR-BERT but is subsequently fine-tuned using global and local multi-modal contrastive learning and image-informed MLM objectives. By combining reports with temporally sequenced image pairs, this approach enhances the understanding of radiological sentences with temporal descriptions.


Drawing inspiration from these works, we adopt BERT as our base model for text encoding. However, unlike prior approaches that aim to improve BERT's representations with a single pre-text task \cite{reimers-gurevych-2019-sentence,make5010005}, we employ a novel domain-specific multi-task learning protocol. This protocol leverages LLMs to generate large-scale supervision alongside expert-curated annotations from domain-specific datasets.

\paragraph{Label Extraction from Radiology Reports.} Our work is also related to the problem of extracting information, usually in the form of labels, from free-text radiology reports. A well-known example in the literature is the CheXpert labeler \cite{irvin2019chexpert}, which uses a rule-based system to infer the presence or absence of 13 observations (plus the label "No findings"). CheXbert \cite{chexbert} and VisualCheXbert \cite{visualchexbert} are subsequent versions that follow the same labeling standard of CheXpert but are based on BERT. 

The Chest ImaGenome \cite{chestimagenome} dataset is another example that used a rule-based NLP system to label reports to construct scene graphs for the corresponding frontal images in the MIMIC-CXR dataset \cite{johnson2019mimicv1}. RadGraph \cite{radgraph} proposed a labeling scheme of entities and relations for radiology reports and trained a variant of BERT, DyGIE++ \cite{wadden-etal-2019-entity}, for entity and relation extraction on examples annotated by radiologists. PadChest \cite{bustos2019padchest} followed a similar approach, by labeling reports with an LSTM that was previously trained on examples annotated by physicians. 

Our work contributes to this field by introducing a more flexible, open-vocabulary approach to information extraction, focused on extracting the essential factual information contained in the report, without imposing constraints that are too rigid. Specifically, we propose extracting factual statements, referred to as "facts," from reports, by leveraging the proven effectiveness of recent LLMs.

\section{Method}
\label{sec:method}

We introduce a two-stage method for encoding the information within a CXR report. In the first stage, called fact extraction (Section \ref{sec:fact-extraction}), we utilize LLMs to extract facts from the original sentences of the report. In the second stage, called fact encoding (Section \ref{sec:fact-encoding}), we employ a BERT-based text encoder to generate sentence embeddings for each extracted fact. 
When used in tandem, these two stages form a cohesive system capable of producing vectorial representations of the factual statements found within a CXR report.

\subsection{Fact Extraction}
\label{sec:fact-extraction}

\begin{figure*}[!htb]
\begin{center}
\includegraphics[width=\textwidth]{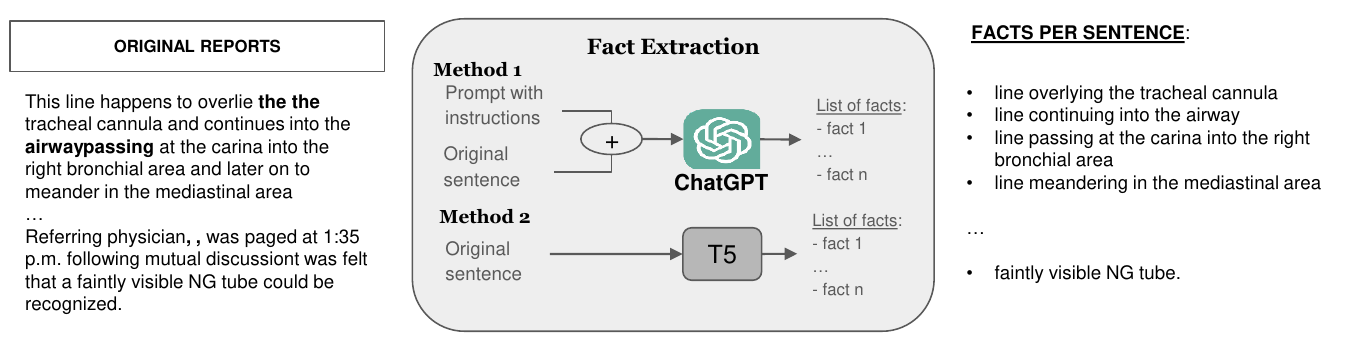}
\end{center}
\caption{Fact extraction from radiology reports, designed to handle noisy input, including repeated words, incorrect sentence tokenization, typos, and verbose sentences, among other issues. When employing ChatGPT, context with instructions is used, whereas T5 is directly applied to the original sentence.}
\label{figure:fact-extraction}
\end{figure*}

Figure \ref{figure:fact-extraction} outlines our method for extracting facts from radiology reports, with examples taken from the MIMIC-CXR dataset \cite{johnson2019mimiccxrjpg}. Initially, we use regular expressions and simple rules to pinpoint relevant radiological sections in MIMIC-CXR reports, mainly \textit{Findings} and \textit{Impression}, but we also handle alternate headings. These sections are then divided into sentences. For simplicity, we use the \textit{sent\_tokenize} function from the NLTK library\footnote{\url{https://www.nltk.org/}}. Next, we proceed to extract concise factual statements from each sentence. The main reason for doing this is that radiologists often write sentences that are noisy or complicated. Figure \ref{figure:fact-extraction} shows two examples of such sentences. The first example contains multiple factual statements in one sentence, which can be simplified into shorter phrases. The second example is overly verbose, but the essential observation can be summarized in a brief phrase. We provide more examples in Table \ref{table:fact-extraction-examples}. Given the recent success of LLMs, an effective strategy to achieve this sort of extraction is by directing ChatGPT using a custom prompt.

\paragraph{T5 as an alternative to ChatGPT.} In theory, this entire process could be executed using off-the-shelf LLMs. However, the expenses associated with accessing the API to annotate the entire dataset can be prohibitive. Therefore, the alternative approach we adopted was to annotate only a strategically selected subset of sentences and then transfer the acquired knowledge from these annotations to a more cost-effective sequence-to-sequence model, such as T5, through fine-tuning. This approach mirrors the strategy employed by Yang et al. \citeyearpar{yang2023language}, where a T5 is fine-tuned to condense verbose descriptions from GPT-3 in LLM-assisted image classification. 

We provide detailed implementation steps for this fact extraction procedure in Appendix \ref{sec:appendix:fact-extraction-details}

\subsection{Fact Encoding}
\label{sec:fact-encoding}

After we extract facts, we encode them to obtain vectors in a latent space via a text encoder model, called \cxrfactencodershortname. In this work, we rely on CXR-BERT \cite{boecking2022making} to implement our fact encoder. Specifically, we use the CXR-BERT-specialized variant available on the Huggingface hub\footnote{\url{https://huggingface.co/microsoft/BiomedVLP-CXR-BERT-specialized}}, leveraging its built-in [CLS] token projection, which yields a 128-D vector serving as the final representation of the text.

Building on top of CXR-BERT-specialized, we explore 6 different training tasks to enhance the latent representation of radiological sentences: triplet loss for sentence ranking (T), natural language inference (NLI), quadruplet loss to enforce a separation between entailment and contradiction pairs (EC), entity and relation extraction (ER), sentence classification tasks (C), and sentence decoding (SD). We provide details on the implementation of each task in Appendix \ref{sec:appendix:tasks}.

Thus, by combining the two stages, the whole framework can accurately extract and encode facts from CXR reports, thus providing a rich and convenient representation of the factual information for downstream applications.

\subsection{\cxrfactencodermetricname}
\label{sec:cxrfescore}

\begin{figure*}[!htb]
\begin{center}
\includegraphics[width=\textwidth]{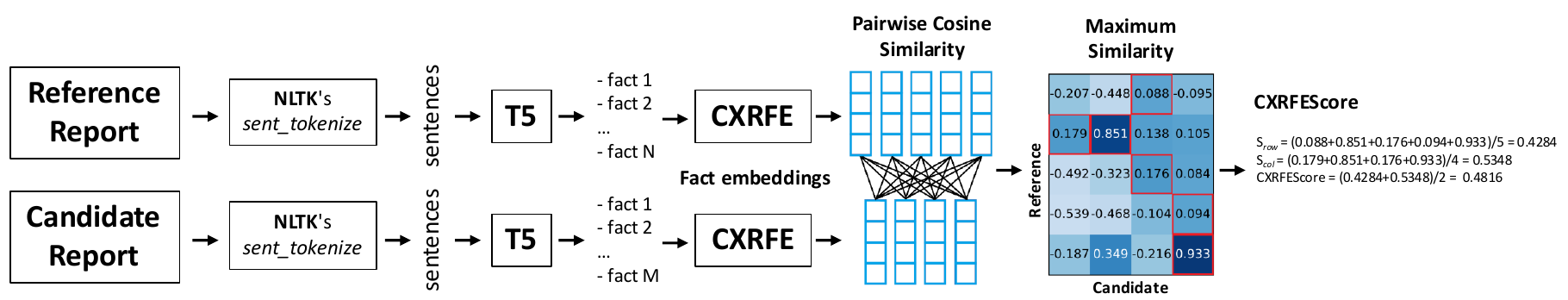}
\end{center}
\caption{Illustration of the computation of \cxrfactencodermetricname. Given a reference report and a candidate report, we employ NLTK's \textit{sent\_tokenize} to extract sentences from each, extract facts from these sentences using T5, generate embeddings from the facts using \cxrfactencodershortname, and finally compute the pairwise cosine similarity matrix. Greedy matching is highlighted in red, with the calculation of the final score explicitly shown on the right. This figure's design is inspired by Figure 1 from BERTScore \cite{bert-score}.}
\label{figure:cxrfescore}
\end{figure*}

A potential application of our framework is in evaluating report generation from chest X-rays. 
We introduce \cxrfactencodermetricname, an embedding-based metric that leverages both T5 and our \cxrfactencodershortname\ model for evaluation. Figure \ref{figure:cxrfescore} illustrates how the metric works.

Given a reference report and a candidate report, we extract facts from each and represent them as embedding vectors, denoting the sets for the reference and candidate reports as \( R \) and \( C \) respectively. The cosine similarity matrix \( M \) of size \( |R| \times |C| \) is formed, where \( M_{i,j} \) represents the cosine similarity between the \( i^{th} \) vector of \( R \) and the \( j^{th} \) vector of \( C \). This allows us to calculate \( S_{\text{row}} \), \( S_{\text{col}} \), and the final \cxrfactencodermetricname\ as follows:

\vspace{0.2cm}
\begin{tabular}{cc}
\( S_{\text{row}} = \frac{\sum_{i} \max_{j} M_{i,j}}{|R|} \) & \( S_{\text{col}} = \frac{\sum_{j} \max_{i} M_{i,j}}{|C|} \) \\
\end{tabular}
\vspace{0.1cm}


\begin{tabular}{c}
\( \text{CXRFEScore} = \frac{S_{\text{row}} + S_{\text{col}}}{2} \) \\
\end{tabular}
\vspace{0.1cm}

The equations of this metric resemble those of BERTScore \cite{bert-score}. The key difference lies in \cxrfactencodermetricname's comparison of fact embeddings rather than token embeddings. This metric illustrates the fusion of the two proposed stages: fact extraction (implemented with T5) and fact encoding (implemented with \cxrfactencodershortname). We provide evidence of the robustness of this metric compared to many existing metrics in Section~\ref{sec:exp}.

\subsection{Dataset construction}
\label{sec:datasets}

In our experiments, we primarily utilize the MIMIC-CXR dataset \cite{johnson2019mimiccxrjpg}, which comprises 227,827 radiology reports associated with 377,110 chest X-ray images. However, we focus solely on utilizing the reports for our experiments, deferring the exploration of images and multi-modality for future work. Additionally, we incorporate annotations from the Chest ImaGenome dataset \cite{chestimagenome}, which provides scene graphs linking observations to anatomical image locations for each frontal view image in MIMIC-CXR. These annotations serve two main purposes: facilitating our creation of a binary multi-label classification task and introducing a radiologist-informed annotation standard covering various observation types and anatomical locations. Similarly, we utilize RadGraph \cite{radgraph}, which offers an entity and relation annotation scheme for radiology reports in MIMIC-CXR, and datasets such as MedNLI \cite{romanov2018lessons}, RadNLI \cite{miura-etal-2021-improving} and MS-CXR-T \cite{bannur2023learning} for experiments on Natural Language Inference (NLI). To assess the performance and generalization ability of our two-stage framework, we also evaluate \cxrfactencodermetricname\ using the 3955 reports and associated tag annotations of the IU X-ray dataset \cite{10.1093/jamia/ocv080}.  It is important to highlight that radiologists or doctors have partially or fully annotated these datasets, which adds significant value to training and evaluation. We direct the reader to Appendix \ref{sec:appendix:tasks} for further details on these datasets and their utilization in our experiments.

\subsubsection{LLM-assisted data augmentations for training supervision}
\label{sec:datasets:augmenting-facts}

As mentioned in Section \ref{sec:fact-encoding}, we use six distinct tasks to train \cxrfactencodershortname. For certain tasks, we leverage ChatGPT to generate additional training data. We elaborate in depth on these aspects in Appendix \ref{sec:appendix:tasks}, so here we only offer a concise overview.

For the triplet loss task (detailed in Section \ref{sec:datasets:triplets}), some triplets incorporate paraphrased facts, which we generate using ChatGPT (an example is depicted in Figure \ref{figure:chatgpt-prompts:fact2paraphrases}). Similarly, we employ ChatGPT to produce challenging triplets, as indicated by the prompt in Figure \ref{figure:chatgpt-prompts:hard-triplets}.

In classification-related tasks, each fact is annotated with a JSON metadata object using ChatGPT. This object encompasses fields such as \textit{anatomical location}, \textit{detailed observation}, \textit{short observation}, \textit{category}, \textit{health status}, and \textit{comparison status}. The respective prompt for this process is shown in Figure \ref{figure:chatgpt-prompts:fact2metadata}. To refine the "comparison status" field, we use another prompt showcased in Figure \ref{figure:chatgpt-prompts:fact2comparison}. Also, we employ ChatGPT to label according to the scheme of the Chest ImaGenome dataset, producing additional observation and anatomical location labels, as shown in Figures \ref{figure:chatgpt-prompts:fact2obs} and \ref{figure:chatgpt-prompts:fact2anat}.

For the task of natural language inference, we extensively utilize GPT-4 to generate training examples with distinct prompts (refer to Figures \ref{figure:chatgpt-prompts:nli:p2ecn}, \ref{figure:chatgpt-prompts:nli:ex2sim}, \ref{figure:chatgpt-prompts:nli:nli-simple}, \ref{figure:chatgpt-prompts:nli:nli-cot-ex}, and \ref{figure:chatgpt-prompts:nli:p2cont}).

\subsubsection{Triplet Sampling}
\label{sec:datasets:triplets}

\cxrfactencodershortname\ is trained to generate sentence embeddings that cluster semantically similar sentences in the embedding space through a triplet ranking task with binary cross-entropy loss. This approach uses a dataset of triplets, each one with an anchor, a positive sample (akin to the anchor), and a negative one. The difference in similarities is computed as \(\Delta \text{sim}(a, p, n) = \text{sim}(a, p) - \text{sim}(a, n)\) from their embeddings' dot product. By minimizing the binary cross-entropy loss, the encoder ensures closely related sentences are nearer and unrelated ones are more distant in the embedding space.
    
We define six triplet sampling rules to guide the selection process. \textbf{Rule 1} prioritizes paraphrases generated with ChatGPT. \textbf{Rule 2} involves sampling triplets based on the consensus of BioVIL-T and Levenshtein distance, with the anchor and positive sample sharing the same health status. \textbf{Rule 3} ensures proximity between short observations, detailed observations, and original facts, along with their paraphrases. \textbf{Rule 4} samples triplets based on Chest ImaGenome labels, ensuring that the anchor and positive sample share at least one label and that BioVIL-T and Levenshtein distance agree. \textbf{Rule 5} ranks triplets according to the overlap of entities and relations from RadGraph. \textbf{Rule 6} includes hard triplets generated by ChatGPT. For each rule, we sampled around 3 to 4 million training triplets and 1,000 each for validation and testing.

These rules encapsulate specific intuitions and heuristics regarding the ranking of sentence embeddings within the semantic space. The design of these sampling rules and the construction of the triplets dataset involve several technical details, which are elaborated upon in Appendix \ref{sec:appendix:triplets-dataset-details}.

\subsubsection{Natural Language Inference}
\label{sec:datasets:nli}

\input{tables/triplet-and-sentence-ranking-results}

Natural Language Inference (NLI) aims to classify the relationship between a premise and a hypothesis into one of three categories: "entailment", "neutral", or "contradiction". For instance, consider a premise stating \textit{``There are no evident signs of pleural effusion''}, while a hypothesis asserts \textit{``There are evident signs of pleural effusion''}. Despite their structural and lexical similarities, these sentences express contradictory meanings. A robust sentence embedding should be able to distinguish between sentences conveying contradictory diagnoses. Thus, our objective in leveraging NLI during training is to refine sentence embeddings to discern such subtle distinctions accurately.

All MedNLI splits \cite{romanov2018lessons} are used for training, amounting to 14,049 annotated sentence pairs. Radiology-specific datasets include RadNLI \cite{miura-etal-2021-improving} with 960 pairs and MS-CXR-T \cite{bannur2023learning}, an evaluation set with 361 pairs emphasizing condition evolution over time. Given the dearth of training data, we build a custom NLI dataset combining MedNLI (14,049), RadNLI development set (480) and GPT-4 generated examples (154,498), resulting in a total of 169,025 pairs (entailment: 25,175, neutral: 44,729, contradiction: 99,121), while RadNLI test set (480) and MS-CXR-T (361) are set apart for evaluation.

\section{Experimental Results}
\label{sec:exp}

Our experiments evaluate different versions of \cxrfactencodershortname, each defined by a subset of the six tasks (T, C, NLI, EC, ER, SD) outlined in Section \ref{sec:fact-encoding}. This results in a total of 64 potential combinations. However, for the sake of simplicity, we heuristically assess only 11 combinations. Further details on our rationale are provided in Appendix~\ref{sec:appendix:tasks}.

\textbf{Triplet Ranking}. We evaluate \cxrfactencodershortname\ and multiple baselines from the literature on triplet ranking accuracy, using a separate test set of 1000 triplets per rule sampled according to the sampling rules detailed in Section \ref{sec:datasets:triplets}. The left side of Table \ref{table:triplet-and-sent-rank-results} presents these results. Notably, all different versions of \cxrfactencodershortname\ outperform all the baselines in triplet rules where ChatGPT is heavily involved, namely, paraphrases (R1(o), R1(a), R3) and hard triplets (R6). The hard triplets are especially challenging for the baselines, with BioViL-T only achieving 0.765 accuracy (row 6), while the best performing version of \cxrfactencodershortname\ achieves 0.952 (row 17). In addition to triplet loss (T), which is key to learning an embedding for these rules, we notice that sentence decoding (SD) and classification (C) appear to be helpful auxiliary tasks since most of the best scores are achieved by variants that include them (rows 8, 10, 15, 17).

\begin{table*}[ht]
         \input{tables/nli-entcont-results}
 
 \end{table*}

\textbf{Sentence Ranking}. To complement the triplet ranking evaluation, which is based on heuristic sampling rules, we conduct a sentence ranking evaluation using 2,412 carefully annotated sentences provided by radiologists from the gold standard of the Chest ImaGenome dataset \cite{chestimagenome}. These sentences are annotated with a vocabulary of 70 observations (\textit{yes} (1), \textit{no} (0), or \textit{omitted} (-1)), and 38 anatomical locations (\textit{mentioned} (1) or \textit{unmentioned} (0)), resulting in a discrete vector of size 108 for each sentence. 

We report the results of this evaluation on the right side of Table \ref{table:triplet-and-sent-rank-results}. In this evaluation, each sentence serves as a query, against which all other sentences are ranked based on the cosine similarity of their embeddings. AUC, acc@k, and cont@k are computed by comparing the labels of the query sentence with those of each ranked sentence. AUC requires defining which sentences are relevant (1) or irrelevant (0) for the query. For two sentences $a$ and $b$, $a$ is deemed relevant for $b$ if $a$'s labels logically entail $b$'s or vice versa; otherwise, they are considered irrelevant. acc@k represents the mean average accuracy up to the $k$th sentence in the ranking, while cont@k represents the mean number of sentences contradicting the query up to the $k$th sentence, by having contradictory values in at least one observation (1 vs. 0). 

\begin{table}[ht]
\input{tables/radnli-results}
\end{table}

CheXbert (row 4) consistently performs the strongest among the baselines, likely due to its training in a similar multi-label classification task. \cxrfactencodershortname\ achieves the best overall performance. Notably, the T+C variant (row 8), combining triplet loss (T) and classification (C), attains the highest AUC and accuracy, while T+EC+NLI (row 13), containing tasks designed for pulling contradictory sentences apart, yields the smallest cont@k scores (where smaller is better). Rows 14-17 represent intermediate points between these two extremes. 



\textbf{NLI}. Table \ref{table:nli-entcont-results} presents the NLI results using cosine similarity between sentence vectors, following a methodology akin to Bannur et al. \citeyearpar{bannur2023learning}. This methodology specifically focuses on \textit{entailment} and \textit{contradiction} pairs, aiming to assess the efficacy of a text embedding in distinguishing between the two given a similarity threshold. We present results across three datasets: our NLI custom dataset mentioned in Section \ref{sec:datasets:nli}, the RadNLI test set, and MS-CXR-T. Our reported results are based on thresholds fine-tuned in the NLI custom dataset, alongside upper bounds obtained by tuning thresholds within the same data used for evaluation.

Notably, 
employing the quadruplet entailment/contradiction loss (EC) and natural language inference (NLI) (rows 11 to 17) is essential to achieve high performance, significantly outperforming all baselines.
In contrast, variants lacking EC and NLI (rows 7 to 10) exhibit a weaker result.

\input{tables/metrics-results}
\input{tables/report-recovery-results}

Among the baselines, CheXbert (row 4) demonstrates superior performance on RadNLI, while BioVil-T (row 6) is the clear victor on MS-CXR-T. However, all baselines struggle considerably in our NLI custom dataset and are outperformed across all three datasets by variants 11-17 of \cxrfactencodershortname.

Additionally, Table \ref{table:radnli-results} presents the accuracy achieved on the RadNLI test set in the context of the typical 3-class classification task encompassing entailment, contradiction, and neutral classes. In this evaluation, we exclusively assess variants of \cxrfactencodershortname\ equipped with an NLI classification head (rows 9-15). For insights into implementing NLI classification, please consult Figure \ref{figure:task-NLI}. 

Within the existing literature, the strongest baseline identified is DoT5 \cite{10.1162/tacl_a_00585} (82.1), employing a sophisticated sequence-to-sequence approach based on T5. Furthermore, we conducted evaluations on GPT-4 (rows 5-6) and 
Meta-Llama-3-8B \cite{llama3modelcard} (rows 7-8) utilizing two distinct prompts: a simple prompt (Figure \ref{figure:chatgpt-prompts:nli:nli-simple}) and a prompt with Chain-of-Thought (CoT) + examples (Figure \ref{figure:chatgpt-prompts:nli:nli-cot-ex}). Notably, the second prompt led to a significant performance boost for GPT-4 (from 82.3 to 89.0), whereas Meta-Llama-3-8B, an open-source LLM from Meta, only experienced a moderate improvement (from 58.1 to 61.5), with very low accuracies overall. Consequently, GPT-4 with the second prompt was selected as our "oracle" for generating additional training data (more details on this in Appendix \ref{sec:appendix:tasks}).

Most versions of \cxrfactencodershortname\ showed superior performance compared to the baselines. Surprisingly, a version fine-tuned explicitly for NLI (row 15) even outperforms GPT-4 with CoT (row 6) by a narrow margin (89.8).

\textbf{CXRFEScore vs. existing metrics}. To assess the quality of our proposed metric, we conduct an evaluation of \cxrfactencodermetricname\ alongside multiple metrics from the literature, as shown in Table \ref{table:metrics-results}. This assessment encompasses four settings: (1) a sentence ranking evaluation using 2412 sentences, (2) a report ranking evaluation with 500 reports, both sourced from the gold standard of Chest ImaGenome, (3) a report ranking evaluation with 3955 reports from the IU X-ray dataset \cite{10.1093/jamia/ocv080} leveraging the manual and automatic tags associated with each report, and (4) a natural language inference evaluation utilizing entailment (336) and contradiction (424) pairs from RadNLI and MS-CXR-T. Note that all these datasets are 
annotated by radiologists, thus serving as 
gold standards for metric comparison.

Among the baseline metrics, RadGraph F1 (rows 10-11) emerges as one of the most promising based on its performance on Chest ImaGenome Gold, which aligns with the findings of Yu et al. \citeyearpar{yu2022evaluating}. However, BERTScore's results (row 5) on Chest ImaGenome Gold are quite similar, achieving the highest AUC among the baselines (0.840). Additionally, BERTScore achieves the highest Jaccard index scores on IU X-ray among the baselines. Notably, CheXbert (rows 8-9), closely followed by CheXpert (rows 6-7), shows the fewest contradictions on Chest ImaGenome Gold. 

All baseline metrics, however, are surpassed by \cxrfactencodermetricname\ (rows 12-15) in all the evaluation metrics. A particularly striking observation is that the baseline metrics struggle significantly to differentiate between entailed and contradictory sentences, as indicated by the AUC results in the last column of Table \ref{table:metrics-results}. RadGraph F1 Full achieves an AUC of only 0.610, whereas the best version of \cxrfactencodermetricname\ (row 15) achieves an AUC of 0.938. This suggests that current metrics assign elevated scores to pairs of sentences with contradictory semantics, highlighting the necessity for improved metrics to discern these subtleties—precisely what \cxrfactencodermetricname\ is designed to accomplish.

Similarly, \cxrfactencodermetricname\ outperforms all the baselines 
on IU X-ray, a dataset not used to develop the metric. This provides valuable evidence of the metric's ability to generalize to radiology reports from a different institution.

We provide additional details and results about these metrics in Appendix \ref{sec:appendix:metrics-details}.

\textbf{Fact Extraction Quality}. To evaluate the quality of the fact extraction stage, we interpret these facts as \textit{open-vocabulary} labels and compare them against three existing radiology report label extraction methods: CheXpert labeler \cite{irvin2019chexpert}, CheXbert \cite{chexbert}, and Chest ImaGenome \cite{chestimagenome}. For Chest ImaGenome, we use the labels from the dataset's scene graphs, as the original NLP algorithm is not publicly available. For fact extraction, we compare T5-small, fine-tuned specifically for this task, against GPT-4 and Meta-Llama-3-8B. The latter two models use the prompt shown in Figure \ref{figure:chatgpt-prompts:fact-extraction}. Our evaluation protocol involves the following: for each MIMIC-CXR test set report and label extraction method, labels are extracted, converted into a report, and evaluated against the original report by several metrics. We adopt the templates suggested by Pino et al. \citeyearpar{pino2021clinically} for CheXpert labeler and CheXbert, while for Chest ImaGenome, we utilize basic templates such as ``(no) \{observation\} in \{anatomical location\}''. For fact extraction, we simply concatenate the facts. These template-based reports are illustrated in Tables \ref{table:template-based-reports} and \ref{table:template-based-reports-2}.


Table \ref{table:report-recovery-results} presents these results. Notably, fact-based reports generated by T5 (row 4) achieve most of the top scores, even slightly outperforming GPT-4 and Meta-Llama-3-8B, demonstrating the efficacy of the fact extraction process. Interestingly, \cxrfactencodermetricname\ suggests that Chest ImaGenome outperforms both the CheXpert labeler and CheXbert, which is reasonable given Chest ImaGenome's broader range of labels, although the improvement is rather marginal. However, it is evident that all three baseline methods fall short of fully capturing the factual information within the reports, likely due to their rigid annotation rules. This shortcoming is highlighted by their results in \cxrfactencodermetricname, RadGraph F1, BERTScore, and most of the other metrics, compared to the fact extraction methods.

\section{Conclusions \& Future Work}
\label{sec:conclusions}

In this work, we present a novel two-stage framework for extracting and encoding factual information from radiology reports. The first stage, fact extraction, uses ChatGPT and T5 to extract factual statements. The second stage, fact encoding, introduces \cxrfactencodershortname, a specialized variant of CXR-BERT, fine-tuned through multitask learning by incorporating tasks that support representation improvement. Our system's effectiveness is validated through comprehensive evaluations. Additionally, we introduce \cxrfactencodermetricname, a novel metric for evaluating radiology text generation, leveraging our two-stage system. We anticipate that our work will stimulate further research in enhanced fact extraction and representation, LLM utilization, advanced training methodologies, and improved evaluation metrics. For future work, we aim to 
expand our framework to integrate visual modality, focusing on image-based fact detection and visual grounding.

\section*{Limitations}
\label{sec:limitations-and-future-work}


Our study acknowledges several limitations and highlights areas for improvement. First, more expert evaluations, particularly from radiologists, are needed to refine the use of large language models (LLMs) in radiology. Although we extensively utilized publicly available gold standards, such as those from the Chest ImaGenome dataset, RadNLI, and MS-CXR-T, there remains room for improvement. For instance, involving radiologists in the prompt engineering process and developing more rigorous evaluation protocols are two strategies we believe will enhance the evaluation and utilization of LLMs for radiological text.

We also see potential in designing better triplet sampling heuristics, especially with input from radiologists. Optimizing LLM prompts for triplet sampling and incorporating more advanced auxiliary embeddings could further enhance our approach.

Furthermore, while our study focuses on text-only analysis, we recognize the importance of integrating visual data, such as chest X-ray images, into a multimodal framework. Devising a training protocol that effectively combines supervision from both images and text is an area of potential improvement for future work.

In this work, we limited our experiments to the sections "findings," "impression," and similar headings providing factual statements about the imaging exam. However, other sections, such as "comparison," "indication," and "history," were left out of the analysis, yet they can provide deeper insights into patient information and context. Investigating how this broader information can be extracted and encoded to enhance downstream applications is another avenue for future exploration and potential improvement.

Lastly, we acknowledge that our fact extraction algorithm may be limited due to its reliance on the \textit{sent\_tokenize} function of the NLTK library, which we use to obtain a preliminary division of the report into coarse sentences (before fact extraction). This method could falter when a fact spans multiple sentences connected through co-reference. While such occurrences are relatively uncommon in our observations, a deeper exploration of this linguistic aspect could guide the development of a more refined fact extraction mechanism that overcomes this challenge.

\section*{Acknowledgements}

This work was funded by the Chilean National Agency for Research and Development (ANID), including Instituto Milenio en Ingeniería e Inteligencia Artificial para la Salud (iHEALTH) ICN2021\_004; Centro Nacional de Inteligencia Artificial (CENIA) FB210017; Fondecyt regular 1231724; Fondecyt 1221425; and the ANID Scholarship Program / Doctorado Becas Chile / 2019 - 21191569.
Additionally, Pablo was supported by the National Institutes of Health (NIH) grant 1R01AG067396. 
We are grateful for the support from all funding sources mentioned above.

\bibliography{custom}

\appendix
\section{Appendix}
\label{sec:appendix}

\subsection{Additional Related Work}
\label{sec:appendix:rel-work}

\textbf{Evaluation of Factual Correctness in Radiology Text Generation}. One important area of application motivating this work is the evaluation of factual correctness in systems that generate radiological text, usually from input medical imaging. Recent research emphasizes enhancing the accuracy of generated facts in applications like report generation \cite{miura-etal-2021-improving, delbrouck-etal-2022-improving, pino2020inspecting, pino2021clinically} and summarization \cite{zhang-etal-2020-optimizing, delbrouck2023overview}. Pino et al. \citeyearpar{pino2020inspecting} conducted an evaluation of several trivial report generation baselines using established metrics such as BLEU \citeyearpar{papineni2002bleu}, ROUGE-L \citeyearpar{lin-2004-rouge}, and CIDEr-D \citeyearpar{vedantam2015cider}. They achieved results comparable to state-of-the-art papers at the time. However, when assessed using the CheXpert labeler \cite{irvin2019chexpert}, a domain-specific NLP tool designed to detect 13 findings, the performance was notably poor, underscoring the urgent necessity for standardizing improved evaluation metrics among researchers. More recently, Delbrouck et al. \citeyearpar{delbrouck-etal-2022-improving} repurposed RadGraph's entity and relation extraction model \cite{radgraph} to create a factual correctness reward. This reward measures the overlap of entities and relations between real and generated reports, serving as the guiding signal to optimize a report generation model through reinforcement learning. Interestingly, their proposed reward aligns functionally with the RadGraph F1 metric introduced by Yu et al. (\citeyear{yu2022evaluating}). Yu et al. conducted a study on metrics for radiology report generation, determining that RadGraph F1 and BLEU show the highest correlation with radiologists' judgement. Recently, the RadSum23 challenge \cite{delbrouck2023overview} evaluated multimodal radiology report summarization quality using BLEU-4 \citeyearpar{papineni2002bleu}, ROUGE-L \citeyearpar{lin-2004-rouge}, BERTScore \citeyearpar{bert-score}, CheXbert F1 \citeyearpar{chexbert}, and RadGraph F1 \citeyearpar{radgraph}.


Our work is highly relevant in this domain because of our development of the \cxrfactencodermetricname\ metric. This metric leverages the strengths of both stages within our framework: fact extraction and encoding. As a result, \cxrfactencodermetricname\ is specifically designed to assess the factual accuracy of generated radiological text against a reference text. Section \ref{sec:exp} offers a comprehensive evaluation, showcasing the effectiveness of our two-stage system. This includes a comparison of \cxrfactencodermetricname\ with commonly used metrics in the literature, with very favorable results.

\textbf{LLMs in Medicine}. Our work falls under the category of applications of LLMs to the medical domain. Specifically, in this work we make use of ChatGPT versions GPT-3.5 and GPT-4 through OpenAI's API\footnote{{\url{https://platform.openai.com/}}}. Recent works have shown the effectiveness of ChatGPT applied to medical tasks. Most notably, recent work by Liu et al. \citeyearpar{liu-etal-2023-exploring-boundaries} explored the boundaries of GPT-4 in radiology, evaluated in tasks such as classification, summarization, and natural language inference, with remarkable performance. Liu et al. (\citeyear{liu2023a}) employed ChatGPT to generate short sentences with plausible symptoms of medical conditions for interpretable zero-shot medical image diagnosis. Adams et al. (\citeyear{adams2023leveraging}) used GPT-4 to transform free-text radiology reports into structured templates, with remarkable results. GPT-4 is also known for having passed the bar exam \cite{katz2023gpt}. Inspired by these results, we make extensive use of ChatGPT to produce abundant annotations through diverse prompts.

\textbf{Knowledge Distillation from LLMs}.
Our approach can also be viewed as a form of LLM knowledge distillation, where a large language model (LLM) "teacher" generates annotations for training a more compact "student" model. Shi et al. \citeyearpar{shi2023chatgraph} illustrated this by using ChatGPT to extract knowledge graphs from text to train a smaller model for text classification. Similarly, Gu et al. \citeyearpar{gu2023distilling} applied this concept in the biomedical field, distilling knowledge from GPT-3.5 for adverse drug event extraction with student models like PubMedBERT and BioGPT. In line with these works, our research presents a form of knowledge distillation as we fine-tune T5 (student) using annotations generated by ChatGPT (teacher). This process trains an imitator that is more cost-effective and easier to run on our machines. We apply this method to the tasks of fact extraction from reports and the generation of metadata and labels for each fact. Additionally, we extensively use GPT-4 to produce high-quality NLI labels. Although we don't use T5 in this case, these labels serve as part of the training supervision for \cxrfactencodershortname, which can also be considered a form of knowledge distillation.

\subsection{Fact Extraction Implementation Details}
\label{sec:appendix:fact-extraction-details}

In our experiments, we processed the 227,827 radiology reports provided by the MIMIC-CXR dataset \cite{johnson2019mimiccxrjpg}. To pinpoint relevant radiological sections in the MIMIC-CXR reports, such as \textit{Findings}, \textit{Impression}, and various other headings, we employed a combination of regular expressions and simple rules. These sections were then segmented into sentences using NLTK's \textit{sent\_tokenize} function, resulting in 677,694 unique sentences after processing the entire dataset. Subsequently, we extracted facts from each sentence.

\begin{figure*}[!htb]
\begin{center}
\includegraphics[width=\textwidth]{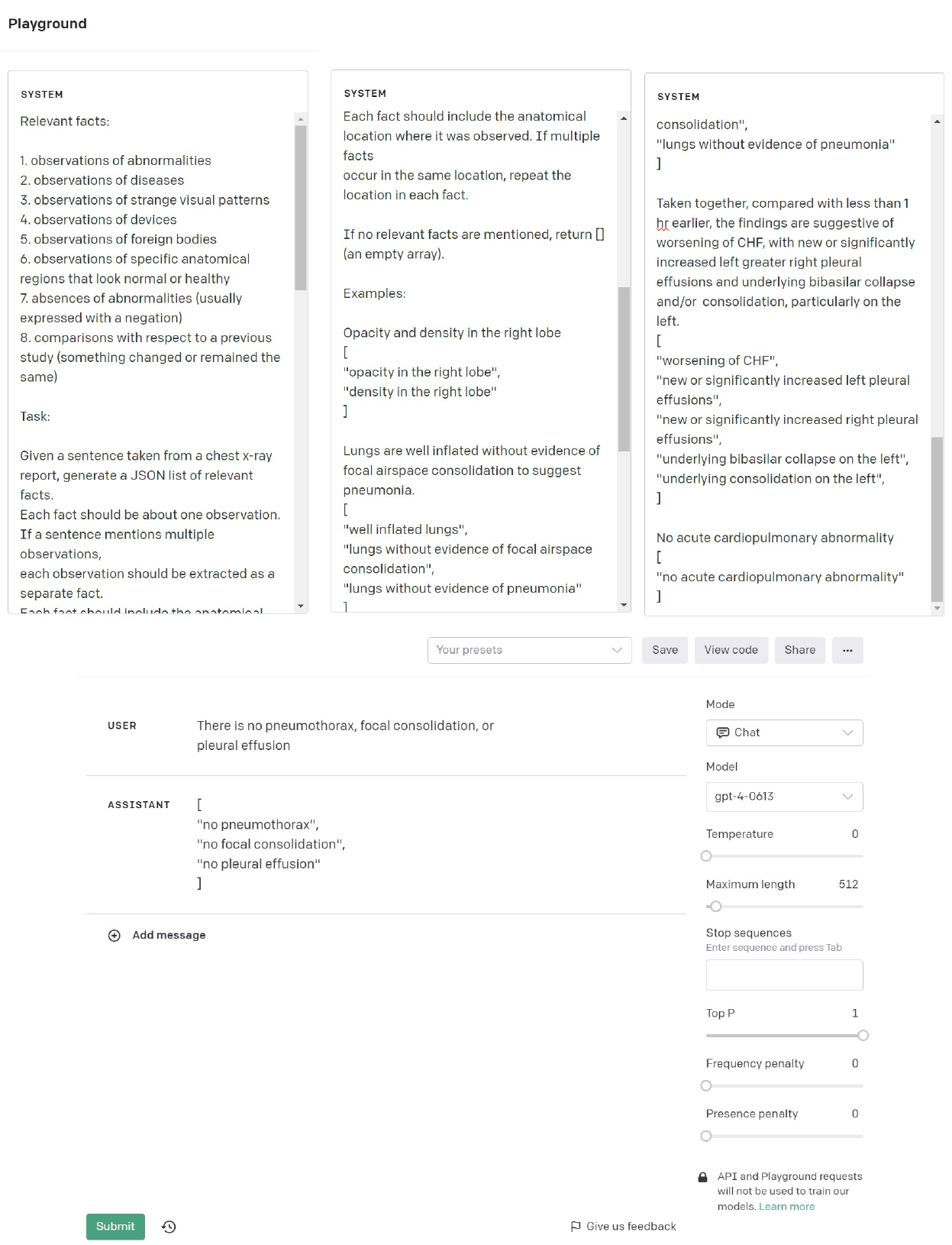}
\end{center}
\caption{Screenshot of OpenAI's playground web  interface showing the prompt used to extract facts from chest X-ray report sentences. The screenshot was manipulated in order to display the full instructions given to ChatGPT.}
\label{figure:chatgpt-prompts:fact-extraction}
\end{figure*}

Extracting factual statements from a sentence of a free-text radiology report using traditional approaches, such as regular expressions, hand-designed rules, and similar heuristics, poses significant challenges due to the complexity and diversity of vocabulary used by radiologists. A more promising alternative is to leverage the capabilities of powerful LLMs like ChatGPT to tackle this task. Table \ref{table:fact-extraction-examples} presents several examples of facts extracted by GPT-4 from challenging sentences. This is achieved by providing the model with a specific set of instructions: (Refer to Figure \ref{figure:chatgpt-prompts:fact-extraction} for a screenshot of OpenAI's web interface displaying the same prompt.)

\begin{quote}
\small
Relevant facts:

1. observations of abnormalities
2. observations of diseases
3. observations of strange visual patterns
4. observations of devices
5. observations of foreign bodies
6. observations of specific anatomical regions that look normal or healthy
7. absences of abnormalities (usually expressed with a negation)
8. comparisons with respect to a previous study (something changed or remained the same)

Task:

Given a sentence taken from a chest x-ray report, generate a JSON list of relevant facts.
Each fact should be about one observation. If a sentence mentions multiple observations,
each observation should be extracted as a separate fact.
Each fact should include the anatomical location where it was observed. If multiple facts
occur in the same location, repeat the location in each fact.

If no relevant facts are mentioned, return [] (an empty array).

Examples:

Opacity and density in the right lobe

[
"opacity in the right lobe",
"density in the right lobe"
]

Lungs are well inflated without evidence of focal airspace consolidation to suggest pneumonia.

[
"well inflated lungs",
"lungs without evidence of focal airspace consolidation",
"lungs without evidence of pneumonia"
]

Taken together, compared with less than 1 hr earlier, the findings are suggestive of worsening of CHF, with new or significantly increased left greater right pleural effusions and underlying bibasilar collapse and/or  consolidation, particularly on the left.

[
"worsening of CHF",
"new or significantly increased left pleural effusions",
"new or significantly increased right pleural effusions",
"underlying bibasilar collapse on the left",
"underlying consolidation on the left",
]

No acute cardiopulmonary abnormality

[
"no acute cardiopulmonary abnormality"
]
\end{quote}

Given the relatively high cost of both GPT-4 and GPT-3.5 models, we opted to annotate a subset of the MIMIC-CXR sentences with these LLMs. To select this subset, we employed a two-pronged approach. Firstly, we ranked sentences based on the sum of the inverse frequency of their tokens, thus prioritizing longer and more complex sentences, which often include infrequent abnormalities, typos, and symbols. Secondly, to ensure diversity of topics, we clustered the sentences into 200 groups using K-Means, utilizing embedding representations obtained with BioViL-T, and sampled equally from each cluster in order of difficulty. This combined strategy allowed us to curate a subset that is both diverse and challenging. GPT-4-0613 extracted facts from 24,998 sentences, while GPT-3.5-turbo-0613 processed 69,936. Subsequently, we trained T5-small for fact extraction using a total of 94,934 training examples, reserving 200 examples for validation. Once trained, T5-small annotated the remaining 582,760 sentences, yielding a total of 1,341,830 facts, including duplicates, of which 591,920 were unique after duplicate removal.

\input{tables/fact-extraction-examples}

\subsection{\cxrfactencodershortname's Tasks Details}
\label{sec:appendix:tasks}

\cxrfactencodershortname\ is a fine-tuned version of CXR-BERT-specialized, accessible for download from \url{https://huggingface.co/microsoft/BiomedVLP-CXR-BERT-specialized}. This fine-tuning process entails multi-task learning, incorporating six distinct tasks: triplet loss for sentence ranking (T), sentence classification (C), sentence decoding (SD), natural language inference (NLI), quadruplet loss for enforcing separation between entailment and contradiction pairs (EC), and entity and relation extraction (ER). While there exist a total of 64 possible task combinations, we focus on 11 combinations in our experiments, namely: T, T+C, T+ER, T+SD, T+EC, T+NLI, T+EC+NLI, T+C+EC+NLI, T+C+EC+NLI+ER, T+C+EC+NLI+SD, T+C+EC+NLI+ER+SD. This selection is guided by the following heuristics:

\begin{itemize}
    \setlength\itemsep{-0.1em}
    \item Triplet loss (T) is always included, given our belief, as discussed in Appendix \ref{sec:appendix:triplets-dataset-details}, that the developed triplets dataset captures many desirable properties for learned sentence embeddings.
    \item We explore each combination of T with the other tasks individually: T+C, T+ER, T+SD, T+EC, T+NLI.
    \item Recognizing the complementary nature of EC and NLI in exploiting natural language inference data, we explore T+EC+NLI.
    \item According to our experimental results, both C and EC+NLI serve as effective auxiliary tasks; thus, we keep them fixed while varying other combinations of SD and ER: T+C+EC+NLI, T+C+EC+NLI+ER, T+C+EC+NLI+SD, T+C+EC+NLI+ER+SD.
\end{itemize}

Next, we delve into the implementation details of each task in our experiments.

\begin{figure}[!htb]
\begin{center}
\includegraphics[width=0.85\linewidth]{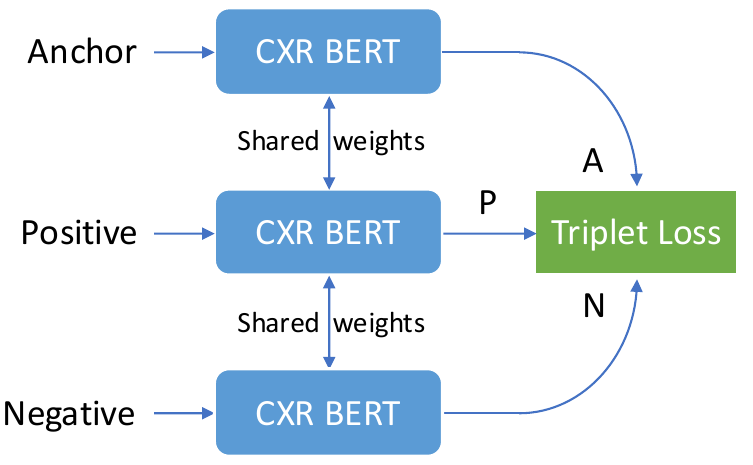}
\end{center}
\caption{Triplet loss (T)}
\label{figure:task-T}
\end{figure}

\textbf{Triplet loss for sentence ranking (T)}. One of the tasks we explore for model fine-tuning is sentence ranking via triplet loss. Figure \ref{figure:task-T} illustrates this task. Concretely, we forward 3 sentences (anchor, positive, negative) through CXR-BERT-specialized with weight sharing, obtaining three vectors \(a\), \(b\), and \(c\) each of dimension 128, and compute \(\Delta \text{sim}(a, p, n) =   a \cdot p - a \cdot n\). This is serves as the input logit for a binary cross-entropy loss.

We provide details on the construction of the triplets dataset used in this task in Section \ref{sec:appendix:triplets-dataset-details}.

\begin{figure}[!htb]
\begin{center}
\includegraphics[width=0.9\linewidth]{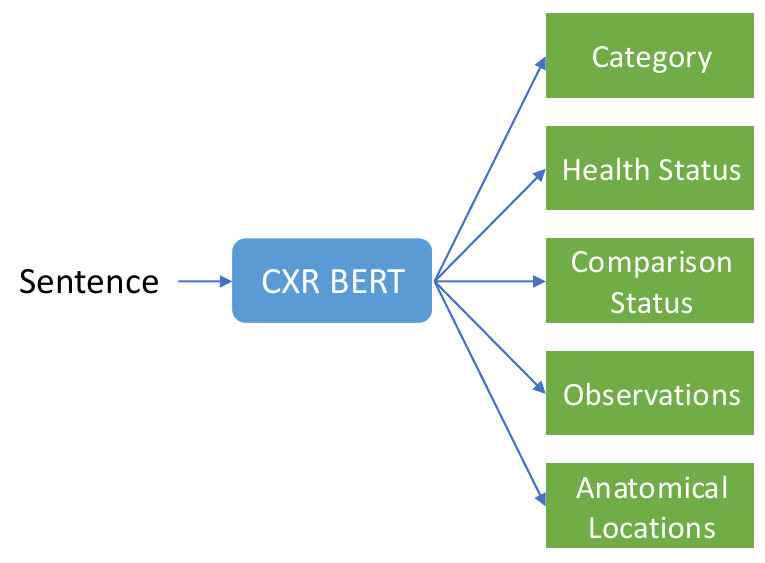}
\end{center}
\caption{Sentence classification (C)}
\label{figure:task-C}
\end{figure}

\begin{figure*}[!htb]
\begin{center}
\includegraphics[width=\textwidth]{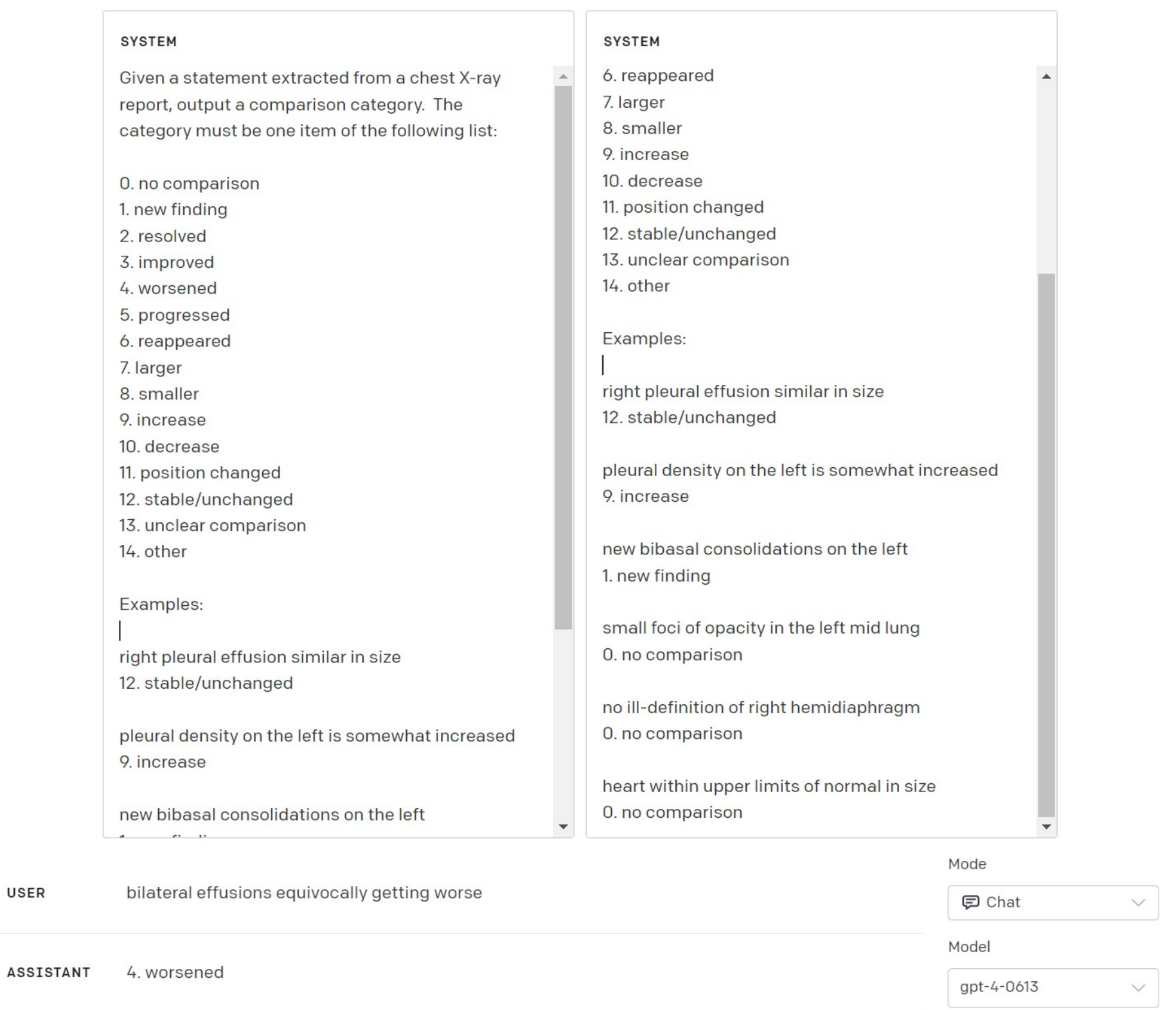}
\end{center}
\caption{ChatGPT prompt. Fact to comparison status}
\label{figure:chatgpt-prompts:fact2comparison}
\end{figure*}

\begin{figure*}[!htb]
\begin{center}
\includegraphics[width=\textwidth]{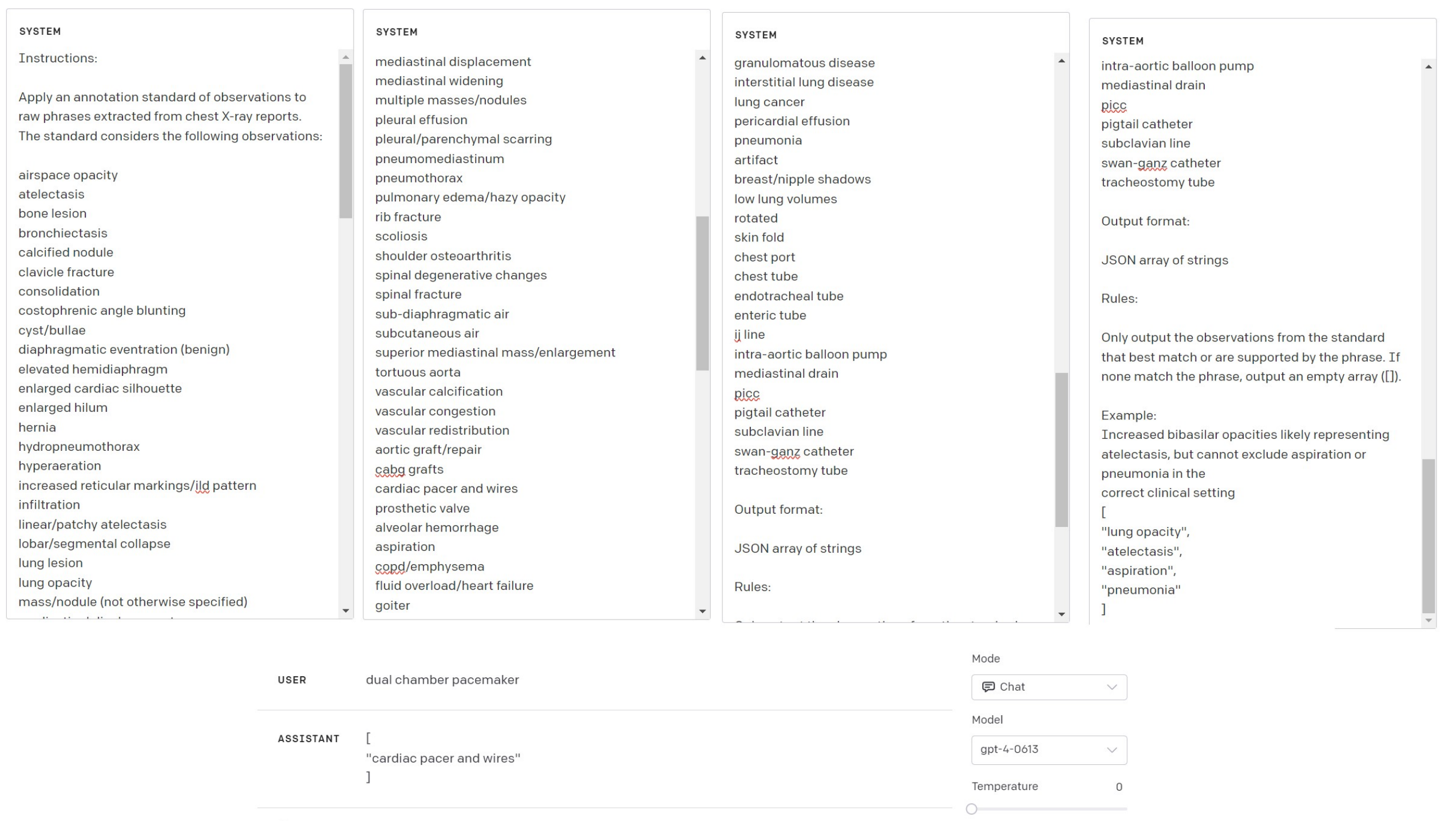}
\end{center}
\caption{ChatGPT prompt. Fact to Chest ImaGenome observations}
\label{figure:chatgpt-prompts:fact2obs}
\end{figure*}

\begin{figure*}[!htb]
\begin{center}
\includegraphics[width=\textwidth]{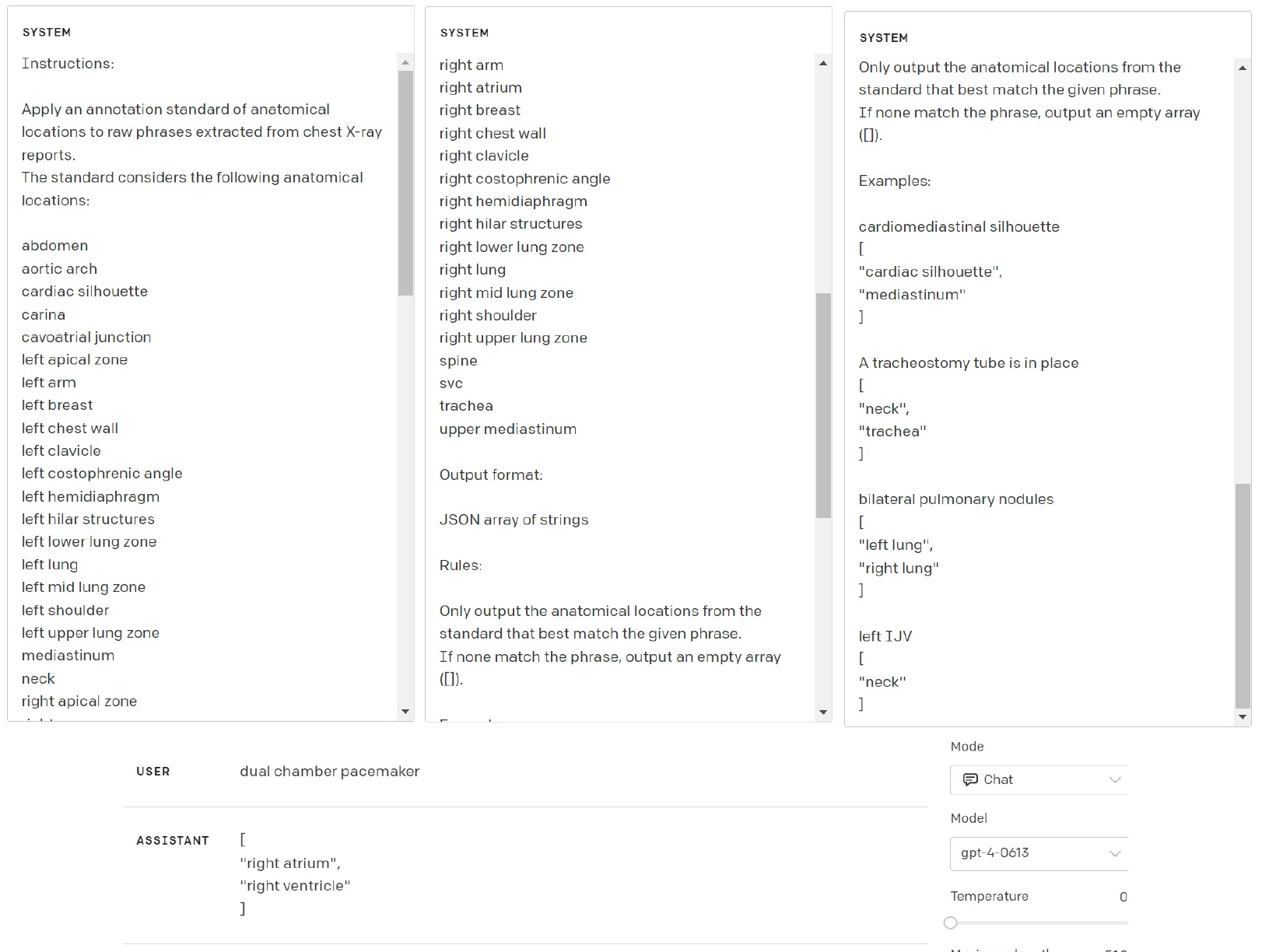}
\end{center}
\caption{ChatGPT prompt. Fact to Chest ImaGenome anatomical locations}
\label{figure:chatgpt-prompts:fact2anat}
\end{figure*}

\textbf{Sentence classification tasks (C)}. A second task group is classification tasks (Figure \ref{figure:task-C}). These include category (5 classes: \textit{anatomical finding}, \textit{disease}, \textit{technical assessment}, \textit{tubes and lines} and \textit{device}), health status (4 classes: \textit{normal}, \textit{abnormal}, \textit{ambiguous}, \textit{unknown}), comparison status (15 classes, see Figure \ref{figure:chatgpt-prompts:fact2comparison}), Chest ImaGenome observations (74 classes, see Figure \ref{figure:chatgpt-prompts:fact2obs}) and anatomical locations (38 classes, see Figure \ref{figure:chatgpt-prompts:fact2anat}). Category, Health Status and Comparison Status are single-label multi-class classification tasks, whereas Chest ImaGenome observations and anatomical locations are multi-label binary classification tasks. Implementing these tasks require attaching fully connected heads on top of CXR-BERT-specialized's built-in projection layer in order to perform the classification.

The data for the classification tasks comes from two primary sources. The first source is the Chest ImaGenome dataset, which provides observation and anatomical location labels in scene graphs, conveniently linked to specific sentences from MIMIC-CXR reports. The second source is ChatGPT, which we leverage to annotate the extracted facts with details such as category, health status, comparison status, observations, and anatomical locations. This process involves a teacher-student approach similar to our method for fact extraction: we use ChatGPT to annotate a subset of facts, incurring some monetary cost, with the prompts shown in Figures \ref{figure:chatgpt-prompts:fact2metadata}, \ref{figure:chatgpt-prompts:fact2comparison}, \ref{figure:chatgpt-prompts:fact2obs}, and \ref{figure:chatgpt-prompts:fact2anat}, and then fine-tune T5 to annotate the remaining facts at no additional cost. 


Concretely, from the Chest ImaGenome scene graphs, we retrieved 556,111 sentences, each annotated with observations (74 classes) and anatomical locations (38 classes). In addition, facts were annotated as follows:
\begin{itemize}[topsep=0pt, partopsep=0pt, itemsep=0pt, parsep=0pt]
    \item 5,000 facts were annotated with observations by \texttt{GPT-4-0613}.
    \item 84,708 facts were annotated with observations by \texttt{GPT-3.5-turbo-0613}.
    \item 2,816,982 facts were annotated with observations by \texttt{T5-small} fine-tuned with the teacher-student approach. These include facts extracted from reports plus additional facts obtained via paraphrases. (For more details on paraphrases, see Section \ref{sec:appendix:triplets-dataset-details} on the construction of the triplets dataset.)
\end{itemize}

Similarly, anatomical locations were annotated as follows:
\begin{itemize}[topsep=0pt, partopsep=0pt, itemsep=0pt, parsep=0pt]
    \item 72,400 facts were annotated with anatomical locations by \texttt{GPT-3.5-turbo-0613}.
    \item 2,598,778 facts were annotated with anatomical locations by \texttt{T5-small} fine-tuned.
\end{itemize}

For other classification tasks (Category, Health Status, Comparison Status), we generated a JSON object with metadata from each fact using the prompt shown in Figure \ref{figure:chatgpt-prompts:fact2metadata}:
\begin{itemize}[topsep=0pt, partopsep=0pt, itemsep=0pt, parsep=0pt]
    \item 59,921 facts were annotated by \texttt{GPT-3.5-turbo-0613}.
    \item 535,959 facts were annotated by \texttt{T5-small} fine-tuned with the teacher-student approach.
\end{itemize}

\begin{figure}[!htb]
\begin{center}
\includegraphics[width=\linewidth]{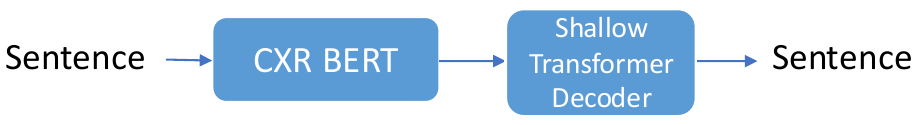}
\end{center}
\caption{Sentence decoding (SD)}
\label{figure:task-SD}
\end{figure}

\textbf{Sentence decoding (SD)}. Another task is sentence decoding (Figure \ref{figure:task-SD}). We attach a lightweight, shallow Transformer Decoder to CXR-BERT-specialized's projection layer in order to generate back the original sentence. This can be viewed a sort of text autoenconder, forcing the projection layer to capture as much information as possible of the input sentence to facilitate the reconstruction of the sentence by the Transformer Decoder. We use a Transfomer Decoder with embedding, hidden and feedforward dimension 256, only one self-attention head and only one layer.

Because this task relies on self-supervision and does not require specialized annotations, any sentence or fact can serve as a training instance.

\begin{figure}[!htb]
\begin{center}
\includegraphics[width=\linewidth]{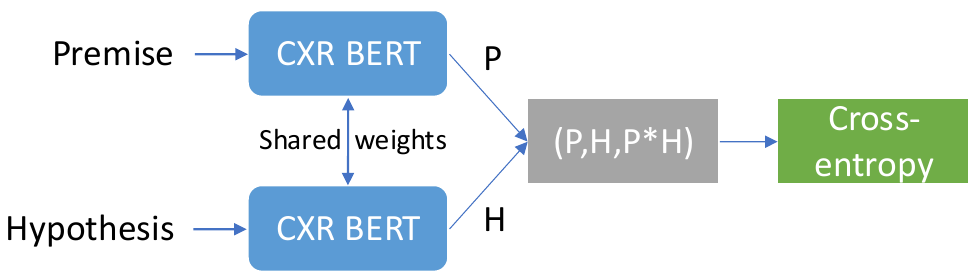}
\end{center}
\caption{Natural language inference (NLI)}
\label{figure:task-NLI}
\end{figure}

\textbf{Natural language inference (NLI)}. For NLI, we adopt an approach similar to that of SBERT \cite{reimers-gurevych-2019-sentence}, by concatenating the embeddings of the premise, hypothesis and their element-wise multiplication, followed by a fully connected layer and a softmax layer for NLI classification (see Figure \ref{figure:task-NLI}).

\begin{figure*}[!htb]
\begin{center}
\includegraphics[width=\textwidth]{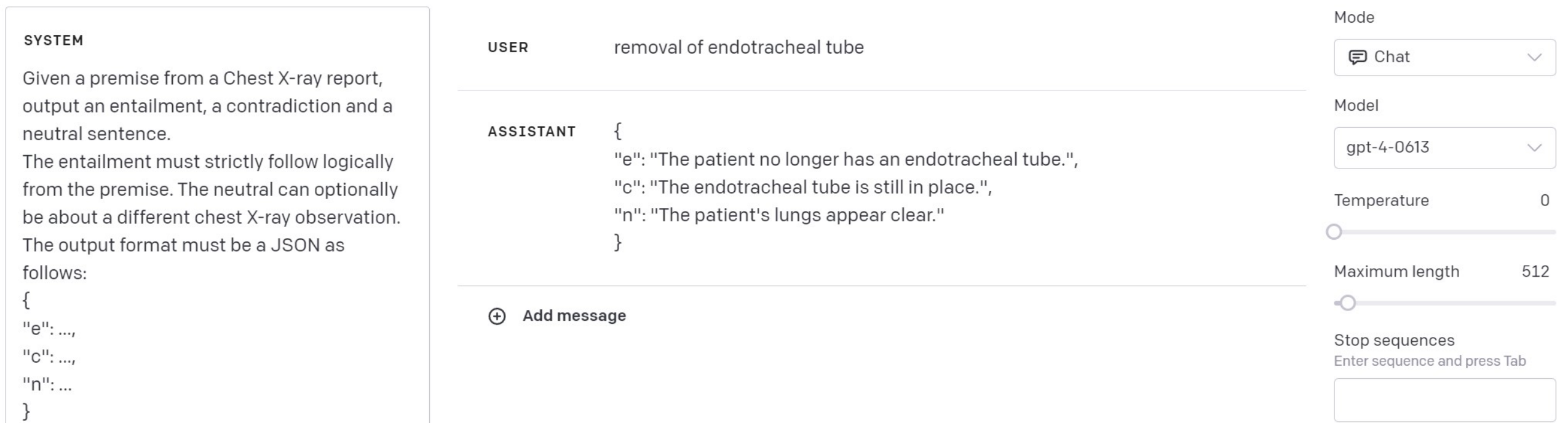}
\end{center}
\caption{NLI Prompt 1. Given a premise, output entailment, contradiction, and neutral hypotheses.}
\label{figure:chatgpt-prompts:nli:p2ecn}
\end{figure*}

\begin{figure*}[!htb]
\begin{center}
\includegraphics[width=\textwidth]{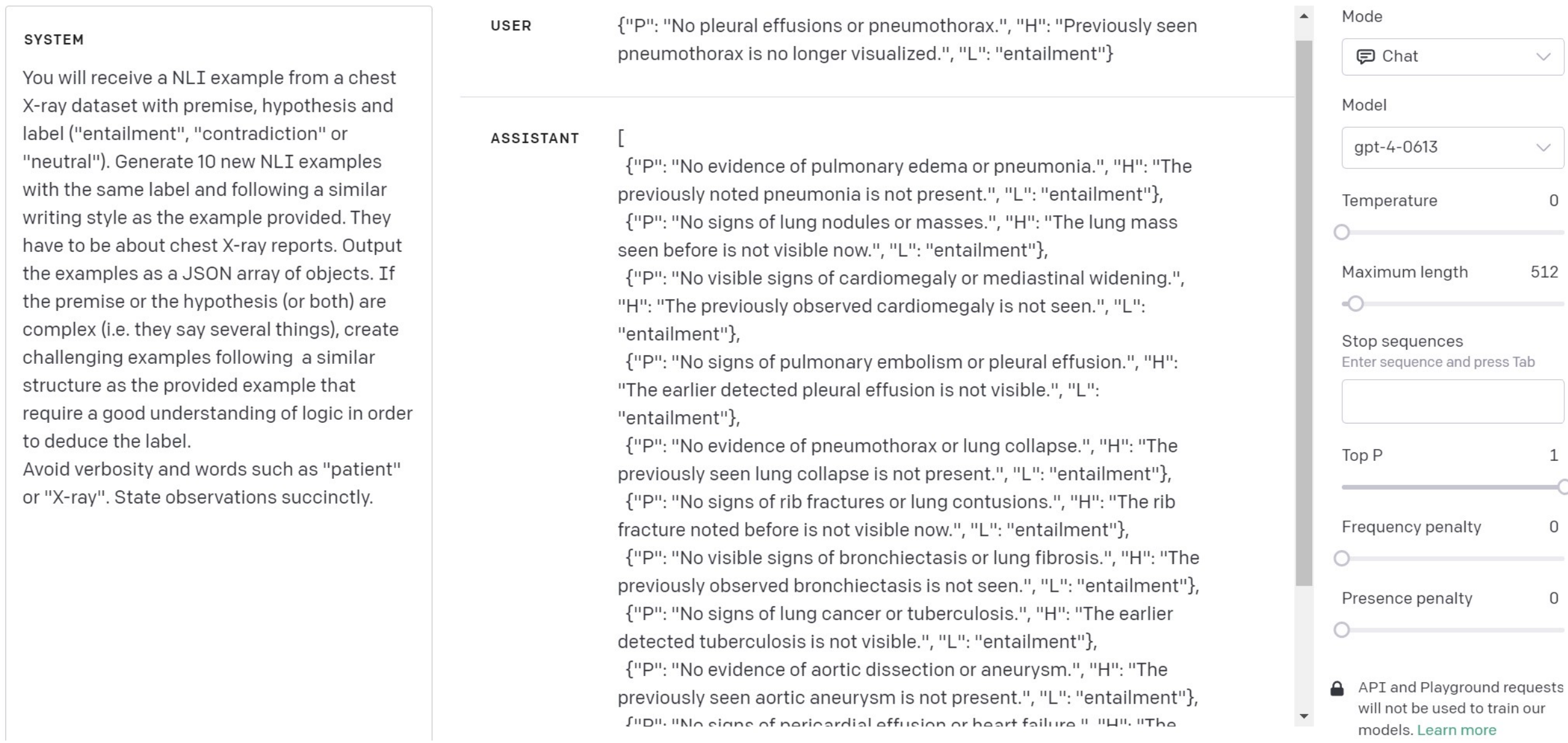}
\end{center}
\caption{NLI Prompt 2. Given a ground-truth NLI example, generate multiple similar  examples.}
\label{figure:chatgpt-prompts:nli:ex2sim}
\end{figure*}

\begin{figure*}[!htb]
\begin{center}
\includegraphics[width=\textwidth]{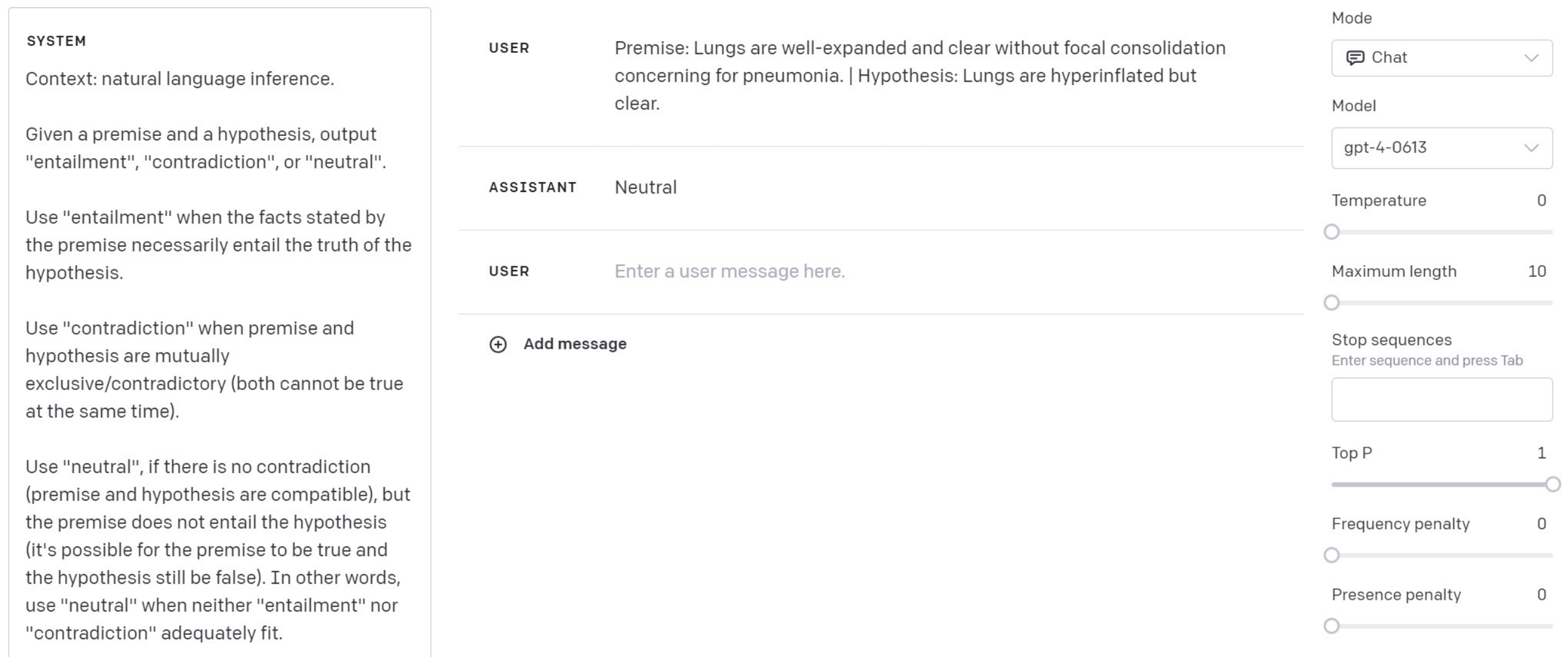}
\end{center}
\caption{NLI Prompt 3. Given a premise and a hypothesis, output the correct label. In this version, GPT-4 has to output the label immediately.}
\label{figure:chatgpt-prompts:nli:nli-simple}
\end{figure*}

\begin{figure*}[!htb]
\begin{center}
\includegraphics[width=\textwidth]{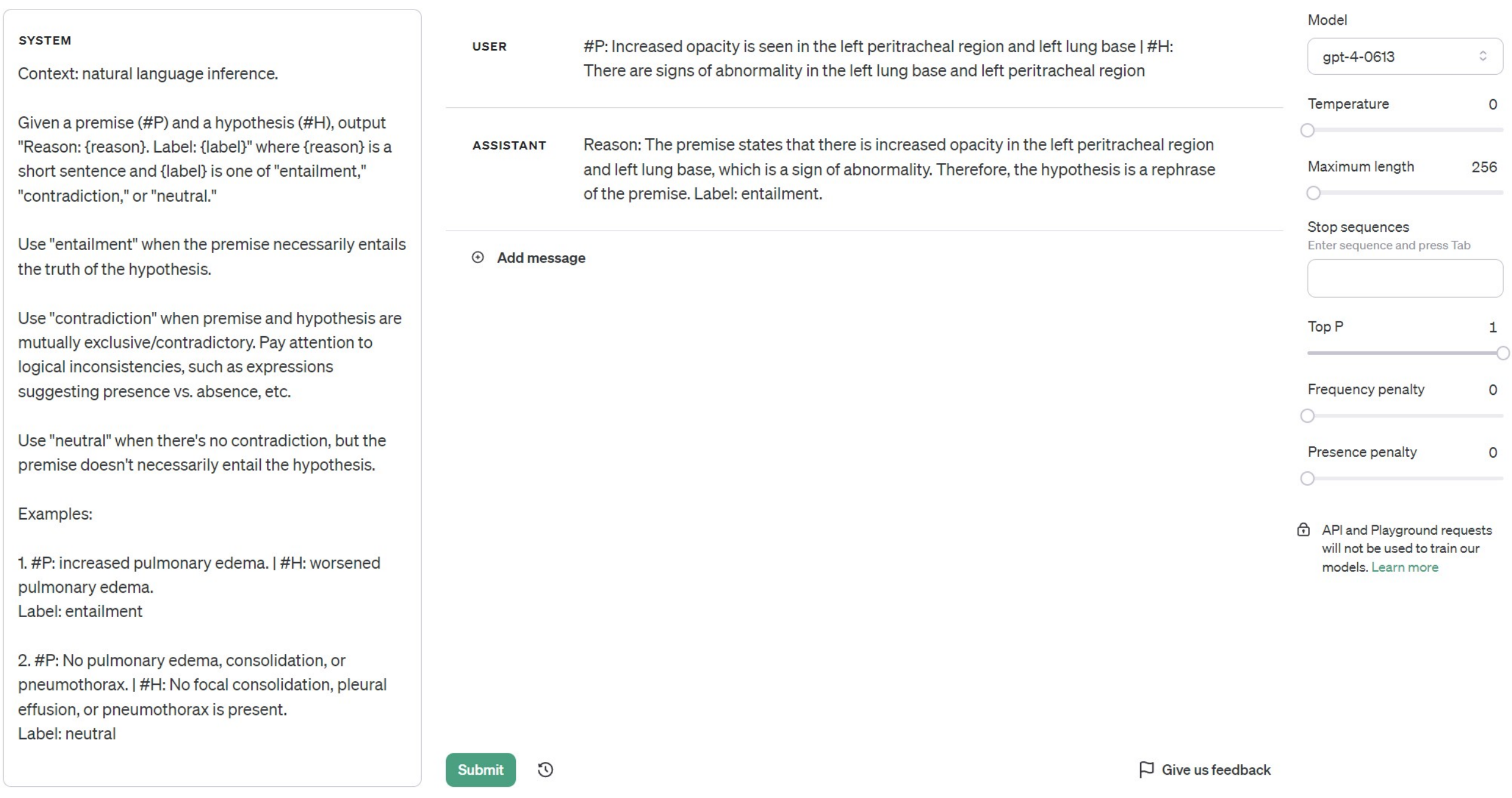}
\end{center}
\caption{NLI Prompt 4. Given a premise and a hypothesis, output the correct label. In this version, GPT-4 is given the chance to reason first and then output the label. It is also shown a couple of examples with the expected labels. These simple changes to the prompt significantly boosted GPT-4's performance.}
\label{figure:chatgpt-prompts:nli:nli-cot-ex}
\end{figure*}

\begin{figure*}[!htb]
\begin{center}
\includegraphics[width=\textwidth]{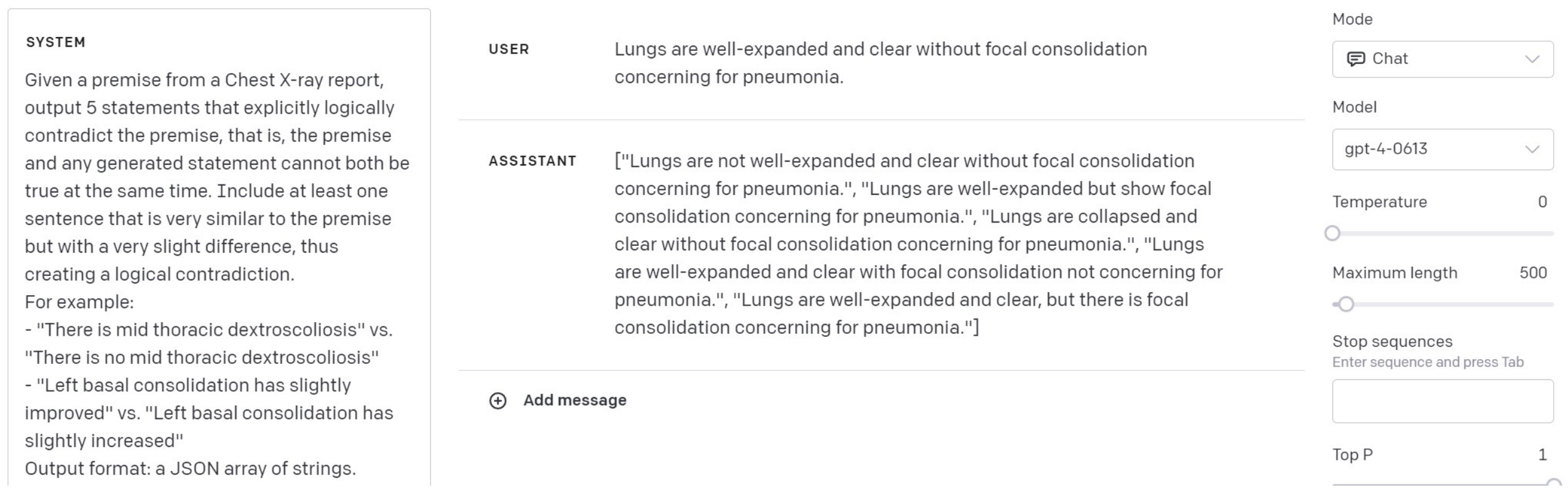}
\end{center}
\caption{NLI Prompt 5. Given a premise, generate diverse hypotheses that contradict it.}
\label{figure:chatgpt-prompts:nli:p2cont}
\end{figure*}

As discussed in Section \ref{sec:datasets:nli}, our NLI dataset was compiled by merging data from MedNLI (14,049 pairs), the RadNLI development set (480 pairs), and examples generated by GPT-4 (154,498 pairs), resulting in a comprehensive collection of 169,025 pairs. These consist of 25,175 entailment pairs, 44,729 neutral pairs, and 99,121 contradiction pairs. Additionally, the RadNLI test set (480 pairs) and MS-CXR-T (361 pairs) were reserved for evaluation purposes. To generate NLI examples using GPT-4, we employed four distinct prompts:


\begin{itemize}[topsep=0pt, partopsep=0pt, itemsep=0pt, parsep=0pt]
    \item A prompt that generates entailment, neutral, and contradiction sentences from a reference sentence (Figure \ref{figure:chatgpt-prompts:nli:p2ecn}).
    \item A prompt that aims to produce examples analogous to a given NLI reference example (Figure \ref{figure:chatgpt-prompts:nli:ex2sim}).
    \item A prompt that predicts the correct label for a given premise and hypothesis, incorporating Chain-of-Thought (CoT) reasoning and examples (Figure \ref{figure:chatgpt-prompts:nli:nli-cot-ex}). A simpler version of this prompt, which only requests the correct label, was also considered (Figure \ref{figure:chatgpt-prompts:nli:nli-simple}). However, as indicated in Table \ref{table:radnli-results}, CoT is crucial for achieving significantly more accurate predictions.
    \item A prompt designed to generate contradictory sentences relative to a reference sentence (Figure \ref{figure:chatgpt-prompts:nli:p2cont}).
\end{itemize}

\begin{figure}[!htb]
\begin{center}
\includegraphics[width=\linewidth]{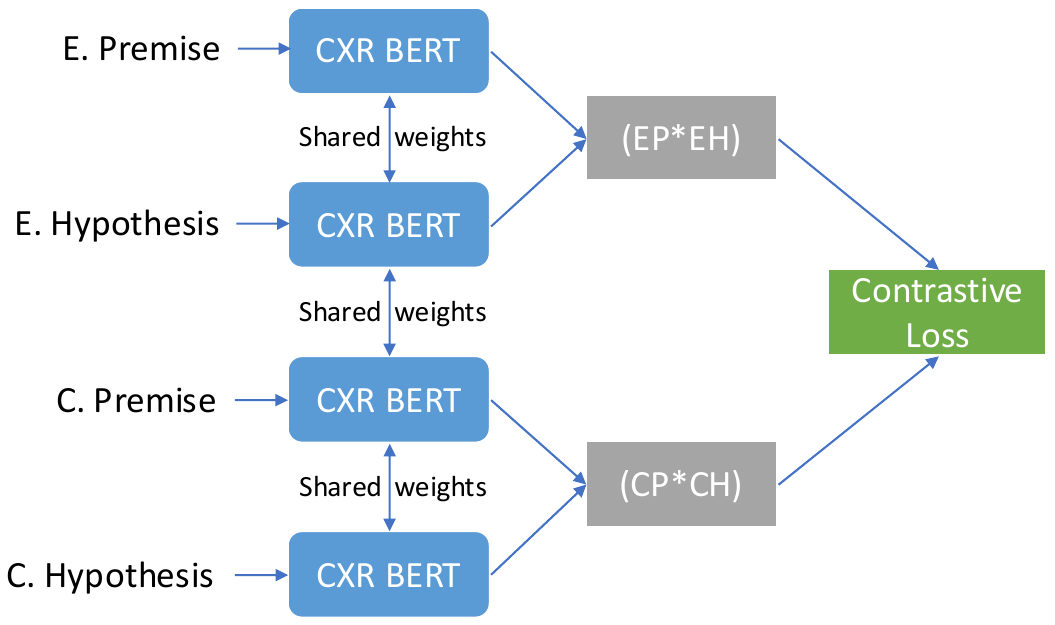}
\end{center}
\caption{Entailment/contradiction quadruplet loss (EC)}
\label{figure:task-EC}
\end{figure}

\textbf{Quadruplet loss: enforcing separation between entailment and contradiction pairs (EC)}. The next task is what we refer to as entailment/contradiction quadruplet loss (Figure \ref{figure:task-EC}). The goal of this task is to promote a generalized separation of entailment and contradiction sentence pairs in the latent space, by randomly sampling entailment and contradiction pairs and requiring that the entailment pair have greater similarity than the contradiction pair. This loss was a key contributor to achieving the state-of-the-art results presented in Table \ref{table:nli-entcont-results}.

Since this task complements the standard NLI task, it utilizes the same dataset but excludes neutral pairs, focusing solely on entailment and contradiction pairs.

\begin{figure}[!htb]
\begin{center}
\includegraphics[width=0.95
\linewidth]{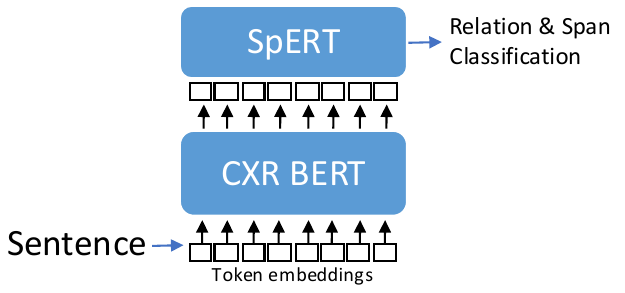}
\end{center}
\caption{Entity and relation extraction (ER) with SpERT}
\label{figure:task-ER}
\end{figure}

\textbf{Entity and relation extraction (ER)}. Lastly, for entity and relation extraction we augment CXR-BERT-specialized with the layers proposed by SpERT \cite{DBLP:conf/ecai/EbertsU20}, as illustrated in Figure \ref{figure:task-ER}. This adaptation was relatively straightforward, since the authors of SpERT released an implementation (\url{https://github.com/lavis-nlp/spert/}) that is compatible with Huggingface models like CXR-BERT-specialized.

To implement this task, we utilize the gold data provided by the RadGraph dataset \cite{radgraph}. This dataset comprises 500 MIMIC-CXR radiology reports, annotated with an entity-and-relation schema by board-certified radiologists. Additionally, a test set containing 50 MIMIC-CXR and 50 CheXpert reports, annotated in the same manner, is included. We use all of this data for training.

\input{tables/template-based-reports}

\input{tables/template-based-reports-2}



\subsection{Triplets Dataset Construction Details}
\label{sec:appendix:triplets-dataset-details}

\cxrfactencodershortname\ is trained to generate sentence embeddings that organize semantically similar sentences into clusters within the embedding space via a triplet ranking task. This task involves a dataset comprising triplets ($a$, $p$, $n$), where $a$ represents an anchor sentence, $p$ is a positive sentence, and $n$ denotes a negative sentence. The objective is to train a text encoder in a manner that ensures \(\text{sim}(a, p) > \text{sim}(a, n)\) holds true for each triplet, with \(\text{sim}(x, y)\) computed as the cosine similarity between sentence embeddings (or dot product if they are already normalized).

The rationale behind selecting triplet loss lies in its versatility, offering flexibility in terms of desired properties for the learned embedding function. For instance, if $a$ and $p$ denote identical medical observations using different vocabulary, while $n$ presents an unrelated observation, it's logical to enforce \(\text{sim}(a, p) > \text{sim}(a, n)\). Similarly, if there's substantial topic overlap between $a$ and $p$, whereas $n$ is largely unrelated or contradictory to $a$, it's reasonable to aim for \(\text{sim}(a, p) > \text{sim}(a, n)\) to hold true.

To guide a text encoder towards learning sentence embeddings consistent with triplets, we can define the difference in similarities as \(\Delta \text{sim}(a, p, n) = \text{sim}(a, p) - \text{sim}(a, n)\). Then, a straightforward approach is to utilize \(\text{sim}(a, p, n)\) as the input for a binary cross-entropy loss, where the ground-truth label is consistently set to 1.

\textbf{Notation}. To define our triplet sampling heuristics, we use the notation E(\(x\)) for the embedding of sentence \(x\), cos(E(\(x\)), E(\(y\))) for the cosine similarity between embeddings of \(x\) and \(y\), lev(\(x\), \(y\)) for the Levenshtein string distance between them, and levsim(\(x\), \(y\)) = 1 - lev(\(x\), \(y\)) / max(len(\(x\)), len(\(y\))). c(\(x\)) indicates the cluster id for sentence \(x\) after running a clustering algorithm like K-Means on the sentence embeddings. This requires having an auxiliary text encoder capable of producing these auxiliary embeddings and clusters. Specifically, we use BioViL-T \cite{bannur2023learning}, a state-of-the-art BERT-based model for radiological text, available on Huggingface\footnote{{\url{https://huggingface.co/microsoft/BiomedVLP-BioViL-T}}}.

\begin{figure*}[!htb]
\begin{center}
\includegraphics[width=\textwidth]{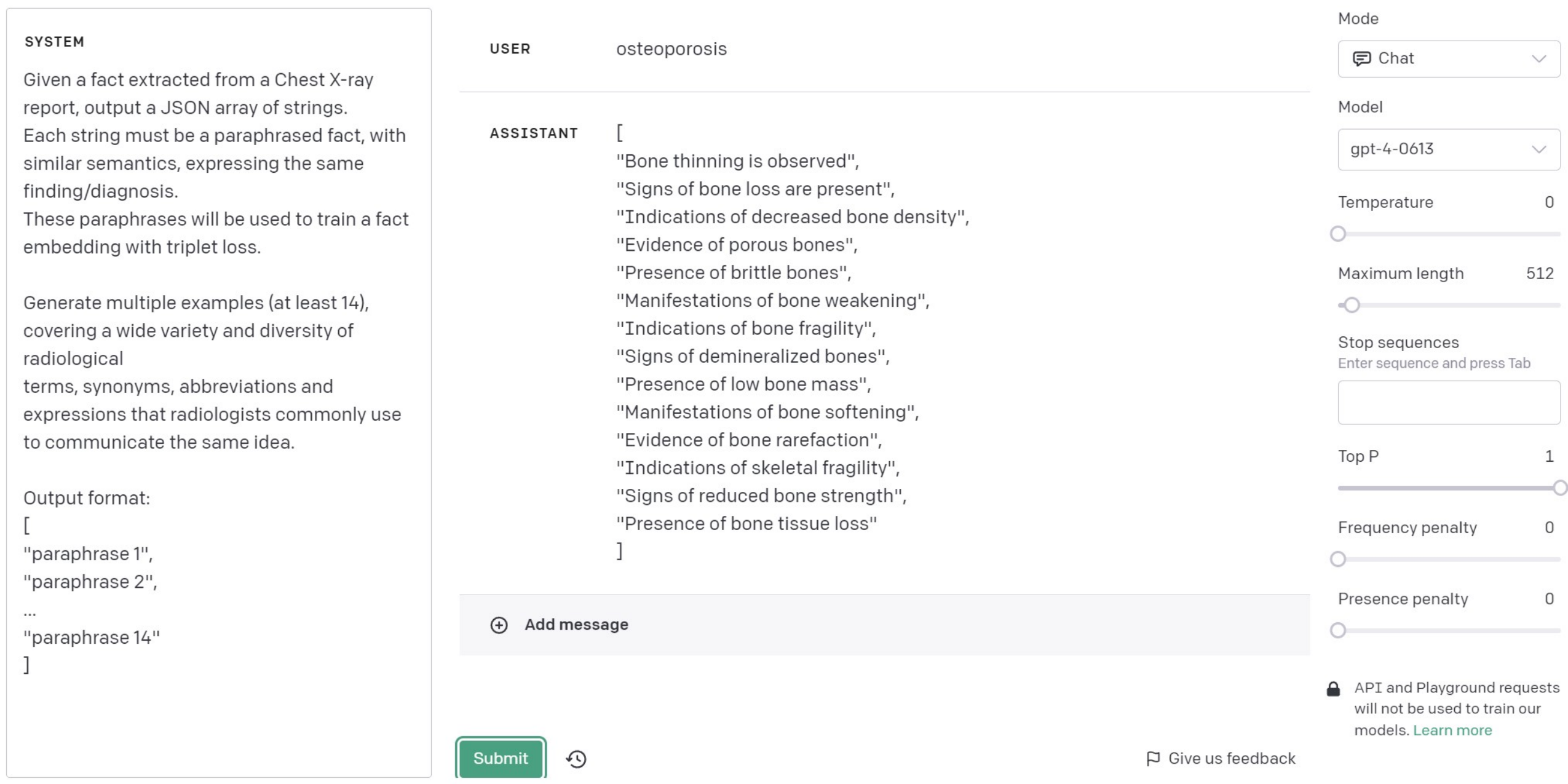}
\end{center}
\caption{ChatGPT prompt. Fact to paraphrases}
\label{figure:chatgpt-prompts:fact2paraphrases}
\end{figure*}

\begin{figure*}[!htb]
\begin{center}
\includegraphics[width=\textwidth]{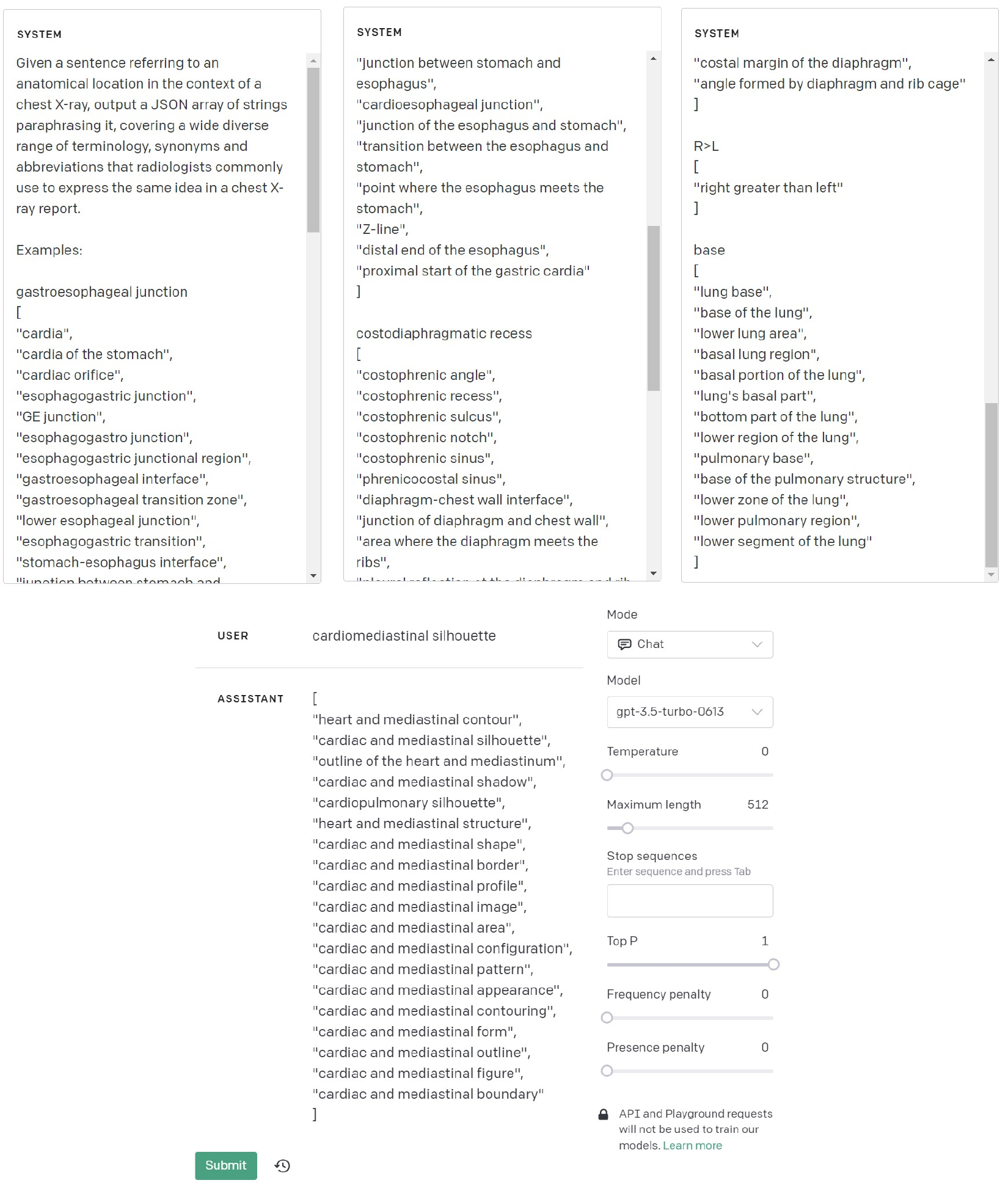}
\end{center}
\caption{ChatGPT prompt. Anatomy location to paraphrases}
\label{figure:chatgpt-prompts:anatomy2paraphrases}
\end{figure*}

\begin{figure*}[!htb]
\begin{center}
\includegraphics[width=\textwidth]{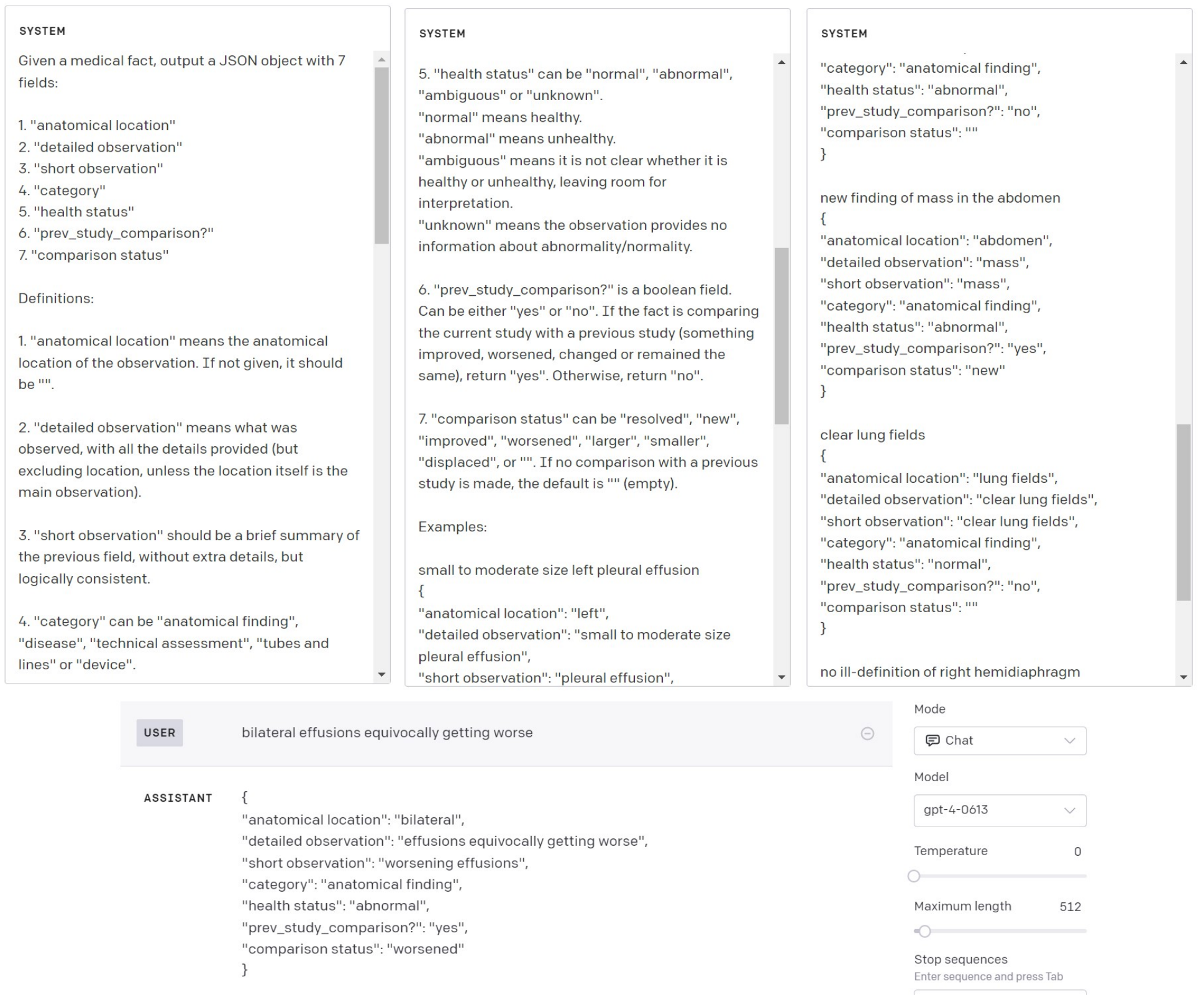}
\end{center}
\caption{ChatGPT prompt. Fact to metadata}
\label{figure:chatgpt-prompts:fact2metadata}
\end{figure*}

With this, we sample triplets based on the following heuristics:

\textbf{Rule 1: Rank paraphrases highly.} $\Delta \text{sim}(a, p, n) > 0$ if \(p\) is a paraphrase of \(a\) generated by ChatGPT and \(n\) is any other sentence (unless cos(E(\(a\)), E(\(p\))) $<$ cos(E(\(a\)), E(\(n\))) and lev(\(a\), \(p\)) $>$ lev(\(a\), \(n\))). To generate paraphrases, we employ the prompts shown in Figures \ref{figure:chatgpt-prompts:fact2paraphrases} and \ref{figure:chatgpt-prompts:anatomy2paraphrases}, one for paraphrasing facts and another for paraphrasing anatomical locations. We decided to paraphrase anatomical locations too in order to strengthen the model's understanding of their vocabulary. As a reminder, the anatomical locations are obtained from facts as part of the metadata generated with the prompt shown in Figure \ref{figure:chatgpt-prompts:fact2metadata}.

\textbf{Rule 2: Sample triplets according to the consensus of E and lev, while anchor and positive share the same health status.} $\Delta \text{sim}(a, p, n) > 0$ if HS(\(a\)) = HS(\(p\)), c(\(p\)) = c(\(a\)),  c(\(p\)) $\neq$ c(\(n\)), cos(E(\(a\)), E(\(p\))) $>$ cos(E(\(a\)), E(\(n\))) + margin$_{\text{cos}}$ and levsim(\(a\), \(p\)) $>$ levsim(\(a\), \(n\)) + margin$_{\text{lev}}$. This rule is very heuristic, and can be interpreted as a sort of mixture of experts. Essentially, we are looking for multiple signals to agree, increasing the likelihood that the resulting triplet passing those tests is in fact a high quality triplet. Levenshtein distance is useful to capture the notion of string similarity, while BioViL-T captures more the semantics (to some extent). The health status condition is adding an additional constraint. As a reminder, the health status is also obtained with the prompt of Figure \ref{figure:chatgpt-prompts:fact2metadata}.

\textbf{Rule 3: Short observation, detailed observation and the original fact (and their paraphrases) should be close to each other.} Given a fact \(f\), $\Delta \text{sim}(a, p, n) > 0$ if \(a\) and \(p\) $\in$ S(\(f\)), \(n\) $\notin$ S(\(f\)) and c(\(a\)) $\neq$ c(\(n\)) (unless cos(E(\(a\)), E(\(p\))) $<$ cos(E(\(a\)), E(\(n\))) and lev(\(a\), \(p\)) $>$ lev(\(a\), \(n\))). Here, S(\(f\)) stands for the union of \(f\), its detailed observation, its short observation and all the paraphrases (if any) generated for all of them with ChatGPT. The intuition here is that they are all closely related semantically, as they are all derived from the same fact (see Figure \ref{figure:chatgpt-prompts:fact2metadata}).

\textbf{Rule 4: Sample triplets according to Chest ImaGenome labels.} $\Delta \text{sim}(a, p, n) > 0$ if CIGL(\(a\)) $\cap$ CIGL(\(p\)) $\neq \emptyset$, CIGL(\(a\)) $\cap$ CIGL(\(n\)) = $\emptyset$, CIGL(\(p\)) $\cap$ CIGL(\(n\)) = $\emptyset$, and if (cos(E(\(a\)), E(\(p\))) $>$ cos(E(\(a\)), E(\(n\))) AND levsim(\(a\), \(p\)) $>$ levsim(\(a\), \(n\))). Here, CIGL(\(x\)) stands for the set of Chest ImaGenome labels of the sentence \(x\). With this rule, our aim is to heuristically utilize the Chest ImaGenome labels to identify semantically similar sentences that warrant clustering together. However, we enhance this approach by incorporating signals from Levenshtein distance and BioViL-T to bolster our confidence in the triplet quality.

\textbf{Rule 5: Rank triplets according to the overlap of entities and relations from RadGraph.} $\Delta \text{sim}(a, p, n) > 0$ if c(\(a\)) = c(\(p\)), c(\(a\)) $\neq$ c(\(n\)), and J(RG(\(a\)), RG(\(p\))) $>$ J(RG(\(a\)), RG(\(n\))) + margin$_{\text{RG}}$. Here, RG(\(x\)) stands for the set of RadGraph entities and relations for the sentence \(x\), and J for Jaccard similarity. Following the same idea of the previous rule, we seek to utilize the entities and relations provided by the RadGraph dataset as valuable cues for identifying semantically similar sentences that ought to be clustered together in the embedding space.

\begin{figure*}[!htb]
\begin{center}
\includegraphics[width=\textwidth]{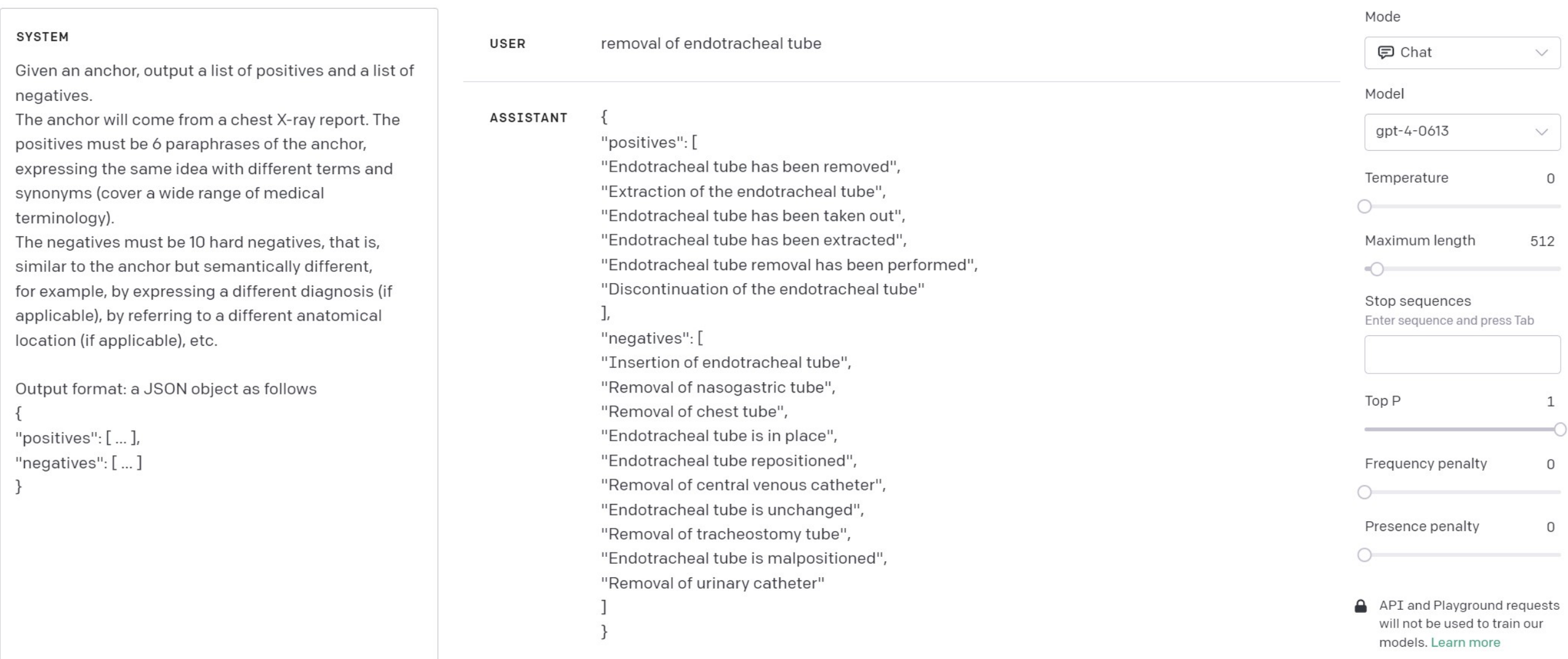}
\end{center}
\caption{ChatGPT prompt. Hard triplets}
\label{figure:chatgpt-prompts:hard-triplets}
\end{figure*}

\textbf{Rule 6: Hard triplets generated by ChatGPT.} $\Delta \text{sim}(a, p, n) > 0$ if (\(a\), \(p\), \(n\)) is a hard triplet generated by ChatGPT. The intuition behind this rule is very simple: we want to leverage ChatGPT's remarkable skills to produce challenging triplets, requiring a good understanding of the text to be ranked correctly. Figure \ref{figure:chatgpt-prompts:hard-triplets} shows the prompt used to generate these triplets along with an example.

For each rule, we sample approximately 3 to 4 million training triplets, along with 1,000 triplets for validation and 1,000 triplets for testing.

\begin{figure*}[!htb]
\begin{center}
\includegraphics[width=0.85\linewidth]{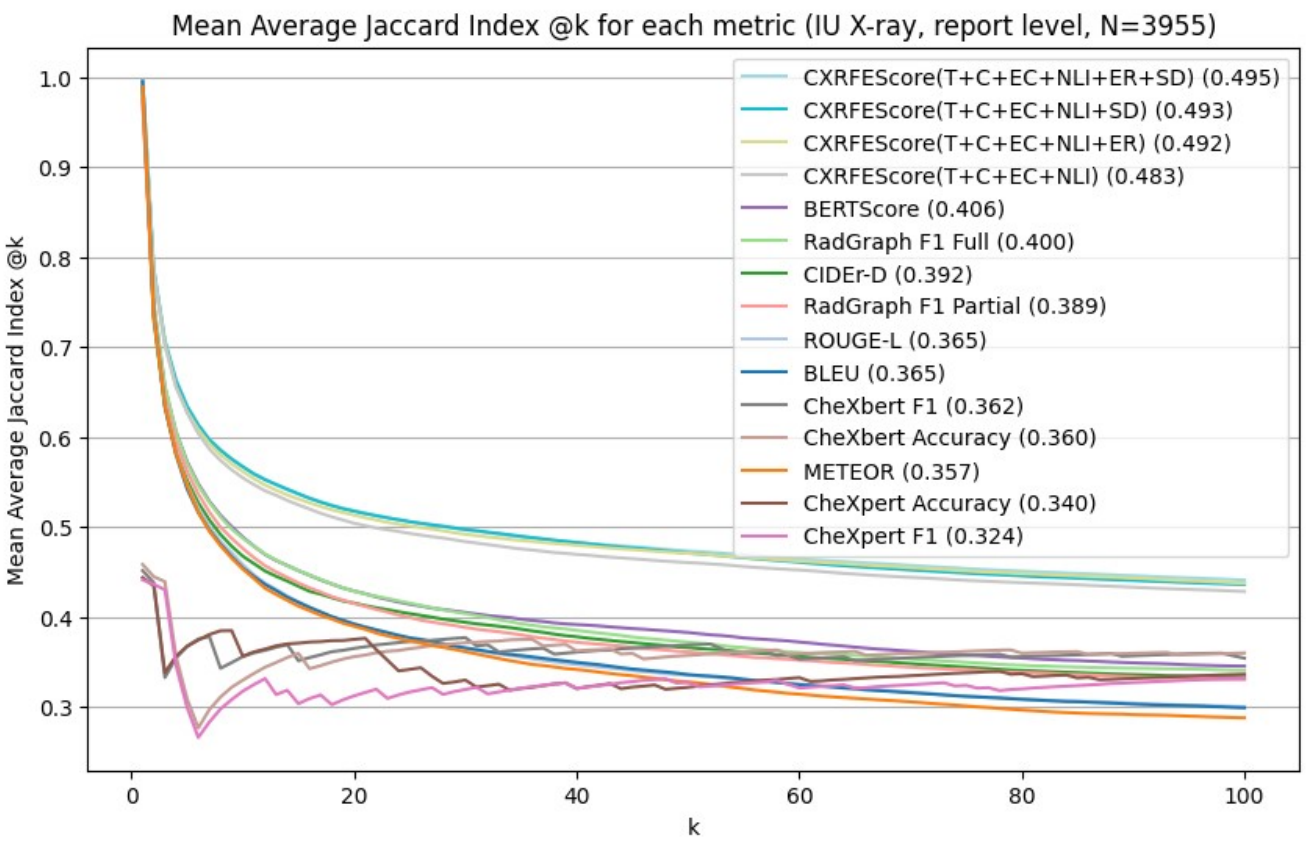}
\end{center}
\caption{Mean average Jaccard Index at k, for the 3955 reports of the IU X-ray dataset. Larger Jaccard Index is better. The Jaccard Index is calculated by comparing bags of words obtained from the manual and automatic tags associated with each report in the IU X-ray dataset.}
\label{figure:mean_avg_jaccard_report_iuxray}
\end{figure*}

\begin{figure*}[!htb]
\begin{center}
\includegraphics[width=0.85\linewidth]{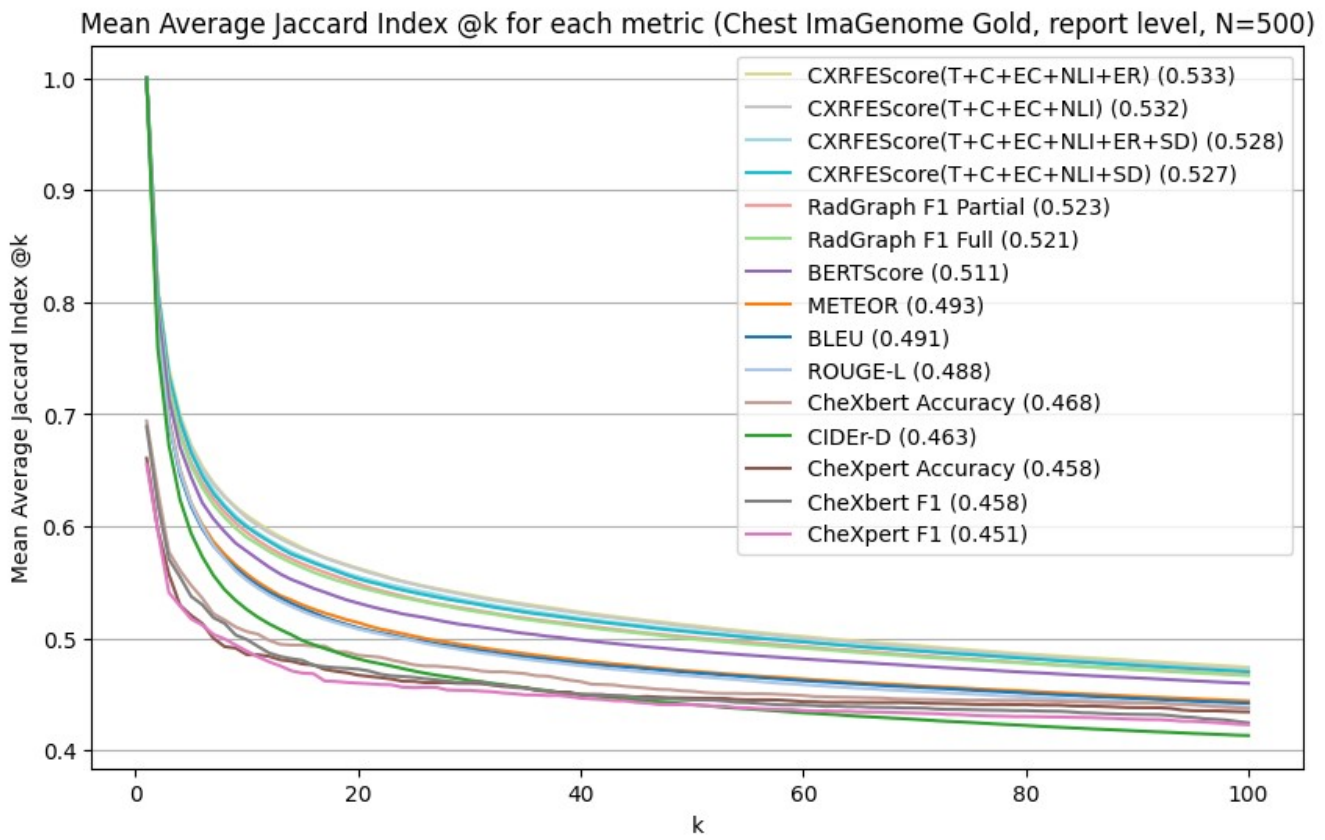}
\end{center}
\caption{Mean average Jaccard Index at k, for the 500 reports in the gold dataset of Chest ImaGenome. Larger Jaccard Index is better.}
\label{figure:mean_avg_jaccard_report_chestimg}
\end{figure*}

\begin{figure*}[!htb]
\begin{center}
\includegraphics[width=0.85\linewidth]{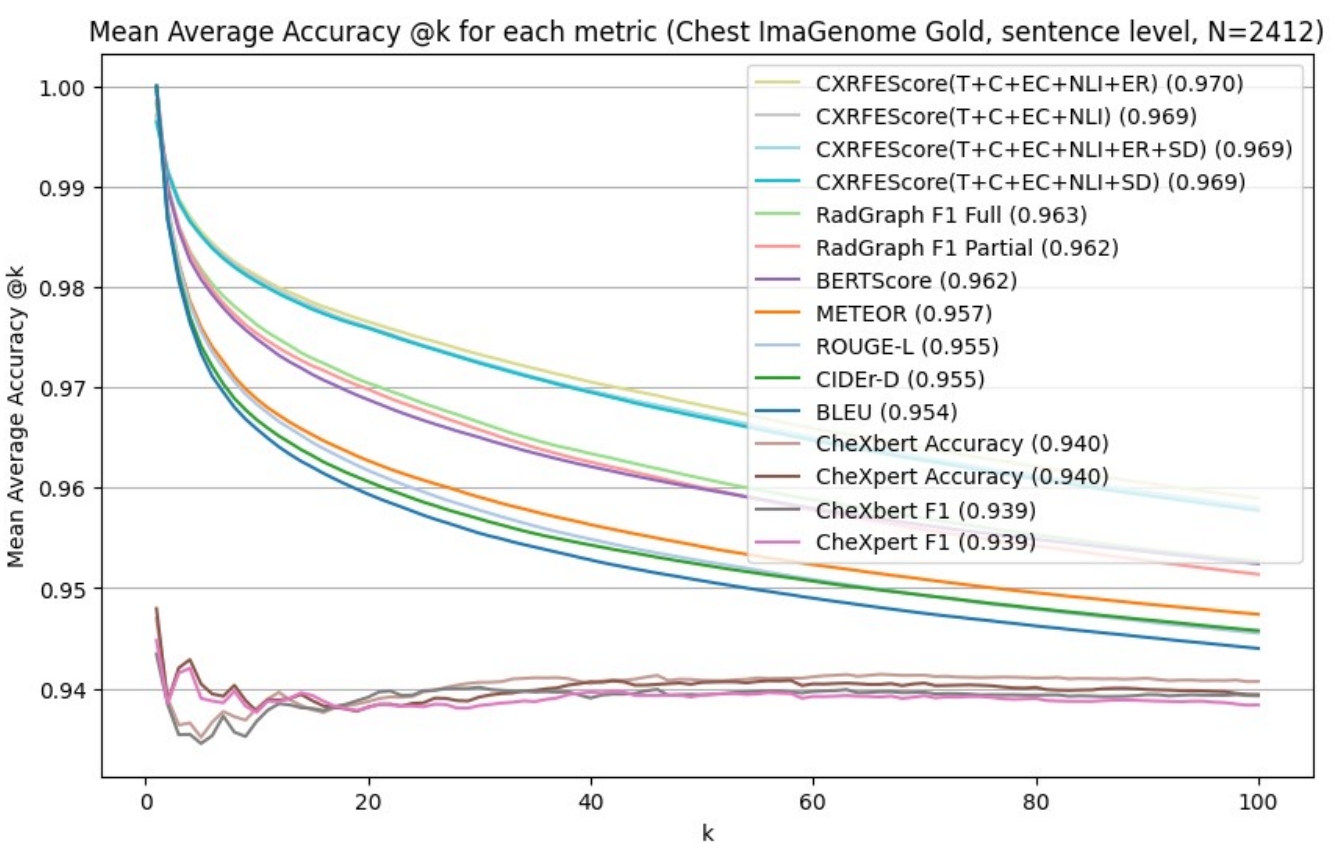}
\end{center}
\caption{Mean average accuracy at k, for 2412 sentences in the gold dataset of Chest ImaGenome. Larger accuracy is better.}
\label{figure:mean_avg_acc_sentences_chestimg}
\end{figure*}

\begin{figure*}[!htb]
\begin{center}
\includegraphics[width=0.9\linewidth]{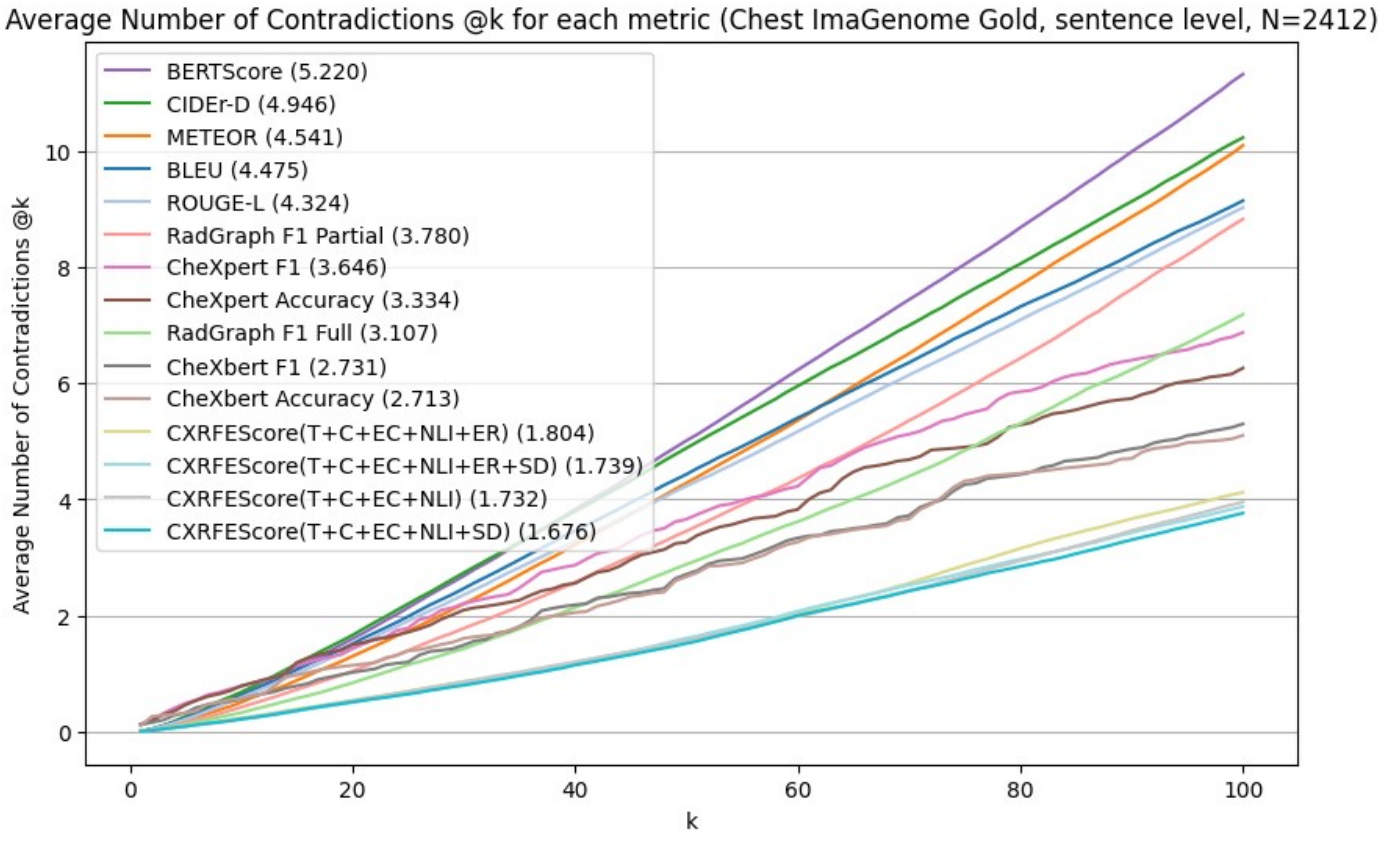}
\end{center}
\caption{Average number of contradictions at k, for 2412 sentences in the gold dataset of Chest ImaGenome. Smaller number of contradictions is better.}
\label{figure:avg_num_contr_sentences_chestimg}
\end{figure*}

\subsection{Additional Metric Evaluation Details and Results}
\label{sec:appendix:metrics-details}

We conduct a thorough comparison involving \cxrfactencodermetricname\ against several established metrics, including BLEU \cite{papineni2002bleu}, ROUGE-L \cite{lin-2004-rouge}, METEOR \cite{banerjee2005meteor}, CIDEr-D \cite{vedantam2015cider}, BERTScore \cite{bert-score}, CheXpert labeler \cite{irvin2019chexpert}, CheXbert \cite{chexbert}, and RadGraph F1 \cite{delbrouck-etal-2022-improving}. The last three are considered domain-specific metrics, tailored to the radiology domain, while the others serve as general-purpose evaluation metrics.

Both the CheXpert labeler and CheXbert output a 14-dimensional discrete vector representing 13 observations along with a label indicating "no findings". The values in this vector denote presence (1), absence (0), uncertainty (-1), and unknown (-2). We binarize this vector, treating both presence and uncertainty as 1, and absence and unknown as 0. This enables the computation of CheXpert F1, CheXbert F1, CheXpert Accuracy, and CheXbert Accuracy.


Regarding the RadGraph F1 metric, drawing inspiration from the methodology of Delbrouck et al. \citeyearpar{delbrouck-etal-2022-improving}, we employ the pretrained entity and relation extraction model provided within the dataset. This model has been conveniently made accessible as an installable package (\url{https://pypi.org/project/radgraph/}). Specifically, we assess the RadGraph F1 Partial variant included in the package, as it was the officially designated variant for the First Shared Task on Clinical Text Generation: RRG24 \cite{xu-etal-2024-overview}. Additionally, we explore another variant, which we term RadGraph F1 Full. This variant is based on the underlying model of the package. We achieve this by creating a "bag" comprising entities, relations without type, and relations with type, and then computing the F1 score between the "bag" of a referenced report and that of a generated report.

In addition to the results that were already presented in Table \ref{table:metrics-results} (Section \ref{sec:exp}), in this appendix we include the plots shown in Figures \ref{figure:mean_avg_jaccard_report_iuxray}, \ref{figure:mean_avg_jaccard_report_chestimg},
\ref{figure:mean_avg_acc_sentences_chestimg}, \ref{figure:avg_num_contr_sentences_chestimg}.

\subsection{Hardware and Other Implementation Details}
\label{sec:appendix:implementation-details}

All of our experiments are implemented using Python 3.10.10 with PyTorch version 1.13.1+cu117 \citep{paszke2017automatic}. All experiments are conducted on a computing node equipped with a 20-core Intel(R) Core(TM) i9-9900X CPU @ 3.50GHz, three NVIDIA GPUs - two GeForce RTX 2080 Ti with 11GB memory and one GeForce RTX 3090 with 24GB memory. The system is complemented by 125GB of RAM. 

We implement multitask learning for \cxrfactencodershortname\ using interleaved dataloaders, multiple model forward passes, and multiple gradient accumulation steps. Specifically, our model features distinct forward functions for each task, with each task assigned its own dataloader. These dataloaders are interleaved according to weights that determine the sampling frequency for each task. To ensure all tasks contribute to the gradients during training, we employ enough gradient accumulation steps so that each task's batch is sampled at least once before performing backpropagation.

We use the AdamW optimizer \cite{loshchilov2018decoupled} with a cyclic exponential learning rate that varies from 8e-5 to 1e-6 over 8 epochs. Here, an epoch consists of roughly 800 batches. Typically, our experiments run for 12-18 hours, after which we observe no significant gains in validation metrics.

\subsection{ChatGPT prompts}

\end{document}

%% file: math_commands.tex
\usepackage{amsmath,amsfonts,bm}

\def\eqref#1{equation~\ref{#1}}

\def\1{\bm{1}}

\DeclareMathAlphabet{\mathsfit}{\encodingdefault}{\sfdefault}{m}{sl}
\SetMathAlphabet{\mathsfit}{bold}{\encodingdefault}{\sfdefault}{bx}{n}



%% file: tables/iu-xray-example.tex
\begin{figure}[!tb]
\centering
    \begin{tabular}{cl}
    \begin{minipage}{0.15\textwidth}
        \includegraphics[width=\linewidth]{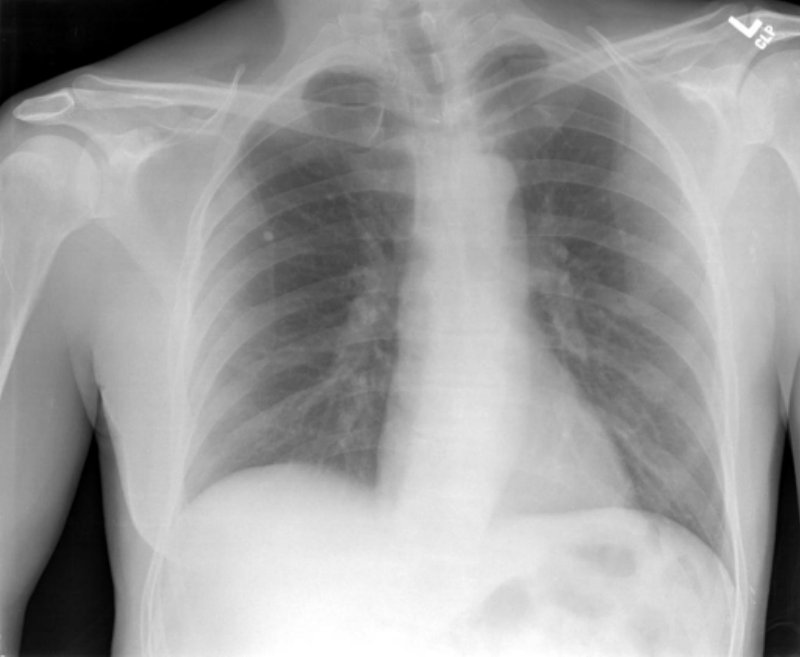}
    \end{minipage}
    &
    \begin{minipage}{0.27\textwidth}
        \tiny
        \textbf{Comparison:}
        Chest radiographs XXXX.\\
        \textbf{Indication:}
        XXXX-year-old male, chest pain.\\
        \textbf{Findings:}
        The cardiomediastinal silhouette is within normal limits for size and contour. The lungs are normally inflated without evidence of focal airspace disease, pleural effusion, or pneumothorax. \textcolor{purple}{Stable calcified granuloma within the right upper lung.} \textcolor{orange}{No acute bone abnormality.}\\
        \textbf{Impression:}
        No acute cardiopulmonary process. \\
    \end{minipage}
    \end{tabular}
\caption{Example image and report from the IU X-ray dataset \cite{10.1093/jamia/ocv080}}
\label{figure:iu-xray-example}
\end{figure}

%% file: tables/triplet-and-sentence-ranking-results.tex
\begin{table*}[!th]
\begin{center}
\caption{Triplet and sentence ranking results. \textbf{Triplet ranking}: 1000 samples per rule. For each rule, we report the fraction of correctly ranked triplets. ($^*$) denotes perfect scores achieved by BioViL-T due to an unfair advantage from how triplets were heuristically sampled in Rules 2 an 4. (o) stands for observations and (a) for anatomical locations. \textbf{Sentence ranking}: 2412 sentences annotated by radiologists, sourced from the Chest ImaGenome Gold dataset. Notation: a@k represents the mean average accuracy of the top k ranked sentences (larger is better), while c@k represents the mean number of contradictory sentences among the top k ranked sentences (smaller is better).}
\label{table:triplet-and-sent-rank-results}
\resizebox{1.9\columnwidth}{!}{
\begin{tabular}{l|l|c@{\hspace{2mm}}c@{\hspace{2mm}}c@{\hspace{2mm}}c@{\hspace{2mm}}c@{\hspace{2mm}}c@{\hspace{2mm}}c|c@{\hspace{2mm}}c@{\hspace{2mm}}c@{\hspace{2mm}}c@{\hspace{2mm}}c}

\toprule

\multirow{2}{*}{\textbf{ID}} & \multirow{2}{*}{\textbf{Text Model}}  & \multicolumn{7}{c|}{\textbf{Triplet Ranking}} & \multicolumn{5}{c}{\textbf{Sentence Ranking}} \\

 & & \textbf{R1(o)} & \textbf{R1(a)} & \textbf{R2} & \textbf{R3} & \textbf{R4} & \textbf{R5} & \textbf{R6}
 &
 \textbf{AUC} & \textbf{a@50} & \textbf{a@100} & \textbf{c@50} & \textbf{c@100}\\

\midrule

1 & BioLinkBERT &
0.753 & 0.725 &  0.786 & 0.756 & 0.644 & 0.774 & 0.520
& 0.717 & 0.951 & 0.945 & 5.523 & 12.158 \\

2 & PubMedBERT &
0.901 & 0.853 & 0.905 & 0.873 & 0.767 & 0.834 & 0.603
& 0.775 & 0.954 & 0.947 & 4.552 & 10.134 \\

3 & BioClinicalBERT &
\textbf{0.922} & \textbf{0.864} & 0.933 & 0.912 & 0.834 & \textbf{0.948} & 0.601
& 0.830 & 0.957 & 0.950 & 3.823 & 8.615 \\

4 & CheXbert &
0.855 & 0.771 & 0.908 & 0.884 & 0.760 & 0.937 & 0.635
& \textbf{0.864} & \textbf{0.962} & \textbf{0.955} & \textbf{1.914} & \textbf{4.299} \\

5 & CXR-BERT-specialized  &
0.880 & 0.804 & 0.992& 0.914& 0.904& 0.932 & 0.717
& 0.837 & 0.953 & 0.947 & 2.905 & 6.230 \\

6 & BioViL-T &
0.910 & 0.851 & \color{red}{\textbf{1.000}}$^*$ &  \textbf{0.938} & \color{red}{\textbf{1.000}}$^*$ &  0.944 & \textbf{0.765}
& 0.845 & 0.956 & 0.949 & 3.158 & 6.903 \\

\midrule

7 & \cxrfactencodershortname (T) & 0.968 & 0.955 & 0.925 & 0.964 & 0.798 & 0.952 & 0.946
& 0.896 & 0.963 & 0.957 & 1.668 & 3.940 \\

8 & \cxrfactencodershortname (T+C) & 0.967 & 0.945 & 0.967 & 0.982 & \textbf{0.926} & \color{red}{\textbf{0.988}} & 0.937
& \textbf{\color{red}{0.919}} & \textbf{\color{red}{0.975}} & \textbf{\color{red}{0.967}} & 2.955 & 9.157 \\

9 & \cxrfactencodershortname (T+ER) & 0.962 & 0.946 & 0.917 & 0.961 & 0.798 & 0.954 & 0.927
& 0.888 & 0.964 & 0.957 & 1.403 & 3.543 \\

10 & \cxrfactencodershortname (T+SD) & \color{red}{\textbf{0.981}} & \color{red}{\textbf{0.966}} & 0.954 & 0.977 & 0.875 & 0.981 & 0.898
& 0.897 & 0.961 & 0.955 & 2.427 & 5.465 \\

11 & \cxrfactencodershortname (T+EC) & 0.957 & 0.953 & 0.950 & 0.971 & 0.809 & 0.965 & 0.943
& 0.840 & 0.957 & 0.951 & 1.237 & 2.522 \\

12 & \cxrfactencodershortname (T+NLI) & 0.910 & 0.898 & 0.958 & 0.970 & 0.861 & 0.968 & 0.903
& 0.820 & 0.951 & 0.945 & 1.378 & 3.240 \\

13 & \cxrfactencodershortname (T+EC+NLI) & 0.928 & 0.950 & 0.923 & 0.961 & 0.777 & 0.934 & 0.925
& 0.812 & 0.950 & 0.943 & \textbf{\color{red}{0.969}} & \textbf{\color{red}{2.478}} \\

14 & \cxrfactencodershortname (T+C+EC+NLI) & 0.971 & 0.932 & 0.974 & 0.980 & 0.892 & 0.982 & 0.945
& 0.890 & 0.969 & 0.960 & 1.437 & 3.789 \\

15 & \cxrfactencodershortname (T+C+EC+NLI+ER) & 0.972 & 0.944 & \textbf{0.978} & \textbf{\color{red}{0.983}} & 0.911 & 0.984 & 0.936
& 0.917 & 0.972 & 0.964 & 1.719 & 4.694 \\

16 & \cxrfactencodershortname (T+C+EC+NLI+SD) & 0.977 & 0.953 & 0.972 & 0.980 & 0.883 & 0.980 & 0.951
& 0.906 & 0.971 & 0.962 & 1.445 & 3.603 \\

17 & \cxrfactencodershortname (T+C+EC+NLI+ER+SD) & 0.976 & 0.958 & 0.976 & 0.982 & 0.880 & 0.980 & \textbf{\color{red}{0.952}}
& 0.907 & 0.970 & 0.962 & 1.502 & 3.804 \\

\bottomrule

\end{tabular}
}
\end{center}
\end{table*}

%% file: tables/nli-entcont-results.tex
\begin{center}
\caption{NLI results as sentence similarity. a$_{\text{E+C}}^{*}$ denotes the upper bound accuracy with optimal similarity threshold tuning on the same evaluation data. BT stands for best threshold (tuned in NLI custom dataset).}
\label{table:nli-entcont-results}
\resizebox{1.7\columnwidth}{!}{
\begin{tabular}{l|l|c@{\hspace{2mm}}r@{\hspace{2mm}}r@{\hspace{2mm}}r|r@{\hspace{2mm}}r@{\hspace{2mm}}r@{\hspace{2mm}}r|r@{\hspace{2mm}}r@{\hspace{2mm}}r@{\hspace{2mm}}r}
\toprule
& & \multicolumn{4}{c|}{\wrappable{3.5cm}{\textbf{NLI custom dataset} \\ Ent. pairs: 25,175 \\ Cont. pairs: 99,121}} & \multicolumn{4}{c|}{\wrappable{3cm}{\textbf{RadNLI test set} \\ Ent. pairs: 102 \\ Cont. pairs: 98}} & \multicolumn{4}{c}{\wrappable{3.5cm}{\textbf{MS-CXR-T} \\ Ent. pairs: 141 \\ Cont. pairs: 220}} \\[20pt]
\textbf{ID} & \textbf{Text Model} &
\textbf{BT} & \textbf{a}$_{\text{E}}$ & \textbf{a}$_{\text{C}}$ &  \textbf{a}$_{\text{E+C}}$ &
\textbf{a}$_{\text{E}}$ &  \textbf{a}$_{\text{C}}$ & \textbf{a}$_{\text{E+C}}$ & \textbf{a}$_{\text{E+C}}^{*}$ &
\textbf{a}$_{\text{E}}$ &  \textbf{a}$_{\text{C}}$ & \textbf{a}$_{\text{E+C}}$ & \textbf{a}$_{\text{E+C}}^{*}$ \\

\midrule


1 & PubMedBERT &
1.000 & 1.9 & 99.8 & 50.8 & 0.0 & 100.0 & 50.0 & 63.8 & 0.0 & 100.0 & 50.0 & 56.5 \\


2 & BioLinkBERT-large &
0.999 & 1.9 & 100.0 & 50.9 & 0.0 & 100.0 & 50.0 & 65.9 & 0.0 & 100.0 & 50.0 & 58.1 \\


3 & BioClinicalBERT &
1.000 & 1.9 & 100.0 & 50.9 & 0.0 & 100.0 & 50.0 & 69.2 & 0.0 & 100.0 & 50.0 & 69.5 \\

4 & CheXbert &
0.556 & 88.3 & 33.8 & 61.0 & 96.1 & 83.7 & \textbf{89.9} & \textbf{90.4} & 100.0 & 1.8 & 50.9 & 63.3 \\

5 & CXR-BERT-specialized &
0.619 & 76.9 & 44.6 & 60.8 & 73.5 & 85.7 & 79.6 & 82.3 & 100.0 & 8.6 & \textbf{54.3} & 77.5 \\

6 & BioViL-T &
0.709 & 70.9 & 51.4 & \textbf{61.2} & 58.8 & 89.8 & 74.3 & 78.0 & 100.0 & 7.7 & 53.9 & \textbf{87.8} \\

\midrule


7 & \cxrfactencodershortname (T) & 0.640 & 79.8 & 57.2 & 68.5 & 69.6 & 89.8 & 79.7 & 87.3 & 100.0 & 18.6 & 59.2 & 78.0 \\

8 & \cxrfactencodershortname (T+C) & 0.934 & 43.3 & 69.9 & 56.6 & 44.1 & 95.9 & 70.0 & 75.4 & 97.9 & 12.3 & 55.1 & 62.6 \\

9 & \cxrfactencodershortname (T+ER) & 0.606 & 81.6 & 54.3 & 67.9 & 70.6 & 89.8 & 80.2 & 85.9 & 100.0 & 16.8 & 58.4 & 78.5 \\

10 & \cxrfactencodershortname (T+SD) & 0.620 & 68.1 & 50.5 & 59.3 & 81.3 & 79.6 & 80.5 & 81.9 & 99.3 & 10.9 & 55.1 & 70.3 \\

11 & \cxrfactencodershortname (T+EC) & 0.308 & 97.0 & 93.7 & \textbf{\color{red}{95.3}} & 98.0 & 93.8 & 96.0 & 96.4 & 96.5 & 69.5 & 83.0 & 93.3 \\

12 & \cxrfactencodershortname (T+NLI) & 0.233 & 89.3 & 78.6 & 84.0 & 99.0 & 93.9 & \textbf{\color{red}{96.4}} & \textbf{\color{red}{97.0}} & 100.0 & 75.9 & 87.9 & 95.7 \\

13 & \cxrfactencodershortname (T+EC+NLI) & 0.267 & 96.7 & 93.3 & 95.0 & 98.0 & 93.9 & 96.0 & \textbf{\color{red}{97.0}} & 99.3 & 84.1 & \textbf{\color{red}{91.7}} & 97.6 \\

14 & \cxrfactencodershortname (T+C+EC+NLI) &  0.288 & 94.9 & 82.4 & 88.6 & 99.0 & 91.8 & 95.4 & 96.0 & 100.0 & 75.9 & 88.9 & 97.2 \\

15 & \cxrfactencodershortname (T+C+EC+NLI+ER) & 0.431 & 95.2 & 83.0 & 89.1 & 99.0 & 89.8 & 94.4 & 95.5 & 100.0 & 59.0 & 79.5 & \textbf{\color{red}{98.5}} \\

16 & \cxrfactencodershortname (T+C+EC+NLI+SD) & 0.480 & 96.6 & 87.3 & 92.0 & 99.0 & 89.8 & 94.4 & 95.5 & 100.0 & 70.5 & 85.2 & 93.6 \\

17 & \cxrfactencodershortname (T+C+EC+NLI+ER+SD) & 0.455 & 97.1 & 87.0 & 92.0 & 98.0 & 91.8 & 94.9 & 96.0 & 100.0 & 72.7 & 86.4 & 96.0 \\

\bottomrule
\end{tabular}
}
\end{center}

%% file: tables/radnli-results.tex
\begin{center}
\caption{RadNLI test set accuracy. Results for CXR-BERT, IFCC, PTUnifier and DoT5 are from the original papers. CoT stands for Chain-of-Thought prompting. }
\label{table:radnli-results}
\resizebox{0.9\columnwidth}{!}{
\small
\begin{tabular}{l|l|c}
\toprule
\textbf{ID} & \textbf{Text Model} & \textbf{Accuracy} \\
\midrule
1 & CXR-BERT \cite{boecking2022making} & 65.2 \\
2 & IFCC \cite{miura-etal-2021-improving} & 77.8 \\
3 & PTUnifier \cite{chen2023towards} & 80.0 \\
4 & DoT5 \cite{10.1162/tacl_a_00585} & \textbf{82.1} \\

\midrule

5 & GPT-4 w/ simple prompt  & 82.3 \\
6 & GPT-4 w/ CoT + examples & \textbf{89.0} \\

7 & Meta-Llama-3-8B w/ simple prompt &  58.1 \\
8 & Meta-Llama-3-8B w/ CoT + examples  & 61.5 \\

\midrule

9 & \cxrfactencodershortname (T+NLI) & 84.2 \\
10 & \cxrfactencodershortname (T+EC+NLI) & 81.3 \\
11 & \cxrfactencodershortname (T+C+EC+NLI) & 86.7 \\
12 & \cxrfactencodershortname (T+C+EC+NLI+SD) & 85.2 \\
13 & \cxrfactencodershortname (T+C+EC+NLI+ER) & 88.1 \\
14 & \cxrfactencodershortname (T+C+EC+NLI+ER+SD) & 88.5 \\
15 & \cxrfactencodershortname (NLI fine-tuning) & \textbf{\color{red}{89.8}} \\

\bottomrule

\end{tabular}
}
\end{center}

%% file: tables/metrics-results.tex
\begin{table*}[!th]
\begin{center}

\caption{Comparative evaluation of text generation metrics. Notation: a@k denotes the mean average accuracy up to the $k$th sentence, c@k represents the mean number of contradictory sentences up to the $k$th sentence, and j@k represents the mean average Jaccard index up to the $k$th sentence.}

\label{table:metrics-results}
\resizebox{2.1\columnwidth}{!}{
\begin{tabular}{l|l|ccccc|cc|cc|c}
\toprule
\multirow{2}{*}{\textbf{ID}} & \multirow{2}{*}{\textbf{Metric}}  & \multicolumn{5}{|c|}{\textbf{Chest ImaGenome Gold Sentences (2412)}} & \multicolumn{2}{|c|}{\wrappable{3.2cm}{\textbf{Chest ImaGenome} \\ \textbf{Gold Reports (500)}}} & \multicolumn{2}{|c|}{\wrappable{2.5cm}{\textbf{IU X-Ray Reports (3955)}}} & \wrappable{2cm}{\textbf{RadNLI + MS-CXR-T}} \\
 & & \textbf{AUC} & \textbf{a@50} & \textbf{a@100} & \textbf{c@50} & \textbf{c@100} & \textbf{j@20} & \textbf{j@50} & \textbf{j@20} & \textbf{j@50} & \textbf{AUC} \\

\midrule

1 & BLEU \cite{papineni2002bleu} &
0.767 & 0.951 & 0.944 & 4.432 & 9.148 & 0.509 & 0.469 & 0.392 & 0.336 & 0.537\\

2 & ROUGE-L \cite{lin-2004-rouge} &
0.773 & 0.953 & 0.946 & 4.228 & 9.029 & 0.508 & 0.466  & 0.391 & 0.335 & 0.547 \\

3 & METEOR \cite{banerjee2005meteor} &
0.829 & 0.954 & 0.947 & 4.282 & 10.100 & 0.514 & 0.471 & 0.390 & 0.328 & 0.596 \\

4 & CIDEr-D \cite{vedantam2015cider} &
0.778 & 0.952 & 0.946 & 4.883 & 10.234 & 0.482 & 0.441 & 0.415 & 0.366 & 0.556 \\

5 & BERTScore \cite{bert-score} &
\textbf{0.840} & 0.960 & 0.952 & 5.012 & 11.328 & 0.531 & 0.489 & \textbf{0.429} & \textbf{0.383} & 0.559 \\

6 & CheXpert Accuracy \cite{irvin2019chexpert} &
0.764 & 0.941 & 0.939 & 3.263 & 6.263 & 0.467 & 0.446 & 0.375 & 0.322  & 0.592 \\

7 & CheXpert F1 \cite{irvin2019chexpert} &
0.742 & 0.939 & 0.938 & 3.658 & 6.877 & 0.460 & 0.441 & 0.312 & 0.327 & 0.582 \\

8 & CheXbert Accuracy \cite{chexbert} &
0.778 & 0.941 & 0.941 & \textbf{2.667} & \textbf{5.103} & 0.485 & 0.452 & 0.356 & 0.357 & 0.592 \\

9 & CheXbert F1 \cite{chexbert} &
0.753 & 0.939 & 0.939 & 2.709 & 5.301 & 0.472 & 0.445 & 0.365 & 0.369 & 0.583 \\


10 & RadGraph F1 (Full) \cite{radgraph} &
0.831 & \textbf{0.961} &  \textbf{0.953} & 2.881 & 7.189 & 0.546 & 0.500 & \textbf{0.429} & 0.371 & \textbf{0.610} \\

11 & RadGraph F1 (Partial) \cite{radgraph} &
0.789 & 0.960 & 0.951 & 3.453 & 8.833 & \textbf{0.549} & \textbf{0.501} & 0.415 & 0.361 & 0.574 \\

\midrule

12 & \cxrfactencodermetricname (T+C+EC+NLI) &
0.912 & 0.967 & 0.958 & 1.563 & 3.953 & 0.562 & 0.510 & 0.504 & 0.460 & 0.927 \\

13 & \cxrfactencodermetricname (T+C+EC+NLI+ER) &
\textbf{\color{red}{0.921}} & \textbf{\color{red}{0.968}} & \textbf{\color{red}{0.959}} & 1.575 & 4.122 & \textbf{\color{red}{0.563}} & \textbf{\color{red}{0.511}} & 0.513 & 0.471 & 0.931 \\

14 & \cxrfactencodermetricname (T+C+EC+NLI+SD) &
0.911 & 0.967 & 0.958 & \textbf{\color{red}{1.518}} & \textbf{\color{red}{3.763}} & 0.553 & 0.505 & 0.517 & 0.471 & 0.936 \\

15 & \cxrfactencodermetricname (T+C+EC+NLI+ER+SD) &
0.911 & 0.967 & 0.958 & 1.606 & 3.874 & 0.555 & 0.507 & \textbf{\color{red}{0.518}} & \textbf{\color{red}{0.473}} & \textbf{\color{red}{0.938}} \\

\bottomrule

\end{tabular}
}
\end{center}
\end{table*}

%% file: tables/report-recovery-results.tex
\begin{table*}[!th]
\begin{center}
\caption{Template-based report generation metrics on MIMIC-CXR test set (3269 reports) for different label extraction methods. \cxrfactencodermetricname\ was calculated using the T+C+EC+NLI+ER+SD variant. Notation: FE = Fact Extraction; BS-F1 = BERTScore F1; B-4 = BLEU-4; C-D = CIDEr-D; R-L = ROUGE-L; MET = METEOR.}
\label{table:report-recovery-results}
\resizebox{1.9\columnwidth}{!}{
\begin{tabular}{l|l|c|c@{\hspace{2mm}}c|c@{\hspace{2mm}}c|c@{\hspace{2mm}}c|c@{\hspace{2mm}}c@{\hspace{2mm}}c@{\hspace{2mm}}c@{\hspace{2mm}}c}

\toprule

\multirow{2}{*}{\textbf{ID}} & \multirow{2}{*}{\wrappable{3.5cm}{\textbf{Label Extraction Method}}}  & \multirow{2}{*}{\textbf{\cxrfactencodermetricname}} & \multicolumn{2}{|c|}{\textbf{RadGraph F1}} & \multicolumn{2}{|c|}{\textbf{CheXpert F1}} & \multicolumn{2}{|c|}{\textbf{CheXbert F1}} &
\multirow{2}{*}{\textbf{BS-F1}} & \multirow{2}{*}{\textbf{B-4}} & \multirow{2}{*}{\textbf{C-D}} & \multirow{2}{*}{\textbf{R-L}} & \multirow{2}{*}{\textbf{MET}} \\

& & & \textbf{Full} & \textbf{Partial} & \textbf{Micro} & \textbf{Macro} & \textbf{Micro} & \textbf{Macro} \\

\midrule

1 & CheXpert labeler &
0.644 & 
0.119 & 
0.161 & 
\color{red}{\textbf{0.998}} & 
\color{red}{\textbf{0.990}} & 
0.939 & 
0.854 & 
\textbf{0.470} & 
\textbf{0.007} & 
\textbf{0.023} & 
\textbf{0.123} & 
\textbf{0.179} \\ 

2 & CheXbert &
0.647 & 
\textbf{0.120} & 
0.162 & 
0.948 & 
0.921 & 
\color{red}{\textbf{0.983}} & 
\textbf{0.907} & 
\textbf{0.470} & 
\textbf{0.007} & 
\textbf{0.023} & 
\textbf{0.123} & 
\textbf{0.179} \\ 

3 & Chest ImaGenome & 
\textbf{0.677} & 
0.104 & 
\textbf{0.237} & 
0.767 & 
0.693 & 
0.776 & 
0.751 & 
0.257 & 
0.003 & 
0.002 & 
0.086 & 
0.170 \\ 

\midrule




4 & FE (T5-small) &
\color{red}{\textbf{0.983}} & 
\color{red}{\textbf{0.784}} & 
0.745 & 
\textbf{0.973} & 
\textbf{0.964} & 
\textbf{0.959} & 
\color{red}{\textbf{0.947}} & 
\color{red}{\textbf{0.789}} & 
0.275 & 
0.672 & 
\color{red}{\textbf{0.533}} & 
\color{red}{\textbf{0.650}} \\ 

5 & FE (GPT-4) &
0.937 & 
0.773 & 
\color{red}{\textbf{0.751}} & 
0.969 & 
0.959 & 
0.957 & 
0.946 & 
0.777 & 
0.275 & 
0.853 & 
0.503 & 
0.635 \\ 

6 & FE (Meta-Llama-3-8B) &
0.922 & 
0.737 & 
0.738 & 
0.944 & 
0.940 & 
0.932 & 
0.925 & 
0.775 & 
\color{red}{\textbf{0.286}} & 
\color{red}{\textbf{0.902}} & 
0.522 & 
0.636 \\ 

\bottomrule

\end{tabular}
}
\end{center}
\end{table*}

%% file: tables/fact-extraction-examples.tex
\begin{table*}[!th]
    \centering
    \caption{Examples of facts extracted by GPT-4, with the prompt shown in Figure \ref{figure:chatgpt-prompts:fact-extraction}}
    \label{table:fact-extraction-examples}

    \scalebox{0.7}
    {
    \begin{tabular}{|c|c|c|}
        \toprule
        
        \textbf{Original sentence} & \textbf{Extracted facts} & 
        \textbf{Comment} \\
        
        \midrule

        \wrappable{6cm}{2.Atelectasis of the left lower lobe 3.Stable left lower lobe laceration.} &
        
        \begin{minipage}{6cm}
        \begin{itemize}[leftmargin=*]
            \setlength\itemsep{-0.5em}
            \item Atelectasis of the left lower lobe
            \item Stable left lower lobe laceration
        \end{itemize}
        \end{minipage} &

        \wrappable{6cm}{In this example, NLTK's sent\_tokenize failed to split the text into two sentences, but GPT-4 has no problems identifying the facts.} \\

        \midrule

        \wrappable{5cm}{A CT of the abdomen showing the lower chest performed on adds the detection of acinar-sized nodules, some of which are confluent at the lung periphery and confirms the presence of small bilateral pleural effusions.} &

        \begin{minipage}{6cm}
        \begin{itemize}[leftmargin=*]
            \setlength\itemsep{-0.5em}
            \item detection of acinar-sized nodules at the lung periphery
            \item some nodules are confluent at the lung periphery
            \item presence of small bilateral pleural effusions
        \end{itemize}
        \end{minipage} &

        \wrappable{6cm}{A long complicated sentence broken down into 3 shorter factual statements.} \\

        \midrule

        \wrappable{6cm}{NO chf, focal inifiltrate, or gross effusionis identified} &

        \begin{minipage}{6cm}
        \begin{itemize}[leftmargin=*]
            \setlength\itemsep{-0.5em}
            \item no CHF identified
            \item no focal infiltrate identified
            \item no gross effusion identified
        \end{itemize}
        \end{minipage} &

        \wrappable{6cm}{A negated sentence becomes 3 short facts, each one preserving the negation. GPT-4 also correctly fixes a typo.} \\

        \midrule

        \wrappable{6cm}{Severe cardiomegaly is again noted as well as aortic valve replacement including stent material at the aortic root compatible with appearance of Carevalve Aortic bioprothesis.} &

        \begin{minipage}{6cm}
        \begin{itemize}[leftmargin=*]
            \setlength\itemsep{-0.5em}
            \item Severe cardiomegaly
            \item aortic valve replacement
            \item stent material at the aortic root
            \item appearance of Carevalve Aortic bioprothesis at the aortic root
        \end{itemize}
        \end{minipage} &

        \wrappable{6cm}{Complex sentence becomes 4 shorter facts.} \\

        \midrule

        \wrappable{6cm}{A right-sided internal jugular central venous catheter is seen, which appears slightly high in position, terminating in the region of the right innominate vein/right internal jugular/innominate vein junction.} &

        \begin{minipage}{6cm}
        \begin{itemize}[leftmargin=*]
            \setlength\itemsep{-0.5em}
            \item right-sided internal jugular central venous catheter is seen
            \item catheter appears slightly high in position'
            \item catheter terminating in the region of the right innominate vein
            \item catheter terminating in the region of the right internal jugular
            \item catheter terminating in the region of the innominate vein junction
        \end{itemize}
        \end{minipage} &

        \wrappable{6cm}{Complex sentence becomes 5 shorter facts. Note GPT-4's ability to handle forward slashes (/).} \\

        \midrule

        \wrappable{6cm}{THEY REPORT TEXT FOLLOWS: The heart and great vessels are normal.} &

        \begin{minipage}{6cm}
        \begin{itemize}[leftmargin=*]
            \setlength\itemsep{-0.5em}
            \item normal heart
            \item normal great vessels
        \end{itemize}
        \end{minipage} &

        \wrappable{6cm}{GPT-4 successfully ignores a distracting header (in uppercase) and correctly distributes the word ``normal`` to each extracted fact.} \\

        \bottomrule
        
    \end{tabular}
    }
\end{table*}

%% file: tables/template-based-reports.tex
\begin{table*}
    \centering
    \caption{Examples of template-based generated reports for different label extraction algorithms.}
    \label{table:template-based-reports}
    \scalebox{0.7}{
    \begin{tabular}[t]{|c|c|c|c|}
        \toprule
        
        \textbf{Ground-truth Report} &
        \wrappablebold{3cm}{FE (T5-small)} & 
        \wrappablebold{3cm}{FE (GPT-4)} & 
        \wrappablebold{5.3cm}{FE (Meta-Llama-3-8B)} \\
        
        \midrule
        
        \begin{minipage}[t]{5.3cm}
            \small New PICC line on the right is projecting with its tip somewhere in the mediastinum. Appears to cross the midline, there is concern for potential arterial location. The initial line concerns were communicated over the telephone at the time of the wet read. Repeat PA and lateral radiograph, taken approximately an hour after the radiograph demonstrated the PICC line in the mid SVC. Potential small right pleural effusion. Stable moderate cardiomegaly.
        \end{minipage} & 
        
        \begin{minipage}[t]{5.3cm}
            \small new PICC line on the right. tip of PICC line in the mediastinum. potential arterial location crossing the midline. PICC line in the mid SVC. potential small right pleural effusion. stable moderate cardiomegaly

            \vspace{1em}

            \textbf{\cxrfactencodermetricname:} 1.000 \\
            \textbf{RadGraph F1 Full:} 0.750 \\
            \textbf{CheXpert Acc:} 1.0 \\
            \textbf{CheXbert Acc:} 1.0
        \end{minipage} & 
        
        \begin{minipage}[t]{5.3cm}
            \small new PICC line on the right. PICC line tip projecting in the mediastinum. potential arterial location crossing the midline. repeat PA and lateral radiograph taken approximately an hour after the previous radiograph. PICC line observed in the mid SVC. potential small right pleural effusion. stable moderate cardiomegaly

            \vspace{1em}
            
            \textbf{\cxrfactencodermetricname:} 0.891 \\
            \textbf{RadGraph F1 Full:} 0.899 \\
            \textbf{CheXpert Acc:} 1.0 \\
            \textbf{CheXbert Acc:} 1.0
        \end{minipage} &

        \begin{minipage}[t]{5.3cm}
            \small new PICC line on the right. PICC line tip in the mediastinum. appears to cross the midline. concern for potential arterial location. PA radiograph. lateral radiograph. PICC line in the mid SVC. potential small right pleural effusion. stable cardiomegaly

            \vspace{1em}
            
            \textbf{\cxrfactencodermetricname:} 0.966 \\
            \textbf{RadGraph F1 Full:} 0.813 \\
            \textbf{CheXpert Acc:} 1.0 \\
            \textbf{CheXbert Acc:} 1.0
        \end{minipage}
        

            
            
        
        \\

        \midrule

         &
        \wrappablebold{3cm}{CheXbert} & 
        \wrappablebold{3cm}{CheXpert labeler} & 
        \wrappablebold{3cm}{Chest ImaGenome} \\

        \midrule
        
        &

        \begin{minipage}[t]{5.3cm}
            \small the heart is enlarged. the cardiomediastinal silhouette is enlarged. no focal consolidation. the lungs are free of focal airspace disease. no atelectasis. a device is seen. pleural effusion is seen. no fibrosis. no pneumonia. no pneumothorax is seen. no pulmonary edema. no pulmonary nodules or mass lesions identified. no fracture is seen

            \vspace{1em}

            \textbf{\cxrfactencodermetricname:} 0.481 \\
            \textbf{RadGraph F1 Full:}  0.017 \\
            \textbf{CheXpert Acc:} 1.0 \\
            \textbf{CheXbert Acc:} 1.0
        \end{minipage} &
        
        \begin{minipage}[t]{5.3cm}
            \small the heart is enlarged. the cardiomediastinal silhouette is enlarged. no focal consolidation. the lungs are free of focal airspace disease. no atelectasis. a device is seen. pleural effusion is seen. no fibrosis. no pneumonia. no pneumothorax is seen. no pulmonary edema. no pulmonary nodules or mass lesions identified. no fracture is seen

            \vspace{1em}

            \textbf{\cxrfactencodermetricname:} 0.481 \\
            \textbf{RadGraph F1 Full:}  0.017 \\
            \textbf{CheXpert Acc:} 1.0 \\
            \textbf{CheXbert Acc:} 1.0
        \end{minipage} &

        \begin{minipage}[t]{5.3cm}
            \small enlarged cardiac silhouette in cardiac silhouette. abnormal cardiac silhouette. picc in left shoulder. picc in mediastinum. lung opacity in right costophrenic angle. pleural effusion in right costophrenic angle. abnormal right costophrenic angle. lung opacity in right lung. pleural effusion in right lung. abnormal right lung. picc in right shoulder. picc in svc. enlarged cardiac silhouette. lung opacity. pleural effusion. picc

            \vspace{1em}
            
            \textbf{\cxrfactencodermetricname:} 0.660 \\
            \textbf{RadGraph F1 Full:}  0.121 \\
            \textbf{CheXpert Acc:} 1.0 \\
            \textbf{CheXbert Acc:} 1.0
        \end{minipage}

        \\

        \midrule
        \toprule

        \textbf{Ground-truth Report} &
        \wrappablebold{3cm}{FE (T5-small)} & 
        \wrappablebold{3cm}{FE (GPT-4)} & 
        \wrappablebold{5.3cm}{FE (Meta-Llama-3-8B)} \\
        
        \midrule
        
        \begin{minipage}[t]{5.3cm}
            \small In comparison with the earlier study of this date, the Dobhoff tube can be seen only to the level of the distal esophagus, were crosses the lower margin of the image, to evaluate the tip, a view of the upper abdomen must be obtained using abdominal technique. The area of increased opacification at the right base seen previously is not appreciated at this time.
        \end{minipage} & 
        
        \begin{minipage}[t]{5.3cm}
            \small Dobhoff tube seen only to the level of the distal esophagus. Dobhoff tube crosses the lower margin of the image. evaluation of the tip of Dobhoff tube. suggestion for a view of the upper abdomen using abdominal technique. area of increased opacification at the right base not appreciated

            \vspace{1em}

            \textbf{\cxrfactencodermetricname:} 0.898 \\
            \textbf{RadGraph F1 Full:} 0.633 \\
            \textbf{CheXpert Acc:} 1.0 \\
            \textbf{CheXbert Acc:} 0.929
        \end{minipage} & 
        
        \begin{minipage}[t]{5.3cm}
            \small Dobhoff tube seen only to the level of the distal esophagus. distal esophagus crosses the lower margin of the image. need to obtain a view of the upper abdomen to evaluate the tip of the Dobhoff tube. previous increased opacification at the right base. current absence of increased opacification at the right base

            \vspace{1em}
            
            \textbf{\cxrfactencodermetricname:} 0.674 \\
            \textbf{RadGraph F1 Full:} 0.575 \\
            \textbf{CheXpert Acc:} 1.0 \\
            \textbf{CheXbert Acc:} 0.929
        \end{minipage} &

        \begin{minipage}[t]{5.3cm}
            \small observations of devices in the esophagus. device crosses the lower margin of the image in the esophagus. device tip cannot be evaluated in the esophagus. device tip must be evaluated in the upper abdomen. absence of increased opacification at the right base. absence of increased opacification at the right base (compared to previous study)

            \vspace{1em}
            
            \textbf{\cxrfactencodermetricname:} 0.489 \\
            \textbf{RadGraph F1 Full:} 0.193 \\
            \textbf{CheXpert Acc:} 1.0 \\
            \textbf{CheXbert Acc:} 0.929
        \end{minipage}
        
        \\

        \midrule

         &
        \wrappablebold{3cm}{CheXbert} & 
        \wrappablebold{3cm}{CheXpert labeler} & 
        \wrappablebold{3cm}{Chest ImaGenome} \\

        \midrule
        
        &

        \begin{minipage}[t]{5.3cm}
            \small heart size is normal. the mediastinal contour is normal. no focal consolidation. one or more airspace opacities are seen. no atelectasis. no pleural effusion. no fibrosis. no pneumonia. no pneumothorax is seen. no pulmonary edema. no pulmonary nodules or mass lesions identified. no fracture is seen

            \vspace{1em}

            \textbf{\cxrfactencodermetricname:} 0.344 \\
            \textbf{RadGraph F1 Full:}  0.0 \\
            \textbf{CheXpert Acc:} 0.929 \\
            \textbf{CheXbert Acc:} 1.0
        \end{minipage} &
        
        \begin{minipage}[t]{5.3cm}
            \small heart size is normal. the mediastinal contour is normal. no focal consolidation. one or more airspace opacities are seen. no atelectasis. a device is seen. no pleural effusion. no fibrosis. no pneumonia. no pneumothorax is seen. no pulmonary edema. no pulmonary nodules or mass lesions identified. no fracture is seen

            \vspace{1em}

            \textbf{\cxrfactencodermetricname:} 0.394 \\
            \textbf{RadGraph F1 full:}  0.0 \\
            \textbf{CheXpert Acc:} 1.0 \\
            \textbf{CheXbert Acc:} 0.929
        \end{minipage} &

        \begin{minipage}[t]{5.3cm}
            \small enteric tube in abdomen. enteric tube in mediastinum. enteric tube in neck. no lung opacity in right lower lung zone. no abnormal right lower lung zone. no lung opacity in right lung. no abnormal right lung. no lung opacity. enteric tube

            \vspace{1em}
            
            \textbf{\cxrfactencodermetricname:} 0.561 \\
            \textbf{RadGraph F1 Full:}  0.066 \\
            \textbf{CheXpert Acc:} 0.857 \\
            \textbf{CheXbert Acc:} 0.786
        \end{minipage}

        \\
        
        \bottomrule
        
    \end{tabular}
    }
\end{table*}

%% file: tables/template-based-reports-2.tex
\begin{table*}
    \centering
    \caption{More examples of template-based generated reports for different label extraction algorithms.}
    \label{table:template-based-reports-2}
    \scalebox{0.7}{
    \begin{tabular}[t]{|c|c|c|c|}
        \toprule
        
        \textbf{Ground-truth Report} &
        \wrappablebold{3cm}{FE (T5-small)} & 
        \wrappablebold{3cm}{FE (GPT-4)} & 
        \wrappablebold{5.3cm}{FE (Meta-Llama-3-8B)} \\
        
        \midrule
        
        \begin{minipage}[t]{5.3cm}
            \small Frontal and lateral chest radiographs were obtained. There are persistent, stable bilateral upper lung reticular nodular opacities consistent with history of sarcoidosis. No focal consolidation, pleural effusion, pneumothorax, or pulmonary edema is seen. The heart size is normal. Mediastinal and hilar contours are stable. 1. No focal consolidation to suggest pneumonia. 2. Stable bilateral upper lungs zone fibrosis consistent with history of sarcoidosis.
        \end{minipage} & 
        
        \begin{minipage}[t]{5.3cm}
            \small frontal chest radiograph obtained. lateral chest radiograph obtained. persistent bilateral upper lung reticular nodular opacities. stable bilateral upper lung reticular nodular opacities. opacities consistent with history of sarcoidosis. no focal consolidation. no pleural effusion. no pneumothorax. no pulmonary edema. normal heart size. stable mediastinal contours. stable hilar contours. no focal consolidation to suggest pneumonia. stable bilateral upper lungs zone fibrosis. fibrosis consistent with history of sarcoidosis

            \vspace{1em}

            \textbf{\cxrfactencodermetricname:} 0.982 \\
            \textbf{RadGraph F1 Full:} 0.862 \\
            \textbf{CheXpert Acc:} 1.0 \\
            \textbf{CheXbert Acc:} 0.857
        \end{minipage} & 
        
        \begin{minipage}[t]{5.3cm}
            \small persistent bilateral upper lung reticular nodular opacities. stable bilateral upper lung reticular nodular opacities. bilateral upper lung reticular nodular opacities consistent with history of sarcoidosis. no focal consolidation. no pleural effusion. no pneumothorax. no pulmonary edema. normal heart size. stable mediastinal contours. stable hilar contours. normal size heart. unremarkable mediastinum. clear lungs. no focal consolidation to suggest pneumonia. observation of diseases. stable bilateral upper lungs zone fibrosis. history of sarcoidosis

            \vspace{1em}
            
            \textbf{\cxrfactencodermetricname:} 0.911 \\
            \textbf{RadGraph F1 Full:} 0.723 \\
            \textbf{CheXpert Acc:} 1.0 \\
            \textbf{CheXbert Acc:} 1.0
        \end{minipage} &

        \begin{minipage}[t]{5.3cm}
            \small persistent bilateral upper lung reticular nodular opacities. consistent with history of sarcoidosis. no focal consolidation. no pleural effusion. no pneumothorax. no pulmonary edema. normal heart size. mediastinal contours are stable. hilar contours are stable. no focal consolidation to suggest pneumonia. diseases present. diseases such as pneumonia present. diseases such as pulmonary embolism present. diseases such as CHF present. stable bilateral upper lung zone fibrosis. fibrosis in the upper lung zone. consistent with history of sarcoidosis

            \vspace{1em}
            
            \textbf{\cxrfactencodermetricname:} 0.858 \\
            \textbf{RadGraph F1 Full:} 0.743 \\
            \textbf{CheXpert Acc:} 0.857 \\
            \textbf{CheXbert Acc:} 0.714
        \end{minipage}
        
        \\

        \midrule

         &
        \wrappablebold{3cm}{CheXbert} & 
        \wrappablebold{3cm}{CheXpert labeler} & 
        \wrappablebold{3cm}{Chest ImaGenome} \\

        \midrule
        
        &

        \begin{minipage}[t]{5.3cm}
            \small heart size is normal. the cardiomediastinal silhouette is enlarged. no focal consolidation. the lungs are free of focal airspace disease. no atelectasis. no pleural effusion. pleural thickening is present. no pneumonia. no pneumothorax is seen. no pulmonary edema. there are pulmonary nodules or mass identified. no fracture is seen

            \vspace{1em}

            \textbf{\cxrfactencodermetricname:} 0.691 \\
            \textbf{RadGraph F1 Full:} 0.363 \\
            \textbf{CheXpert Acc:} 0.857 \\
            \textbf{CheXbert Acc:} 0.929
        \end{minipage} &
        
        \begin{minipage}[t]{5.3cm}
            \small heart size is normal. the cardiomediastinal silhouette is enlarged. no focal consolidation. one or more airspace opacities are seen. no atelectasis. no pleural effusion. pleural thickening is present. there is evidence of pneumonia. no pneumothorax is seen. no pulmonary edema. there are pulmonary nodules or mass identified. no fracture is seen

            \vspace{1em}

            \textbf{\cxrfactencodermetricname:} 0.641 \\
            \textbf{RadGraph F1 Full:} 0.359 \\
            \textbf{CheXpert Acc:} 1.0 \\
            \textbf{CheXbert Acc:} 0.786
        \end{minipage} &

        \begin{minipage}[t]{5.3cm}
            \small no enlarged cardiac silhouette in cardiac silhouette. no abnormal cardiac silhouette. no consolidation in left costophrenic angle. no pleural effusion in left costophrenic angle. no pneumothorax in left costophrenic angle. no pulmonary edema or hazy opacity in left hilar structures. no consolidation in left lung. lung lesion in left lung. lung opacity in left lung. multiple masses or nodules in left lung. no pleural effusion in left lung. pleural or parenchymal scarring in left lung. no pneumothorax in left lung. no pulmonary edema or hazy opacity in left lung. interstitial lung disease in left lung. no pneumonia in left lung. abnormal left lung. lung lesion in left upper lung zone. lung opacity in left upper lung zone. multiple masses or nodules in left upper lung zone. pleural or parenchymal scarring in left upper lung zone. interstitial lung disease in left upper lung zone. abnormal left upper lung zone. no consolidation in right costophrenic angle. no pleural effusion in right costophrenic angle. no pneumothorax in right costophrenic angle. no pulmonary edema or hazy opacity in right hilar structures. no consolidation in right lung. lung lesion in right lung. lung opacity in right lung. multiple masses or nodules in right lung. no pleural effusion in right lung. pleural or parenchymal scarring in right lung. no pneumothorax in right lung. no pulmonary edema or hazy opacity in right lung. interstitial lung disease in right lung. no pneumonia in right lung. abnormal right lung. lung lesion in right upper lung zone. lung opacity in right upper lung zone. multiple masses or nodules in right upper lung zone. pleural or parenchymal scarring in right upper lung zone. interstitial lung disease in right upper lung zone. abnormal right upper lung zone. no consolidation. no enlarged cardiac silhouette. lung lesion. lung opacity. multiple masses or nodules. no pleural effusion. pleural or parenchymal scarring. no pneumothorax. no pulmonary edema or hazy opacity. interstitial lung disease. no pneumonia

            \vspace{1em}
            
            \textbf{\cxrfactencodermetricname:} 0.691 \\
            \textbf{RadGraph F1 Full:}  0.057 \\
            \textbf{CheXpert Acc:} 0.714 \\
            \textbf{CheXbert Acc:} 0.857
        \end{minipage}

        \\
        
        \bottomrule
        
    \end{tabular}
    }
\end{table*}

%% file: acl_latex.bbl
\begin{thebibliography}{52}
\expandafter\ifx\csname natexlab\endcsname\relax\def\natexlab#1{#1}\fi

\bibitem[{Adams et~al.(2023)Adams, Truhn, Busch, Kader, Niehues, Makowski, and Bressem}]{adams2023leveraging}
Lisa~C Adams, Daniel Truhn, Felix Busch, Avan Kader, Stefan~M Niehues, Marcus~R Makowski, and Keno~K Bressem. 2023.
\newblock Leveraging gpt-4 for post hoc transformation of free-text radiology reports into structured reporting: a multilingual feasibility study.
\newblock \emph{Radiology}, 307(4):e230725.

\bibitem[{AI@Meta(2024)}]{llama3modelcard}
AI@Meta. 2024.
\newblock \href {https://github.com/meta-llama/llama3/blob/main/MODEL_CARD.md} {Llama 3 model card}.

\bibitem[{Alsentzer et~al.(2019)Alsentzer, Murphy, Boag, Weng, Jin, Naumann, and McDermott}]{alsentzer2019publicly}
Emily Alsentzer, John~R Murphy, Willie Boag, Wei-Hung Weng, Di~Jin, Tristan Naumann, and Matthew McDermott. 2019.
\newblock Publicly available clinical bert embeddings.
\newblock \emph{arXiv preprint arXiv:1904.03323}.

\bibitem[{Araujo et~al.(2023)Araujo, Moens, and Soto}]{make5010005}
Vladimir Araujo, Marie-Francine Moens, and Alvaro Soto. 2023.
\newblock \href {https://doi.org/10.3390/make5010005} {Learning sentence-level representations with predictive coding}.
\newblock \emph{Machine Learning and Knowledge Extraction}, 5(1):59--77.

\bibitem[{Banerjee and Lavie(2005)}]{banerjee2005meteor}
Satanjeev Banerjee and Alon Lavie. 2005.
\newblock {METEOR}: An automatic metric for {MT} evaluation with improved correlation with human judgments.
\newblock In \emph{Proc of the {ACL} Workshop on Intrinsic and Extrinsic Evaluation Measures for Machine Translation and/or Summarization}, pages 65--72, Ann Arbor, Michigan. ACL.

\bibitem[{Bannur et~al.(2023)Bannur, Hyland, Liu, Perez-Garcia, Ilse, Castro, Boecking, Sharma, Bouzid, Thieme et~al.}]{bannur2023learning}
Shruthi Bannur, Stephanie Hyland, Qianchu Liu, Fernando Perez-Garcia, Maximilian Ilse, Daniel~C Castro, Benedikt Boecking, Harshita Sharma, Kenza Bouzid, Anja Thieme, et~al. 2023.
\newblock Learning to exploit temporal structure for biomedical vision-language processing.
\newblock In \emph{Proceedings of the IEEE/CVF Conference on Computer Vision and Pattern Recognition}, pages 15016--15027.

\bibitem[{Boecking et~al.(2022)Boecking, Usuyama, Bannur, Castro, Schwaighofer, Hyland, Wetscherek, Naumann, Nori, Alvarez-Valle et~al.}]{boecking2022making}
Benedikt Boecking, Naoto Usuyama, Shruthi Bannur, Daniel~C Castro, Anton Schwaighofer, Stephanie Hyland, Maria Wetscherek, Tristan Naumann, Aditya Nori, Javier Alvarez-Valle, et~al. 2022.
\newblock Making the most of text semantics to improve biomedical vision--language processing.
\newblock In \emph{European conference on computer vision}, pages 1--21. Springer.

\bibitem[{Bustos et~al.(2019)Bustos, Pertusa, Salinas, and de~la Iglesia-Vay{\'a}}]{bustos2019padchest}
Aurelia Bustos, Antonio Pertusa, Jose-Maria Salinas, and Maria de~la Iglesia-Vay{\'a}. 2019.
\newblock Padchest: A large chest x-ray image dataset with multi-label annotated reports.
\newblock \emph{arXiv:1901.07441}.

\bibitem[{Chen et~al.(2023{\natexlab{a}})Chen, Diao, Wang, Li, and Wan}]{chen2023towards}
Zhihong Chen, Shizhe Diao, Benyou Wang, Guanbin Li, and Xiang Wan. 2023{\natexlab{a}}.
\newblock Towards unifying medical vision-and-language pre-training via soft prompts.
\newblock \emph{arXiv preprint arXiv:2302.08958}.

\bibitem[{Chen et~al.(2023{\natexlab{b}})Chen, Varma, Wan, Langlotz, and Delbrouck}]{chen-etal-2023-toward}
Zhihong Chen, Maya Varma, Xiang Wan, Curtis Langlotz, and Jean-Benoit Delbrouck. 2023{\natexlab{b}}.
\newblock \href {https://doi.org/10.18653/v1/2023.acl-short.41} {Toward expanding the scope of radiology report summarization to multiple anatomies and modalities}.
\newblock In \emph{Proceedings of the 61st Annual Meeting of the Association for Computational Linguistics (Volume 2: Short Papers)}, pages 469--484, Toronto, Canada. Association for Computational Linguistics.

\bibitem[{Delbrouck et~al.(2022)Delbrouck, Chambon, Bluethgen, Tsai, Almusa, and Langlotz}]{delbrouck-etal-2022-improving}
Jean-Benoit Delbrouck, Pierre Chambon, Christian Bluethgen, Emily Tsai, Omar Almusa, and Curtis Langlotz. 2022.
\newblock \href {https://doi.org/10.18653/v1/2022.findings-emnlp.319} {Improving the factual correctness of radiology report generation with semantic rewards}.
\newblock In \emph{Findings of the Association for Computational Linguistics: EMNLP 2022}, pages 4348--4360, Abu Dhabi, United Arab Emirates. Association for Computational Linguistics.

\bibitem[{Delbrouck et~al.(2023)Delbrouck, Varma, Chambon, and Langlotz}]{delbrouck2023overview}
Jean-Benoit Delbrouck, Maya Varma, Pierre Chambon, and Curtis Langlotz. 2023.
\newblock Overview of the radsum23 shared task on multi-modal and multi-anatomical radiology report summarization.
\newblock In \emph{The 22nd Workshop on Biomedical Natural Language Processing and BioNLP Shared Tasks}, pages 478--482.

\bibitem[{Demner-Fushman et~al.(2015)Demner-Fushman, Kohli, Rosenman, Shooshan, Rodriguez, Antani, Thoma, and McDonald}]{10.1093/jamia/ocv080}
Dina Demner-Fushman, Marc~D. Kohli, Marc~B. Rosenman, Sonya~E. Shooshan, Laritza Rodriguez, Sameer Antani, George~R. Thoma, and Clement~J. McDonald. 2015.
\newblock {Preparing a collection of radiology examinations for distribution and retrieval}.
\newblock \emph{Journal of the American Medical Informatics Association}, 23(2):304--310.

\bibitem[{Devlin et~al.(2019)Devlin, Chang, Lee, and Toutanova}]{devlin-etal-2019-bert}
Jacob Devlin, Ming-Wei Chang, Kenton Lee, and Kristina Toutanova. 2019.
\newblock \href {https://doi.org/10.18653/v1/N19-1423} {{BERT}: Pre-training of deep bidirectional transformers for language understanding}.
\newblock In \emph{Proceedings of the 2019 Conference of the North {A}merican Chapter of the Association for Computational Linguistics: Human Language Technologies, Volume 1 (Long and Short Papers)}, pages 4171--4186, Minneapolis, Minnesota. Association for Computational Linguistics.

\bibitem[{Eberts and Ulges(2020)}]{DBLP:conf/ecai/EbertsU20}
Markus Eberts and Adrian Ulges. 2020.
\newblock \href {https://doi.org/10.3233/FAIA200321} {Span-based joint entity and relation extraction with transformer pre-training}.
\newblock In \emph{{ECAI} 2020 - 24th European Conference on Artificial Intelligence, 29 August-8 September 2020, Santiago de Compostela, Spain, August 29 - September 8, 2020 - Including 10th Conference on Prestigious Applications of Artificial Intelligence {(PAIS} 2020)}, volume 325 of \emph{Frontiers in Artificial Intelligence and Applications}, pages 2006--2013. {IOS} Press.

\bibitem[{Gu et~al.(2020)Gu, Tinn, Cheng, Lucas, Usuyama, Liu, Naumann, Gao, and Poon}]{pubmedbert}
Yu~Gu, Robert Tinn, Hao Cheng, Michael Lucas, Naoto Usuyama, Xiaodong Liu, Tristan Naumann, Jianfeng Gao, and Hoifung Poon. 2020.
\newblock \href {http://arxiv.org/abs/arXiv:2007.15779} {Domain-specific language model pretraining for biomedical natural language processing}.

\bibitem[{Gu et~al.(2023)Gu, Zhang, Usuyama, Woldesenbet, Wong, Sanapathi, Wei, Valluri, Strandberg, Naumann et~al.}]{gu2023distilling}
Yu~Gu, Sheng Zhang, Naoto Usuyama, Yonas Woldesenbet, Cliff Wong, Praneeth Sanapathi, Mu~Wei, Naveen Valluri, Erika Strandberg, Tristan Naumann, et~al. 2023.
\newblock Distilling large language models for biomedical knowledge extraction: A case study on adverse drug events.
\newblock \emph{arXiv preprint arXiv:2307.06439}.

\bibitem[{Irvin et~al.(2019)Irvin, Rajpurkar, Ko, Yu, Ciurea-Ilcus, Chute, Marklund, Haghgoo, Ball, Shpanskaya et~al.}]{irvin2019chexpert}
Jeremy Irvin, Pranav Rajpurkar, Michael Ko, Yifan Yu, Silviana Ciurea-Ilcus, Chris Chute, Henrik Marklund, Behzad Haghgoo, Robyn Ball, Katie Shpanskaya, et~al. 2019.
\newblock Chexpert: A large chest radiograph dataset with uncertainty labels and expert comparison.
\newblock In \emph{Proceedings of the AAAI conference on artificial intelligence}, volume~33, pages 590--597.

\bibitem[{Jain et~al.(2021{\natexlab{a}})Jain, Agrawal, Saporta, Truong, Duong, Bui, Chambon, Zhang, Lungren, Ng, Langlotz, Rajpurkar, and Rajpurkar}]{radgraph}
Saahil Jain, Ashwin Agrawal, Adriel Saporta, Steven Truong, Du~Nguyen Duong~Nguyen Duong, Tan Bui, Pierre Chambon, Yuhao Zhang, Matthew Lungren, Andrew Ng, Curtis Langlotz, Pranav Rajpurkar, and Pranav Rajpurkar. 2021{\natexlab{a}}.
\newblock \href {https://datasets-benchmarks-proceedings.neurips.cc/paper_files/paper/2021/file/c8ffe9a587b126f152ed3d89a146b445-Paper-round1.pdf} {Radgraph: Extracting clinical entities and relations from radiology reports}.
\newblock In \emph{Proceedings of the Neural Information Processing Systems Track on Datasets and Benchmarks}, volume~1. Curran.

\bibitem[{Jain et~al.(2021{\natexlab{b}})Jain, Smit, Truong, Nguyen, Huynh, Jain, Young, Ng, Lungren, and Rajpurkar}]{visualchexbert}
Saahil Jain, Akshay Smit, Steven~QH Truong, Chanh~DT Nguyen, Minh-Thanh Huynh, Mudit Jain, Victoria~A. Young, Andrew~Y. Ng, Matthew~P. Lungren, and Pranav Rajpurkar. 2021{\natexlab{b}}.
\newblock \href {https://doi.org/10.1145/3450439.3451862} {Visualchexbert: Addressing the discrepancy between radiology report labels and image labels}.
\newblock In \emph{Proceedings of the Conference on Health, Inference, and Learning}, CHIL '21, page 105–115, New York, NY, USA. Association for Computing Machinery.

\bibitem[{Johnson et~al.(2019{\natexlab{a}})Johnson, Pollard, Berkowitz, Greenbaum, Lungren, Deng, Mark, and Horng}]{johnson2019mimicv1}
Alistair E.~W. Johnson, Tom~J. Pollard, Seth~J. Berkowitz, Nathaniel~R. Greenbaum, Matthew~P. Lungren, Chih-ying Deng, Roger~G. Mark, and Steven Horng. 2019{\natexlab{a}}.
\newblock Mimic-cxr, a de-identified publicly available database of chest radiographs with free-text reports.
\newblock \emph{Scientific Data}, 6(1):317.

\bibitem[{Johnson et~al.(2019{\natexlab{b}})Johnson, Pollard, Greenbaum, Lungren, Deng, Peng, Lu, Mark, Berkowitz, and Horng}]{johnson2019mimiccxrjpg}
Alistair~EW Johnson, Tom~J Pollard, Nathaniel~R Greenbaum, Matthew~P Lungren, Chih-ying Deng, Yifan Peng, Zhiyong Lu, Roger~G Mark, Seth~J Berkowitz, and Steven Horng. 2019{\natexlab{b}}.
\newblock \href {http://arxiv.org/abs/1901.07042} {Mimic-cxr-jpg, a large publicly available database of labeled chest radiographs}.
\newblock \emph{arXiv:1901.07042}.

\bibitem[{Johnson et~al.(2016)Johnson, Pollard, Shen, Lehman, Feng, Ghassemi, Moody, Szolovits, Anthony~Celi, and Mark}]{johnson2016mimic}
Alistair~EW Johnson, Tom~J Pollard, Lu~Shen, Li-wei~H Lehman, Mengling Feng, Mohammad Ghassemi, Benjamin Moody, Peter Szolovits, Leo Anthony~Celi, and Roger~G Mark. 2016.
\newblock Mimic-iii, a freely accessible critical care database.
\newblock \emph{Scientific data}, 3(1):1--9.

\bibitem[{Katz et~al.(2023)Katz, Bommarito, Gao, and Arredondo}]{katz2023gpt}
Daniel~Martin Katz, Michael~James Bommarito, Shang Gao, and Pablo Arredondo. 2023.
\newblock Gpt-4 passes the bar exam.
\newblock \emph{Available at SSRN 4389233}.

\bibitem[{Lin(2004)}]{lin-2004-rouge}
Chin-Yew Lin. 2004.
\newblock {ROUGE}: A package for automatic evaluation of summaries.
\newblock In \emph{Text Summarization Branches Out}, pages 74--81, Barcelona, Spain. ACL.

\bibitem[{Liu et~al.(2023{\natexlab{a}})Liu, Liu, Bannur, Pérez-García, Usuyama, Zhang, Naumann, Nori, Poon, Alvarez-Valle, Oktay, and Hyland}]{10.1162/tacl_a_00585}
Fangyu Liu, Qianchu Liu, Shruthi Bannur, Fernando Pérez-García, Naoto Usuyama, Sheng Zhang, Tristan Naumann, Aditya Nori, Hoifung Poon, Javier Alvarez-Valle, Ozan Oktay, and Stephanie~L. Hyland. 2023{\natexlab{a}}.
\newblock \href {https://doi.org/10.1162/tacl_a_00585} {{Compositional Zero-Shot Domain Transfer with Text-to-Text Models}}.
\newblock \emph{Transactions of the Association for Computational Linguistics}, 11:1097--1113.

\bibitem[{Liu et~al.(2023{\natexlab{b}})Liu, Hu, Zhang, Gai, FENG, and Liu}]{liu2023a}
Jiaxiang Liu, Tianxiang Hu, Yan Zhang, Xiaotang Gai, YANG FENG, and Zuozhu Liu. 2023{\natexlab{b}}.
\newblock \href {https://openreview.net/forum?id=DFAVQIxDqL} {A chat{GPT} aided explainable framework for zero-shot medical image diagnosis}.
\newblock In \emph{ICML 3rd Workshop on Interpretable Machine Learning in Healthcare (IMLH)}.

\bibitem[{Liu et~al.(2023{\natexlab{c}})Liu, Hyland, Bannur, Bouzid, Castro, Wetscherek, Tinn, Sharma, P{\'e}rez-Garc{\'\i}a, Schwaighofer, Rajpurkar, Khanna, Poon, Usuyama, Thieme, Nori, Lungren, Oktay, and Alvarez-Valle}]{liu-etal-2023-exploring-boundaries}
Qianchu Liu, Stephanie Hyland, Shruthi Bannur, Kenza Bouzid, Daniel Castro, Maria Wetscherek, Robert Tinn, Harshita Sharma, Fernando P{\'e}rez-Garc{\'\i}a, Anton Schwaighofer, Pranav Rajpurkar, Sameer Khanna, Hoifung Poon, Naoto Usuyama, Anja Thieme, Aditya Nori, Matthew Lungren, Ozan Oktay, and Javier Alvarez-Valle. 2023{\natexlab{c}}.
\newblock \href {https://doi.org/10.18653/v1/2023.emnlp-main.891} {Exploring the boundaries of {GPT}-4 in radiology}.
\newblock In \emph{Proceedings of the 2023 Conference on Empirical Methods in Natural Language Processing}, pages 14414--14445, Singapore. Association for Computational Linguistics.

\bibitem[{Loshchilov and Hutter(2019)}]{loshchilov2018decoupled}
Ilya Loshchilov and Frank Hutter. 2019.
\newblock \href {https://openreview.net/forum?id=Bkg6RiCqY7} {Decoupled weight decay regularization}.
\newblock In \emph{International Conference on Learning Representations}.

\bibitem[{Ma et~al.(2023)Ma, Wu, Wang, Xu, Wei, Liu, Guo, Cai, Zhang, Zhang et~al.}]{ma2023impressiongpt}
Chong Ma, Zihao Wu, Jiaqi Wang, Shaochen Xu, Yaonai Wei, Zhengliang Liu, Lei Guo, Xiaoyan Cai, Shu Zhang, Tuo Zhang, et~al. 2023.
\newblock Impressiongpt: an iterative optimizing framework for radiology report summarization with chatgpt.
\newblock \emph{arXiv preprint arXiv:2304.08448}.

\bibitem[{Messina et~al.(2022)Messina, Pino, Parra, Soto, Besa, Uribe, Andia, Tejos, Prieto, and Capurro}]{messina2022survey}
Pablo Messina, Pablo Pino, Denis Parra, Alvaro Soto, Cecilia Besa, Sergio Uribe, Marcelo Andia, Cristian Tejos, Claudia Prieto, and Daniel Capurro. 2022.
\newblock A survey on deep learning and explainability for automatic report generation from medical images.
\newblock \emph{ACM Computing Surveys (CSUR)}, 54(10s):1--40.

\bibitem[{Miura et~al.(2021)Miura, Zhang, Tsai, Langlotz, and Jurafsky}]{miura-etal-2021-improving}
Yasuhide Miura, Yuhao Zhang, Emily Tsai, Curtis Langlotz, and Dan Jurafsky. 2021.
\newblock \href {https://doi.org/10.18653/v1/2021.naacl-main.416} {Improving factual completeness and consistency of image-to-text radiology report generation}.
\newblock In \emph{Proceedings of the 2021 Conference of the North American Chapter of the Association for Computational Linguistics: Human Language Technologies}, pages 5288--5304, Online. Association for Computational Linguistics.

\bibitem[{Papineni et~al.(2002)Papineni, Roukos, Ward, and Zhu}]{papineni2002bleu}
Kishore Papineni, Salim Roukos, Todd Ward, and Wei-Jing Zhu. 2002.
\newblock {B}leu: a method for automatic evaluation of machine translation.
\newblock In \emph{Proc of the 40th Annual Meeting of the ACL}, pages 311--318, Philadelphia, Pennsylvania, USA. ACL, ACL.

\bibitem[{Paszke et~al.(2017)Paszke, Gross, Chintala, Chanan, Yang, DeVito, Lin, Desmaison, Antiga, and Lerer}]{paszke2017automatic}
Adam Paszke, Sam Gross, Soumith Chintala, Gregory Chanan, Edward Yang, Zachary DeVito, Zeming Lin, Alban Desmaison, Luca Antiga, and Adam Lerer. 2017.
\newblock Automatic differentiation in pytorch.

\bibitem[{Pino et~al.(2021)Pino, Parra, Besa, and Lagos}]{pino2021clinically}
Pablo Pino, Denis Parra, Cecilia Besa, and Claudio Lagos. 2021.
\newblock Clinically correct report generation from chest x-rays using templates.
\newblock In \emph{Machine Learning in Medical Imaging: 12th International Workshop, MLMI 2021, Held in Conjunction with MICCAI 2021, Strasbourg, France, September 27, 2021, Proceedings 12}, pages 654--663. Springer.

\bibitem[{Pino et~al.(2020)Pino, Parra, Messina, Besa, and Uribe}]{pino2020inspecting}
Pablo Pino, Denis Parra, Pablo Messina, Cecilia Besa, and Sergio Uribe. 2020.
\newblock Inspecting state of the art performance and nlp metrics in image-based medical report generation.
\newblock \emph{arXiv preprint arXiv:2011.09257}.

\bibitem[{Raffel et~al.(2020)Raffel, Shazeer, Roberts, Lee, Narang, Matena, Zhou, Li, and Liu}]{raffel2020exploring}
Colin Raffel, Noam Shazeer, Adam Roberts, Katherine Lee, Sharan Narang, Michael Matena, Yanqi Zhou, Wei Li, and Peter~J Liu. 2020.
\newblock Exploring the limits of transfer learning with a unified text-to-text transformer.
\newblock \emph{The Journal of Machine Learning Research}, 21(1):5485--5551.

\bibitem[{Reimers and Gurevych(2019)}]{reimers-gurevych-2019-sentence}
Nils Reimers and Iryna Gurevych. 2019.
\newblock \href {https://doi.org/10.18653/v1/D19-1410} {Sentence-{BERT}: Sentence embeddings using {S}iamese {BERT}-networks}.
\newblock In \emph{Proceedings of the 2019 Conference on Empirical Methods in Natural Language Processing and the 9th International Joint Conference on Natural Language Processing (EMNLP-IJCNLP)}, pages 3982--3992, Hong Kong, China. Association for Computational Linguistics.

\bibitem[{Romanov and Shivade(2018)}]{romanov2018lessons}
Alexey Romanov and Chaitanya Shivade. 2018.
\newblock \href {http://arxiv.org/abs/1808.06752} {Lessons from natural language inference in the clinical domain}.

\bibitem[{Shi et~al.(2023)Shi, Ma, Zhong, Mai, Li, Liu, and Huang}]{shi2023chatgraph}
Yucheng Shi, Hehuan Ma, Wenliang Zhong, Gengchen Mai, Xiang Li, Tianming Liu, and Junzhou Huang. 2023.
\newblock Chatgraph: Interpretable text classification by converting chatgpt knowledge to graphs.
\newblock \emph{arXiv preprint arXiv:2305.03513}.

\bibitem[{Smit et~al.(2020)Smit, Jain, Rajpurkar, Pareek, Ng, and Lungren}]{chexbert}
Akshay Smit, Saahil Jain, Pranav Rajpurkar, Anuj Pareek, Andrew~Y. Ng, and Matthew~P. Lungren. 2020.
\newblock \href {http://arxiv.org/abs/2004.09167} {Chexbert: Combining automatic labelers and expert annotations for accurate radiology report labeling using {BERT}}.
\newblock \emph{CoRR}, abs/2004.09167.

\bibitem[{Tanida et~al.(2023)Tanida, M{\"u}ller, Kaissis, and Rueckert}]{tanida2023interactive}
Tim Tanida, Philip M{\"u}ller, Georgios Kaissis, and Daniel Rueckert. 2023.
\newblock Interactive and explainable region-guided radiology report generation.
\newblock In \emph{Proceedings of the IEEE/CVF Conference on Computer Vision and Pattern Recognition}, pages 7433--7442.

\bibitem[{Vedantam et~al.(2015)Vedantam, Lawrence~Zitnick, and Parikh}]{vedantam2015cider}
Ramakrishna Vedantam, C~Lawrence~Zitnick, and Devi Parikh. 2015.
\newblock Cider: Consensus-based image description evaluation.
\newblock In \emph{Proc of the IEEE Conf. on Computer Vision and Pattern Recognition (CVPR)}, pages 4566--4575.

\bibitem[{Wadden et~al.(2019)Wadden, Wennberg, Luan, and Hajishirzi}]{wadden-etal-2019-entity}
David Wadden, Ulme Wennberg, Yi~Luan, and Hannaneh Hajishirzi. 2019.
\newblock \href {https://doi.org/10.18653/v1/D19-1585} {Entity, relation, and event extraction with contextualized span representations}.
\newblock In \emph{Proceedings of the 2019 Conference on Empirical Methods in Natural Language Processing and the 9th International Joint Conference on Natural Language Processing (EMNLP-IJCNLP)}, pages 5784--5789, Hong Kong, China. Association for Computational Linguistics.

\bibitem[{Wang et~al.(2022)Wang, Zhou, Wang, Vardhanabhuti, and Yu}]{wang2022multi}
Fuying Wang, Yuyin Zhou, Shujun Wang, Varut Vardhanabhuti, and Lequan Yu. 2022.
\newblock Multi-granularity cross-modal alignment for generalized medical visual representation learning.
\newblock \emph{Advances in Neural Information Processing Systems}, 35:33536--33549.

\bibitem[{Wu et~al.(2021)Wu, Agu, Lourentzou, Lourentzou, Sharma, Paguio, Yao, Dee, Mitchell, Kashyap, Giovannini, Celi, and Moradi}]{chestimagenome}
Joy~T Wu, Nkechinyere Agu, Ismini Lourentzou, Ismini Lourentzou, Arjun Sharma, Joseph~Alexander Paguio, Jasper~Seth Yao, Edward~C Dee, William Mitchell, Satyananda Kashyap, Andrea Giovannini, Leo~Anthony Celi, and Mehdi Moradi. 2021.
\newblock \href {https://datasets-benchmarks-proceedings.neurips.cc/paper_files/paper/2021/file/17e62166fc8586dfa4d1bc0e1742c08b-Paper-round2.pdf} {Chest imagenome dataset for clinical reasoning}.
\newblock In \emph{Proceedings of the Neural Information Processing Systems Track on Datasets and Benchmarks}, volume~1. Curran.

\bibitem[{Xu et~al.(2024)Xu, Chen, Johnston, Blankemeier, Varma, Langlotz, and Delbrouck}]{xu-etal-2024-overview}
Justin Xu, Zhihong Chen, Andrew Johnston, Louis Blankemeier, Maya Varma, Curtis Langlotz, and Jean-Benoit Delbrouck. 2024.
\newblock Overview of the first shared task on clinical text generation: Rrg24 and {``}discharge me!{''}.
\newblock In \emph{The 23rd Workshop on Biomedical Natural Language Processing and BioNLP Shared Tasks}, Bangkok, Thailand. Association for Computational Linguistics.

\bibitem[{Yang et~al.(2023)Yang, Panagopoulou, Zhou, Jin, Callison-Burch, and Yatskar}]{yang2023language}
Yue Yang, Artemis Panagopoulou, Shenghao Zhou, Daniel Jin, Chris Callison-Burch, and Mark Yatskar. 2023.
\newblock Language in a bottle: Language model guided concept bottlenecks for interpretable image classification.
\newblock In \emph{Proceedings of the IEEE/CVF Conference on Computer Vision and Pattern Recognition}, pages 19187--19197.

\bibitem[{Yasunaga et~al.(2022)Yasunaga, Leskovec, and Liang}]{yasunaga2022linkbert}
Michihiro Yasunaga, Jure Leskovec, and Percy Liang. 2022.
\newblock Linkbert: Pretraining language models with document links.
\newblock In \emph{Association for Computational Linguistics (ACL)}.

\bibitem[{Yu et~al.(2022)Yu, Endo, Krishnan, Pan, Tsai, Reis, Fonseca, Lee, Abad, Ng et~al.}]{yu2022evaluating}
Feiyang Yu, Mark Endo, Rayan Krishnan, Ian Pan, Andy Tsai, Eduardo~Pontes Reis, Eduardo Kaiser Ururahy~Nunes Fonseca, Henrique Min~Ho Lee, Zahra Shakeri~Hossein Abad, Andrew~Y Ng, et~al. 2022.
\newblock Evaluating progress in automatic chest x-ray radiology report generation.
\newblock \emph{medRxiv}, pages 2022--08.

\bibitem[{Zhang et~al.(2020{\natexlab{a}})Zhang, Kishore, Wu, Weinberger, and Artzi}]{bert-score}
Tianyi Zhang, Varsha Kishore, Felix Wu, Kilian~Q. Weinberger, and Yoav Artzi. 2020{\natexlab{a}}.
\newblock \href {https://openreview.net/forum?id=SkeHuCVFDr} {Bertscore: Evaluating text generation with bert}.
\newblock In \emph{International Conference on Learning Representations}.

\bibitem[{Zhang et~al.(2020{\natexlab{b}})Zhang, Merck, Tsai, Manning, and Langlotz}]{zhang-etal-2020-optimizing}
Yuhao Zhang, Derek Merck, Emily Tsai, Christopher~D. Manning, and Curtis Langlotz. 2020{\natexlab{b}}.
\newblock \href {https://doi.org/10.18653/v1/2020.acl-main.458} {Optimizing the factual correctness of a summary: A study of summarizing radiology reports}.
\newblock In \emph{Proceedings of the 58th Annual Meeting of the Association for Computational Linguistics}, pages 5108--5120, Online. Association for Computational Linguistics.

\end{thebibliography}
